\documentclass{article}
\usepackage{arxiv}

\usepackage[utf8]{inputenc} 
\usepackage[T1]{fontenc}    
\usepackage{hyperref}       
\hypersetup{hypertex=true,
	colorlinks=true,
	linkcolor=blue,
	anchorcolor=blue,
	citecolor=blue
}
\usepackage{url}            
\usepackage{booktabs}       
\usepackage{amsfonts}       
\usepackage{nicefrac}       
\usepackage{microtype}      
\usepackage{lipsum}		
\usepackage{graphicx}
\usepackage{natbib}
\usepackage{doi}

\usepackage{subfigure}
\usepackage[ruled]{algorithm2e}
\usepackage{amsmath}
\usepackage{amssymb}
\usepackage{mathtools} 
\usepackage{multirow}
\usepackage{graphicx} 

\title{Temperature Field Inversion of Heat-Source Systems via Physics-Informed Neural Networks}

\author{ {\hspace{1mm}Xu Liu} \\
	Defense Innovation Institute\\
	Chinese Academy of Military Science\\
	China \\
	\texttt{liuxu18054448691@126.com}
	\And
	{\hspace{1mm}Wei Peng} \\
	Defense Innovation Institute\\
	Chinese Academy of Military Science\\
	China \\
	\texttt{weipeng0098@126.com} \\
	\And
	{\hspace{1mm}Zhiqiang Gong} \\
	Defense Innovation Institute\\
	Chinese Academy of Military Science\\
	China \\
	\texttt{gongzhiqiang13@nudt.edu.cn} \\
    \And
	{\hspace{1mm}Weien Zhou} \\
	Defense Innovation Institute\\
	Chinese Academy of Military Science\\
	China \\
	\texttt{weienzhou@outlook.com} \\
	\And
	{\hspace{1mm}Wen Yao }\thanks{Corresponding author} \\
	Defense Innovation Institute\\
	Chinese Academy of Military Science\\
	China \\
	\texttt{Wendy0782@126.com} \\
}

\renewcommand{\shorttitle}{\textit{arXiv} Template}

\hypersetup{
pdftitle={A template for the arxiv style},
pdfsubject={q-bio.NC, q-bio.QM},
pdfauthor={David S.~Hippocampus, Elias D.~Striatum},
pdfkeywords={First keyword, Second keyword, More},
}

\begin{document}
\maketitle

\begin{abstract}
Temperature field inversion of heat-source systems (TFI-HSS) with limited observations is essential to monitor the system health. 
Although some methods such as interpolation have been proposed to solve TFI-HSS, those existing methods ignore correlations between data constraints and physics constraints, causing the low precision. 
In this work, we develop a physics-informed neural network-based temperature field inversion (PINN-TFI) method to solve the TFI-HSS task and a coefficient matrix condition number based position selection of observations (CMCN-PSO) method to select optima positions of noise observations. 
For the TFI-HSS task, the PINN-TFI method encodes constrain terms into the loss function, thus the task is transformed into an optimization problem of minimizing the loss function. 
In addition, we have found that noise observations significantly affect reconstruction performances of the PINN-TFI method. 
To alleviate the effect of noise observations, the CMCN-PSO method is proposed to find optima positions, where the condition number of observations is used to evaluate positions.
The results demonstrate that the PINN-TFI method can significantly improve prediction precisions and the CMCN-PSO method can find good positions to acquire a more robust temperature field.
\end{abstract}

\keywords{Temperature field inversion of heat-source systems \and Physics-informed neural network \and Position selection\and Condition number\and Upper bound of error}

\section{Introduction}
\label{intro}
Heat management \citep{grujicic2005effect,laloya2015heat} is essential for heat-source systems (HSS), where heat is generated internally, especially over systems that involve components with smaller size and higher power density. 
Statistically, when the allowable temperature is exceeded \citep{emam2019thermal}, the failure rate of such electronic devices doubles for the temperature rise of $10\sim20^{\circ} \mathrm{C}$, and its failure rate decreases by about $4$ $\%$ for the temperature reduction of $1^{\circ} \mathrm{C}$. 
Temperature monitoring is of importance to ensure the normal work of components in heat management systems. 
Temperature field inversion (TFI) through limited temperature transducers (observations) is an important method to monitor component temperatures \citep{evans2002pacific,kong2020numerical}. 
Generally, TFI requires a large number of temperature transducers (observations), which causing expensive costs and affecting the availability of systems. 
Moreover, too many transducers are not accessible in most real-life applications, which increases the difficulty to reconstruct the whole temperature field. 
Therefore, temperature field inversion of heat-source systems (TFI-HSS) in a small data setting is an urgent and challenging problem to be solved in practical engineering systems.

The mechanism of the TFI calculation and the temperature adjustment is complicated, thus it is difficult to reconstruct the temperature field through reasonable formula transformations. 
Interpolation is an easy-implement but low-precision procedure \citep{steffensen2006interpolation}. 
A range of interpolation methods essentially use discrete points to fit the function, such as Gaussian process regression (kriging) \citep{stein2012interpolation}, random forest regression \citep{segal2004machine}, nearest neighbor \citep{jiang2015quantum}. 
In addition to this, reconstructing intensities of heat sources is also a feasible method to reconstruct the temperature field. 
A boundary element method has been proposed \citep{le2001method,le2000experimental} to identify intensities of point-heat sources, which is a time regularization based procedure. Two sources that have different shapes and close distances may cause inaccurate estimations. 
To this end, a numerical algorithm coupled with the concept of future time is proposed \citep{yang1999determination}. 
Subsequently, \cite{shuai2011inversion} propose an adaptive iterative algorithm. 
They regard intensities as variables and then use a multivariate iterative approach to minimize the discrepancy between prediction and measurements. 
This iterative method has a large search space and is low efficient, which is appropriate to solve reconstruction problems with a small number of iterations. 
In the small data setting, these existing TFI methods ignore correlations between data constraints and physics constraints (governing equations), resulting in low precisions and even unavailability.

Recently, physics-informed neural networks (PINNs) have been successfully used to solve partial differential equations (PDEs) or PDE-based problems \citep{raissi2019physics,shen2021physics,zobeiry2021physics} by adding data constraints and physics constraints in the loss function to constrain the space of admissible solutions.
Then, the problem of solving PDEs is transformed into an optimization problem of minimizing the loss function. Furthermore, benefited from physics constraints, the PINN can reduce the requirement of label data and be easily applied to the small data regime. 
In addition, compared to traditional numerical approaches such as finite difference
method \citep{narasimhan1976integrated}, mesh-free PINNs can easily solve irregular-domain problems \citep{gao2020phygeonet} and avoid the curse of dimensionality \citep{grohs2018proof}. 
Moreover, PINNs have offered a powerful new paradigm for dealing with diverse inverse problems, including systems biology \citep{yazdani2020systems}, hidden fluid mechanics \citep{raissi2020hidden,tartakovsky2020physics}, biomedicine \citep{sahli2020physics,kissas2020machine} and the parameter identification of wave, stochastic and fractional PDEs \citep{song2020wavefield,tartakovsky2020physics,pang2019fpinns}. 
By introducing a data-constrained term with the mean-squared error, Pang et al. \citep{pang2019fpinns} use PINNs to identify the diffusion coefficients in fractional advection-diffusion equations from synthetic data. 
Subsequently, parameterizing the data-constrained term as polynomial functions by prior knowledge, Chen et al. \citep{chen2021solving} extend PINNs to solve the inverse problem that the initial probability density function of Fokker-Plank equations is agnostic. 
Similarly, considering the unknowable initial and boundary condition, PINNs are employed to reconstruct the flow field based on scattered observation about the velocity \citep{raissi2020hidden,tartakovsky2020physics}. 
For higher-dimensional problems, PINNs are employed by Kissas et al. \citep{kissas2020machine} to recover the entire three-dimensional velocity flow field given four-dimensional flow magnetic resonance imaging data, which shows that PINNs have a great feature for high-dimensional and real-life problems. 
TFI-HSS is essentially an inverse problem, where intensities of heat sources are unknown. 
Thus, PINNs are promising for solving TFI-HSS. 

However, there still exist two major difficulties in applying PINNs to solve TFI-HSS.
\begin{enumerate}
	\item[$\bullet$] Using PINNs to reconstruct efficiently the temperature field is an urgent problem, given a limited number of observations and the governing equation.
	\item[$\bullet$] Observations are inevitably perturbed in real-life engineering systems. Given noise observations, positions of observations significantly affect the reconstructed performance. At present, there is no appropriate strategy to choose reasonable positions to alleviate the effect of noises.
\end{enumerate}

To solve the first issue, we employ the physics characteristic and develop a novel physics-informed neural network-based temperature field inversion (PINN-TFI) method to solve TFI-HSS. The PINN-TFI method encodes physics and data constraint terms into the loss function, thus the problem of TFI-HSS is transformed into an optimization problem of minimizing the physics-informed loss function. In addition, large computational costs are one of main limitations of the PINN for TFI-HSS. 
In prior works, transfer learning has been widely used to improve the convergence of training for diverse deep learning problems \citep{yang2013theory}. 
Inspired by this idea, we first utilize transfer learning to obtain a retrained PINN and then use the retrained PINN with observations for TFI. 
To solve the latter one, we develop a coefficient matrix condition number based position selection of observations (CMCN-PSO) method, where the condition number of noise observations is used to evaluate positions. In summary, the following contributions are made in this paper. 

\begin{enumerate}
	\item[$\bullet$] We refine the Temperature Field Inversion of Heat-Source Systems (TFI-HSS) task from engineering applications by giving the formula expression.
	\item[$\bullet$] We propose a physics-informed neural network-based temperature field inversion (PINN-TFI) method to reconstruct the temperature field, where transfer learning strategy is used to accelerate the training process.
	\item[$\bullet$]  We develop a coefficient matrix condition number based position selection of observations (CMCN-PSO) method to alleviate the effect of noise observations for the PINN-TFI method.
\end{enumerate}

The experiments considering three types of boundary conditions have been conducted. Experiment results show that the PINN-TFI method has superior performance for the TFI-HSS task. Moreover, the CMCN-PSO method can find good positions of observations. Given noise observations, the PINN-TFI method uses these observations to reconstruct a more robust temperature field.

\section{Temperature Field Inversion of Heat-Source Systems (TFI-HSS)}
\label{sec:2}
Heat-source systems (HSS) will be discussed in this chapter, where observations and the principle of heat transfer equations are taken into account. In this work, HSS is modeled as a two-dimensional plane. 
There are several electronic components on this plane and these components generate heat when working. 
Each component is simplified as a heat source. 
The purpose of the TFI-HSS task is to reconstruct the temperature field given some observations, which is very important to ensure working temperatures of components and monitor the HSS health.

Generally, given a HSS with m observations, the TFI-HSS task can be written as 
\begin{equation}
\label{eq1}
\arg \min _{\theta}\left(\sum\limits_{i=1}^{m}\left|T_{\theta}\left(x_{obs}^{i}, y_{obs}^{i} \mid \phi\right)-T_{obs}^{i}\left(x_{obs}^{i}, y_{obs}^{i}\right)\right|\right),
\end{equation}
where $T_{\theta}(\cdot)$ is the model with the parameter $\theta$ to reconstruct the temperature field, unknown $\phi$ is the intensity distribution, and $T_{obs}^{i}$ represents
the temperature of observations. In addition, the stead-state temperature field $T$ has to satisfy
\begin{equation}
\label{2}
\begin{array}{r}
\frac{\partial}{\partial x}\left(k \frac{\partial T}{\partial x}\right)+\frac{\partial}{\partial y}\left(k \frac{\partial T}{\partial y}\right)+\phi(x, y)=0, \vspace{1ex}\\
Boundary:  \quad T=T_{0} \quad  or  \quad k \frac{\partial T}{\partial \mathbf{n}}=0 \\
or  \quad k \frac{\partial T}{\partial \mathbf{n}}=h\left(T-T_{0}\right),
\end{array}
\end{equation}
where $k$ is the thermal conductivity and is set to be 1 in this work, $\phi$ is the intensity distribution function, and $T_{0}$ is the temperature value of the isothermal boundary or the convective boundary conditions. Eq.(\ref{2}) represents three boundary conditions including Dirichlet (isothermal), Neumann (adiabatic), and Robin (convective) boundary conditions, where $h$ denotes the convective heat transfer coefficient.

The area of a single piece of heat source is set as a rectangle in this work, modeled as
\begin{equation}
\label{eq3}
\phi(x, y)=\left\{\begin{array}{ll}
\phi_{i}, & (x, y) \in \Gamma_{i}, i=1,2, \cdots, n \\
0, & (x, y) \notin \Gamma_{i}
\end{array}\right.,
\end{equation}
where $\phi_{i}$ represents the (constant) intensity of $i$th heat source, $\Gamma_{i}$ represents the area of $i$th heat source, and $n$ is the number of heat sources. In summary, the TFI-HSS task can be expressed as a constrained optimization problem, which is formulated as
\begin{equation}
\label{eq_optim}
\begin{array}{l}
\min\limits_{\theta}\left(\sum\limits_{i=1}^{m}\left|T_{\theta}\left(x_{obs}^{i}, y_{obs}^{i} \mid \phi\right)-T_{obs}^{i}\left(x_{obs}^{i}, y_{obs}^{i}\right)\right|\right), \vspace{1.2ex} \\s.t. \quad \frac{\partial}{\partial x}\left(k \frac{\partial T}{\partial x}\right)+\frac{\partial}{\partial y}\left(k \frac{\partial T}{\partial y}\right)+\phi(x, y)=0,\vspace{1.2ex} \\
\quad T=T_{0}\quad or \quad k \frac{\partial T}{\partial \mathbf{n}}=0  \quad
or  \quad k \frac{\partial T}{\partial \mathbf{n}}=h\left(T-T_{0}\right).
\end{array}
\end{equation}

\begin{figure}[htbp]
	\centering
	\subfigure[The VP problem]{
		\centering
		\includegraphics[width=0.147\textheight]{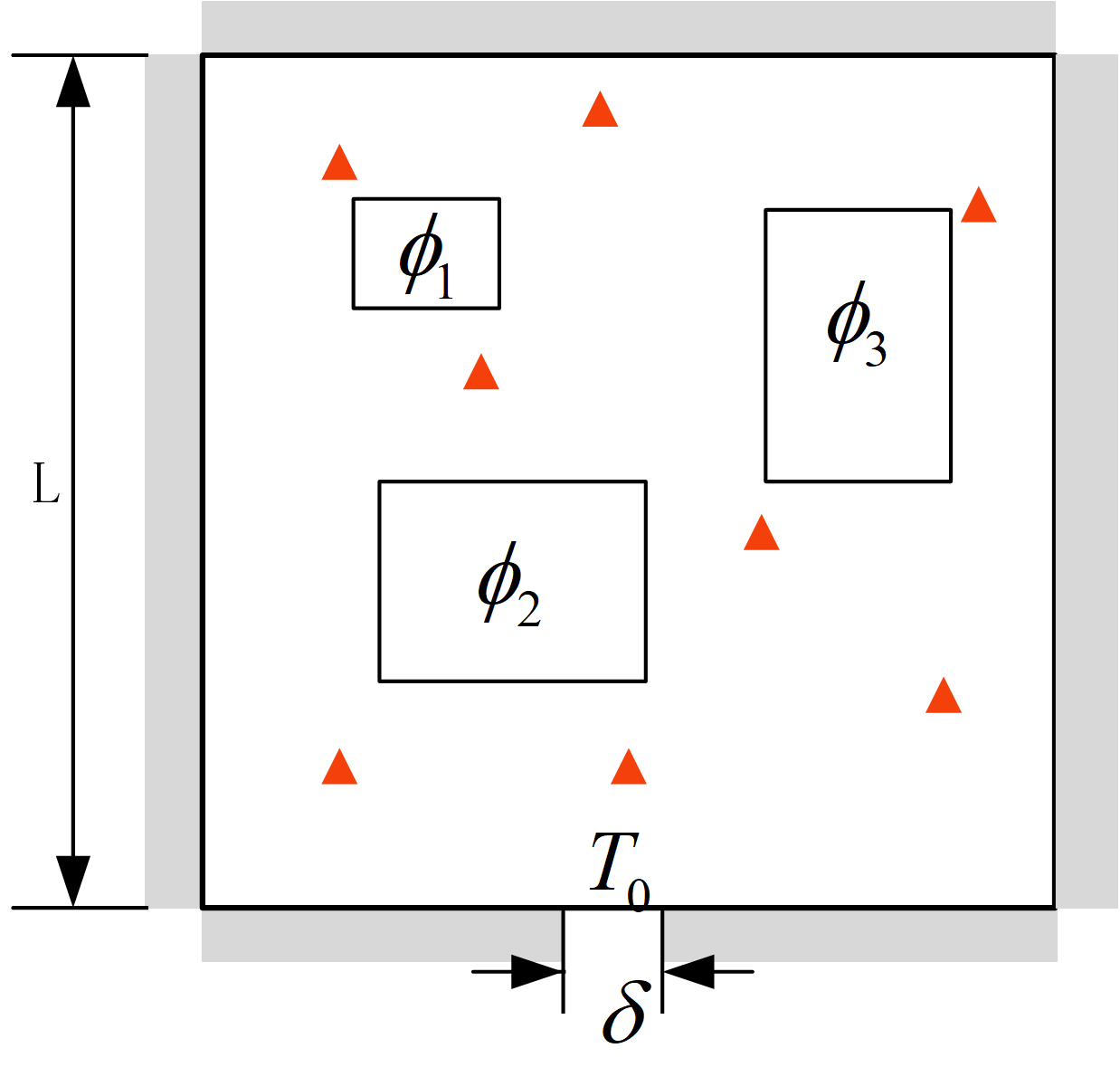}
		\label{fig:1a}
	}
	\subfigure[The VB problem]{
		\centering
		\includegraphics[width=0.15\textheight]{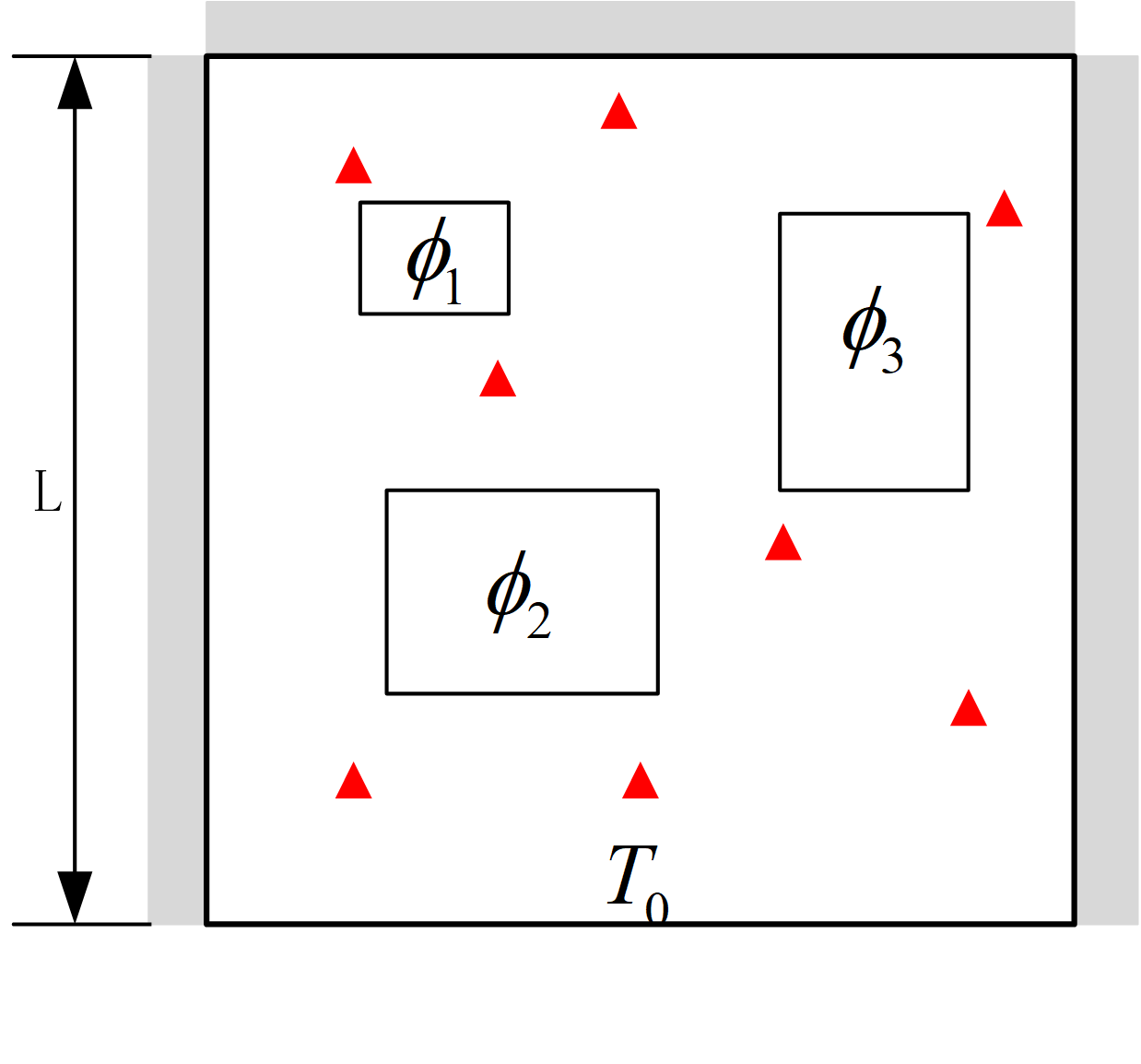}
		\label{fig:1b}
	}
	\caption{The illustration of the VP and the VB problem in a square domain.}
	\label{fig:1}
\end{figure}

For further research on the TFI-HSS task, the volume-to-point (VP) and the volume-to-boundary (VB) problems are commonly used in heat conduction problems \citep{aslan2018heat,chen2016optimization,chen2021deep}. From the perspective of mathematical models, the difference between VB and VP problems is the boundary condition. For the VP heat conduction problem, it considers a finite heat-generating volume cooled by a small patch of the common heat sink with the temperature of $T_{0}$ located on the middle of the bottom boundary, shown in Fig. \ref{fig:1a}. For the VB problem, the condition of each boundary can be isothermal, adiabatic, or convective, shown in Fig. \ref{fig:1b}.

\section{Methodology}

\subsection{Overview}
\label{sec:3.1}
\begin{figure*}[htp]
	\centering
	\includegraphics[width=0.68\linewidth]{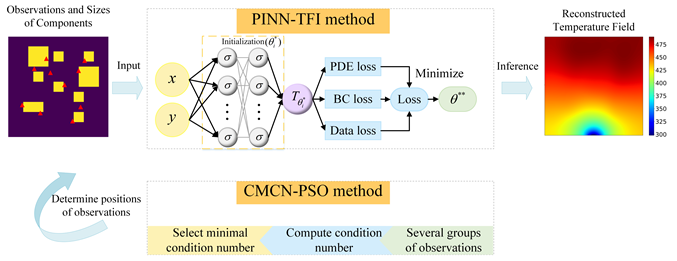}
	\caption{Conceptual flow of the PINN-TFI method and the CMCN-PSO method.}
	\label{fig:flowchart}
\end{figure*}

For the TFI-HSS task, given the unknown intensity distribution, temperature field $T(x, y)$ can not be solved numerically by discretizing Eq.(\ref{2}) using the finite difference method \citep{narasimhan1976integrated} or the finite volume method \citep{eymard2000finite}. 
Interpolations can reconstruct the temperature field. However, reconstruction errors are high and costs of much observation acquisition are prohibitive. 
Moreover, the vast majority of state-of-the-art machine learning techniques (e.g., deep/convolutional/recurrent neural networks) may become intractable when it comes to the small data regime. 
To this end, a novel Physic-Informed Neural Network-based Temperature Field Inversion (PINN-TFI) method is proposed to reconstruct the temperature field in the small data setting. 
The PINN-TFI method is expected to provide a rapid and accurate prediction of the temperature field. 
In addition, we have found that noise observations significantly affect reconstructed performance of the PINN-TFI method but reasonably position selection of observations can alleviate the effect of noise observations. 
In this work, a coefficient matrix condition number based position selection of observations (CMCN-PSO) method is proposed to find optimal positions of noise observations. 
Then, the optimal observations are used by the PINN-TFI method to reconstruct a more robust temperature field. 
A schematic diagram of the PINN-TFI method and the CMCN-PSO method is shown in Fig. \ref{fig:flowchart}. 
Details of the PINN-TFI method and the CMCN-PSO method will be presented in the following subsections.

\subsection{Physics-informed neural network-based temperature field inversion (PINN-TFI) method}
\label{sec:3.2}
TFI-HSS can be considered as an inverse problem, where the temperature field can be learned from observations. 
To solve the inverse problem, we propose a Physics-Informed Neural Network based temperature field inversion (PINN-TFI) method, where the transfer learning strategy is used to accelerate the training process. 
Concretely, the weights and biases for predicting a new temperature field are initialized with those for the temperature field under rated intensity values. 
Then, the weights and biases are trained to learn from observations. 
The PINN-TFI method is mainly divided into two parts, namely the model initialization and the model for temperature field inversion, and its detailed description is in algorithm \ref{alg_PINN_TFI}. 
The key of the PINN-TFI method is that the constrained optimization problem of TFI-HSS is transformed into an unconstrained optimization problem of minimizing the loss function by adding data and physics constrain terms to the loss function.

\begin{figure*}[htbp]
	\centering
	\includegraphics[width=0.75\linewidth]{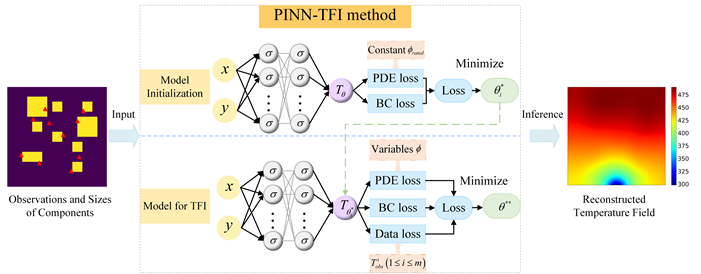}
	\caption{Conceptual flow of the PINN-TFI method with observations.}
	\label{fig:PINN_TFI}
\end{figure*}

Let $T_{\theta}(\cdot)$ be a NN of depth $D$, where $\theta$ is the vector of weights and bias. The NN has an input layer, $D-2$ hidden layers and an output layer, which is a mapping from $\mathbb{R}^{d}$ into $\mathbb{R}^{N}$. In this work, a Multi-Layer Perception (MLP) is employed with the activation function $\sigma(\cdot)$, i.e,

\begin{equation}T_{\theta}:=L_{D} \circ \sigma \circ L_{D-1} \circ \sigma \circ \cdots \circ \sigma \circ L_{1},
\end{equation}
where 
\begin{equation}\begin{array}{l}
L_{1}(x) :=W_{1} x+b_{1}, \quad W_{1} \in \mathbb{R}^{d_{1} \times d}, b_{1} \in \mathbb{R}^{d_{1}}, \\
L_{k}(x) :=W_{k} x+b_{k}, \quad W_{ik} \in \mathbb{R}^{d_{k} \times d_{k-1}}, b_{k} \in \mathbb{R}^{d_{k}},\\
L_{D}(x) :=W_{D} x+b_{D}, \quad W_{D} \in \mathbb{R}^{N \times d_{D-1}}, b_{D} \in \mathbb{R}^{N}.
\end{array}\end{equation}

\textbf{The model initialization}: The rated intensity distribution $\phi_{rated}$ of heat sources is available. PINNs are employed to solve Eq.(\ref{2}) without observations and obtain initialization parameter $\theta_{i}^{*}$. The left-hand-side of Eq.(\ref{2}) is first defined as
\begin{equation}
\begin{array}{r}
f = \frac{\partial}{\partial x}\left(k \frac{\partial T}{\partial x}\right)+\frac{\partial}{\partial y}\left(k \frac{\partial T}{\partial y}\right)+\phi(x, y).
\end{array}
\end{equation}

\begin{algorithm*}[!ht]
	\caption{\textbf{The Physics-Informed Neural Network-based Temperature Field Inversion method (PINN-TFI method)}}
	\label{alg_PINN_TFI}
	\LinesNumbered
	\KwIn{
		\\$\phi_{rated}$, the intensity distribution function under the rated powers,
		\\$\mathcal{P}_{data} = \{T(x_{obs}^{i},y_{obs}^{i}),i = 1,\cdots,n\}$, the observation set.
	}
	\KwOut{\\$\hat{T}(x,y,\theta^{**})$ as the surrogate of the temperature field}
	Construct a NN with the parameter $\theta$, $T_{\theta}(\cdot)$.\\
	
	// \textbf{The model initialization}\\
	Specify two training sets $\mathcal{P}_{pde}^{i}$ and $\mathcal{P}_{bc}^{i}$ for PDE and boundary conditions.\\
	Calculate PDE loss $L_{pde}^{i}$ and BC loss $L_{bc}^{i}$ by Eq.(\ref{pde_bc_ini}).\\
	Obtain total loss $L_{PINN}^{i}(\theta)$ by Eq.(\ref{PDE_ini}).\\
	Train the NN by minimizing the loss function $\mathcal{L}_{PINN}(\theta)$ until finding the best parameters $\theta_{i}^{*}$.\\
	// \textbf{The model for temperature field inversion}\\
	The NN is initialized with the parameter $\theta_{i}^{*}$.\\
	Specify two training sets $\mathcal{P}_{pde}$ and $\mathcal{P}_{bc}$ for PDE and boundary conditions and confirm $\mathcal{P}_{data}$.\\
	Set $\phi$ to be a variable initialized with the value $\phi_{rated}$ and put it in the optimizer.\\
	Calculate the $L_{pde}$, $L_{bc}$ and $L_{data}$ by Eq.(\ref{pde_bc_data}).\\
	Obtain the $L_{PINN}(\theta^{*}_{i})$ by Eq.(\ref{PDE_data}).\\
	Train the NN by minimizing the loss $L_{PINN}(\theta^{*}_{i})$ until finding the best parameters $\theta^{**}$.
\end{algorithm*}

Then, the training sets $\mathcal{P}_{pde}^{i}$ and $\mathcal{P}_{bc}^{i}$ for PDE and boundary conditions (e.g., randomly sampling points in the domain and boundary) are specified. 
To measure the difference between the NN and physics constraints, the loss function including partial differentiable equation loss (PDE loss), boundary condition loss (BC loss) is defined by
\begin{equation}
\label{PDE_ini}
\mathcal{L}_{PINN}^{i}\left(\theta\right)=w_{pde}^{i} \mathcal{L}_{pde}^{i}+w_{bc}^{i} \mathcal{L}_{bc}^{i},
\end{equation}
where
\begin{equation}
\label{pde_bc_ini}
\begin{array}{l}
\mathcal{L}_{pde}^{i}=\frac{1}{\left|\mathcal{P}_{pde}^{i}\right|} \sum\limits_{(\mathbf{x}, \mathbf{y}) \in \mathcal{P}_{pde}^{i}}\left\|f(T_{\theta}, x, y, \phi_{rated})\right\|_{2}^{2}, \\
\mathcal{L}_{bc}=\frac{1}{\left|\mathcal{P}_{bc}^{i}\right|} \sum\limits_{(\mathbf{x}, \mathbf{y}) \in \mathcal{P}_{bc}^{i}}\left\|T_{\theta}-T_{0}\right\|_{2}^{2} \quad or\\
\mathcal{L}_{bc}=\frac{1}{\left|\mathcal{P}_{bc}^{i}\right|} \sum\limits_{(\mathbf{x}, \mathbf{y}) \in \mathcal{P}_{bc}^{i}}\left\|k \frac{\partial T_{\theta}}{\partial \mathbf{n}}\right\|_{2}^{2},
\end{array}
\end{equation}
and $w_{pde}^{i}$ as well as $w_{bc}^{i}$ are weight hyper-parameters, which need to be predefined. $\mathcal{P}_{pde}^{i}$ and $\mathcal{P}_{bc}^{i}$ denote the sets of residual points from PDE and boundary conditions, respectively.

Finally, the NN is trained by minimizing the loss function $L_{PINN}(\theta)^{i}$ until finding the best parameters $\theta_{i}^{*}$.

\textbf{The model for temperature field inversion}: First, transfer learning strategy is used to initialize the NN model with parameter $\theta_{i}^{*}$ for accelerating the training process. When components are working normally, then real intensity distribution $\phi$ is unknown, which is set to be a variable initialized the value $\phi_{rated}$ and is put in the optimizer as trainable parameters. In addition to training sets $\mathcal{P}_{f}$ and $\mathcal{P}_{bc}$ for PDE and boundary conditions, the observation set $\mathcal{P}_{data}$ is confirmed. By adding a data constrained term, the loss function is formulated as
\begin{equation}
\label{PDE_data}
\mathcal{L}_{PINN}(\theta_{i}^{*}) = w_{pde} \mathcal{L}_{pde}+ w_{bc} \mathcal{L}_{bc}+ w_{data} \mathcal{L}_{data},
\end{equation}
where 
\begin{equation}
\label{pde_bc_data}
\begin{array}{l}
\mathcal{L}_{p d e}=\frac{1}{\left|\mathcal{P}_{pde}\right|} \sum\limits_{(\mathbf{x}, \mathbf{y}) \in \mathcal{P}_{pde}}\left\|f(T_{\theta_{i}^{*}}, x, y, \phi)\right\|_{2}^{2}, \\
\mathcal{L}_{data}=\frac{1}{\left|\mathcal{P}_{data}\right|} \sum\limits_{(\mathbf{x}, \mathbf{y}) \in \mathcal{P}_{data}}\left\|T_{\theta_{i}^{*}}-T_{o b s}\right\|_{2}^{2},\\
\mathcal{L}_{bc}=\frac{1}{\left|\mathcal{P}_{bc}\right|} \sum\limits_{(\mathbf{x}, \mathbf{y}) \in \mathcal{P}_{bc}}\left\|T_{\theta_{i}^{*}}-T_{0}\right\|_{2}^{2} \quad or\\
\mathcal{L}_{bc}=\frac{1}{\left|\mathcal{P}_{bc}\right|} \sum\limits_{(\mathbf{x}, \mathbf{y}) \in \mathcal{P}_{bc}}\left\|k \frac{\partial T_{\theta_{i}^{*}}}{\partial \mathbf{n}}\right\|_{2}^{2},
\end{array}
\end{equation}
and $w_{pde}$, $w_{bc}$ as well as $w_{data}$ are predefined weight hyper-parameters.
Thus, the constrained optimization problem in Eq.(\ref{2}) is transformed into an unconstrained optimization problem of minimizing the loss $\mathcal{L}_{PINN}(\theta_{i}^{*})$ in Eq.(\ref{PDE_data}).
In the last step, the training procedure is to search for the best parameter $\theta^{**}$ by minimizing the loss $\mathcal{L}_{PINN}(\theta_{i}^{*})$, where \textit{automatic differentiation} is used to minimize the highly nonlinear and non-convex loss.

In summary, the PINN-TFI-method can reconstruct the temperature field with limited observations.
However, in real-life engineering systems, observations are inevitably perturbed by noises, which significantly affects reconstruction performances of the PINN-TFI-method. We have found that reasonable position selection of observations can alleviate the effect of noise observations.

\subsection{The coefficient matrix condition number based position selection of observations method}
\label{sec:3.3}

\begin{figure*}[htbp]
	\centering
	\includegraphics[width=0.9\linewidth]{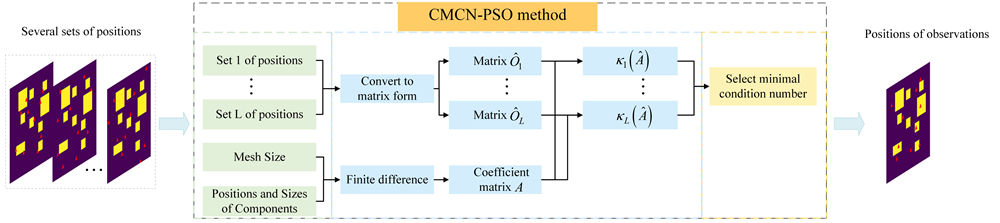}
	\caption{Conceptual flow of the CMCN-PSO method with observations.}
	\label{fig:CMCN}
\end{figure*}

\begin{algorithm*}[!h]
	\caption{\textbf{A coefficient matrix condition number based position selection of observations (CMCN-PSO) method}}
	\label{alg_CMCN-PSO}
	\LinesNumbered
	\KwIn{\\$\left\{\left(x_{1}, y_{1}\right),\ldots, \left(x_{m}, y_{m}\right)\right\}$, $\ldots$, $\left\{\left(x_{L m+1}, y_{L m+1}\right), \ldots,\left(x_{(L+1) m}, y_{(L+1) m}\right)\right\}$, several sets of positions.
		\\The positions and sizes of $n$ components
		\\ The mesh size $K \times K$
	}
	\KwOut{\\The optimal position of observations $\hat{O}_{optimal}$}
	// Determine the coefficient matrix $A$\\
	Mesh the domain as a $K \times K$ grid system and mesh the components in the grid system\\
	Number the $K \times K$ grid points from left to right, bottom to top\\
	Calculate the coefficient matrix $A_{1}$ by finite difference according to the boundary conditions\\
	\For{i=$1,2, \dots, K$}
	{
		\eIf{grid points is location in the compoments' area}
		{
			Set the $i$th row of the maxtrix $A_{2}$ to $-h^{2}$
		}
		{
			Set the $i$th row of the maxtrix $A_{2}$ to $0$
		}
	}	
	Construct the matrix $A$ by $A=\left(A_{1}, A_{2}\right)^{T}$\\

	// Determine the matrix $\hat{O}$ according to positions of observations\\
	\For{j=$1,2, \dots, L$}
	{
		\For{i=$1,2, \dots, n$}
		{
			Set the $i$th column of the maxtrix $\hat{O}_{j}$ to $1$ and other columns to $0$
		}
	}
	// Calculate the condition number\\
	Calculate the condition number of $\hat{A}$ by the function (e.g., $np.linalg.cond(\hat{A})$)\\
	\For{j=$1,2, \dots, L$}
	{
		$\kappa_{j}(\hat{A})=cond((\lambda A,\hat{O}_{j})^{T})$($\lambda$ is set to be 1 in this work)
	}
	//Select the optimal positions\\
	Select the minimal condition number $\kappa_{optimal}=min(\kappa_{1}, \dots, \kappa_{L})$\\
	Find the optimal positions $\hat{O}_{optimal}$ (corresponding to $\kappa_{optimal}$)
\end{algorithm*}

\begin{figure}[htbp]
	\centering
	\subfigure[Grid discretization]{
		\centering
		\includegraphics[width=0.15\textheight]{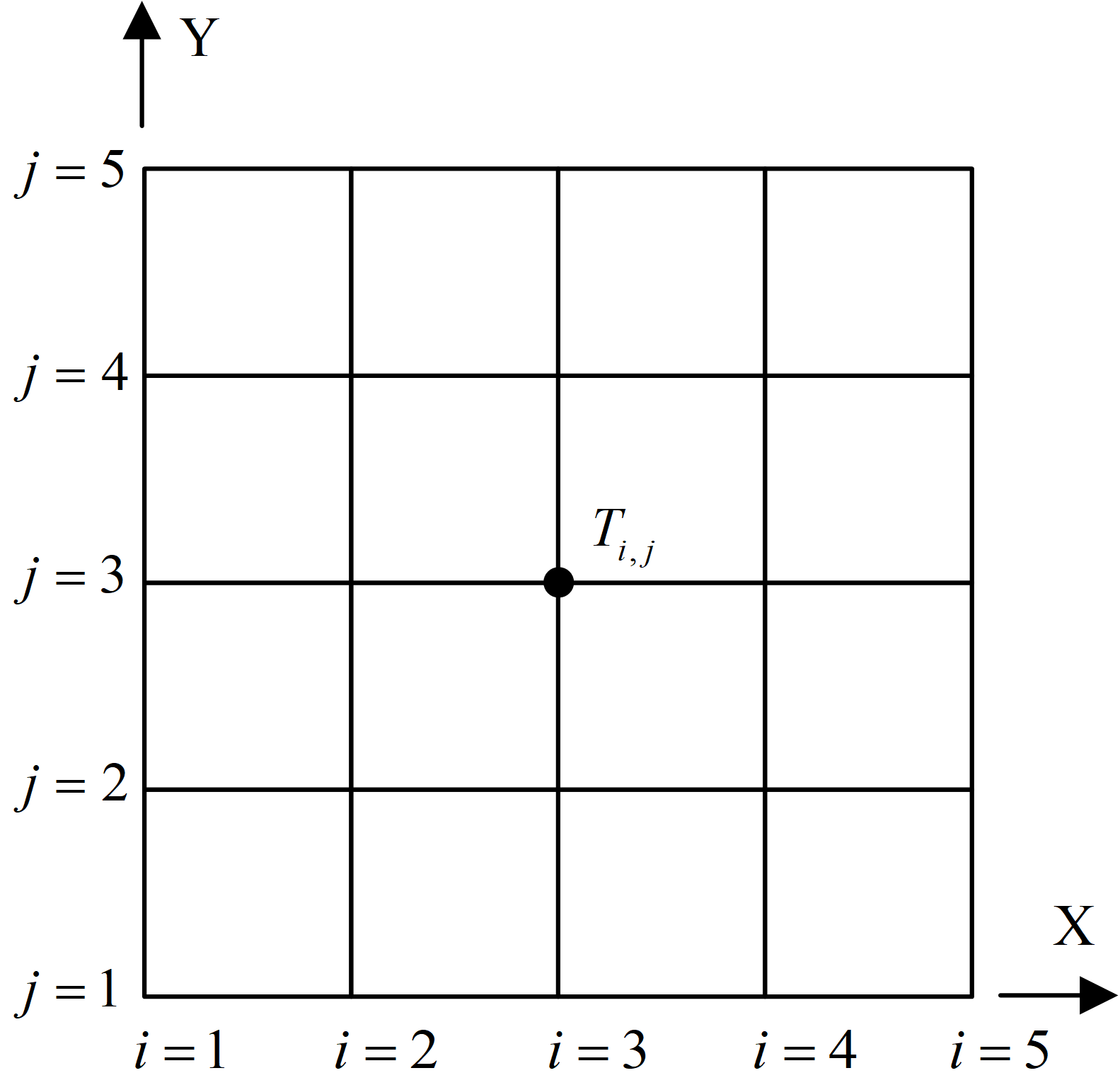}
		\label{fig:mesh1}
	}
	\subfigure[Sorted grid points]{
		\centering
		\includegraphics[width=0.15\textheight]{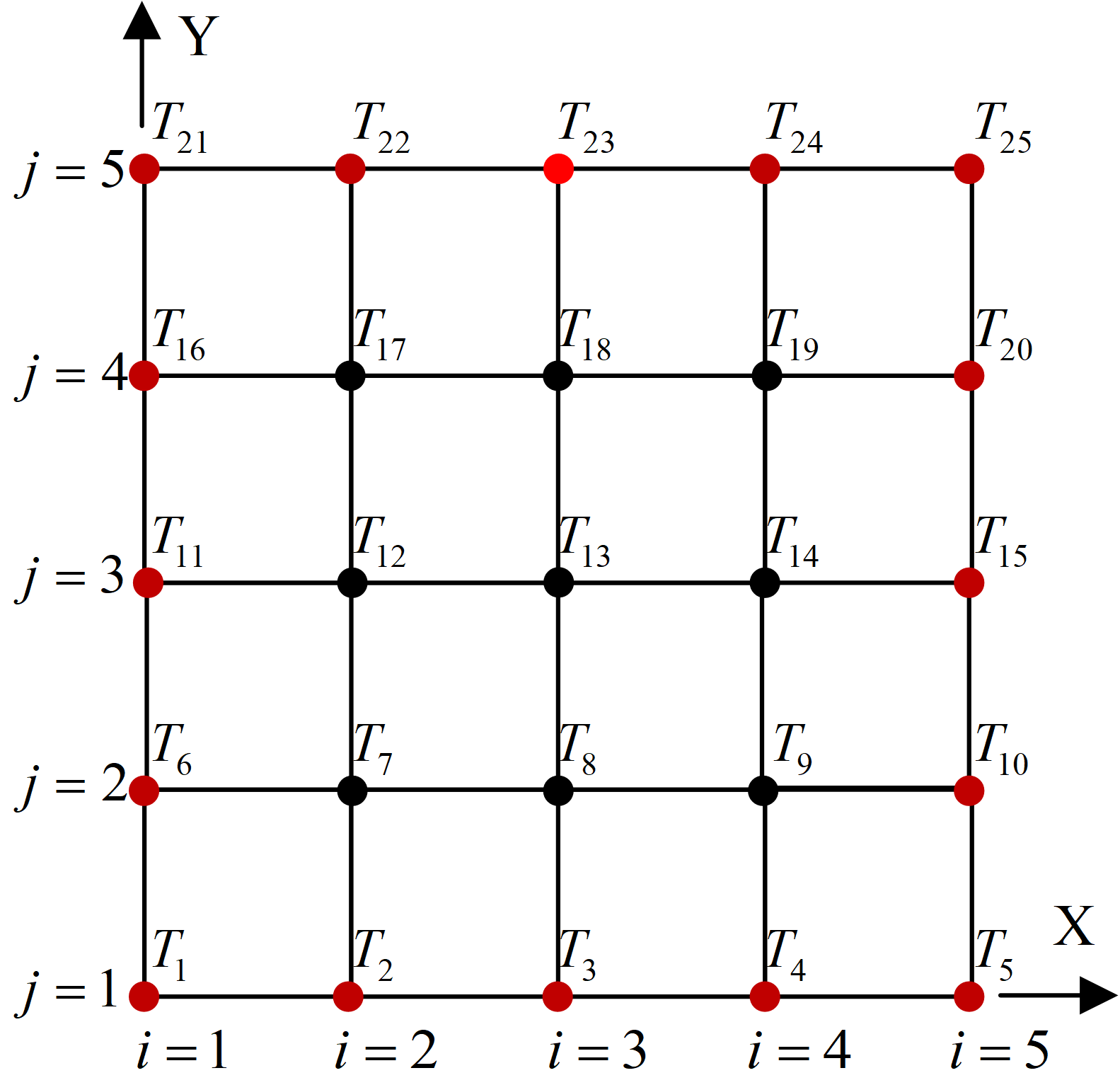}
		\label{fig:sort_mesh1}
	}
	\caption{The grid discretization of the two-dimensional domain.}
	\label{fig:mesh_sort}
\end{figure}
In this chapter, we propose a coefficient matrix condition number based position selection of observations (CMCN-PSO) method to alleviate the effect of noise observations. 
First, the finite difference method is utilized to discretize the governing equation, where the unknown intensity is set to be a variable. Then, combing with matrix forms about observations, the constrained optimization problem of TFI-HSS in Eq.(\ref{eq_optim}) is transformed into an unconstrained optimization problem by the penalty method. 
Furthermore, through analyses of noise observations, the upper bound of the reconstruction error is proved to be related to the coefficient matrix condition number, which is determined by positions of observations. 
After that, the condition number becomes the principle to evaluate positions of observations from many groups of positions. 
Many groups of positions are sampled by Latin hypercube, low discrepancy and grid-based sampling methods in this work. Finally, the optimal positions corresponding to the minimum condition number are selected. The CMCN-PSO method is described in Algorithm \ref{alg_CMCN-PSO} and the detail steps are as follows.

The two-dimensional domain is first discretized by using a Cartesian shown in Fig. \ref{fig:mesh_sort}. The grid points are divided into boundary points (red) and internal points (black). The temperatures of boundary points are controlled by boundary conditions. The governing equation in Eq.(\ref{eq_optim}) is discretized by the finite difference method, which is written as
\begin{equation}
\label{grid_point}
\begin{array}{c}
-4 t_{i, j}+t_{i-1, j}+t_{i+1, j}+t_{i, j-1}+t_{i, j+1}=h^{2} \phi, \\
\end{array}
\end{equation}
where $1 \leq i, j \leq K$ and $t_{j, k}$ denotes the temperature value of node $(i,j)$ and $h$ is the step length of two adjacent nodes. Assume that the two-dimensional domain is meshed as a $K \times K$ system and the temperature of the grid points is sorted as $T_{m \times 1}=\left(t_{1}, t_{2}, \cdots, t_{m}\right)^{T}, m=K \times K$. The governing equation and boundary conditions in Eq.(\ref{eq_optim}) are expressed as
\begin{equation}
\label{eq_line}
A_{1} \cdot T=h^{2} \phi+C_{1},
\end{equation}
where 
\begin{equation}
A_{1}=\left[\begin{array}{ccc}
\mathbf{V}+2 \mathbf{I} & -\mathbf{I} & 0 \\
-\mathbf{I} & \mathbf{V}+2 \mathbf{I} & -\mathbf{I} \\
0 & -\mathbf{I} & \mathbf{V}+2 \mathbf{I}
\end{array}\right]_{m \times m}
\end{equation}	
is the coefficient matrix and $\boldsymbol{V}=\boldsymbol{tridiag}_{L}(-1,2,-1) \in \mathbb{R}^{L, L}$. $C_{1}$ is a vector that contains boundary values. Given the HSS with $n$ heat sources, $\phi_{i}$ forms a matrix $Y_{n \times 1}=\left(\phi_{1}, \phi_{2}, \cdots, \phi_{n}\right)^{T}$. According to Eq.(\ref{eq3}), $\phi$ can also be expressed as
\begin{equation}
\begin{aligned}
\phi &=\left(\begin{array}{ccc}
b_{1,1} & \ldots & b_{1, n} \\
\vdots & \ddots & \vdots \\
b_{s, 1} & \cdots & b_{m, n}
\end{array}\right)_{\operatorname{m} \times n}\left(\begin{array}{l}
\phi_{1} \\
\phi_{2} \\
\vdots \\
\phi_{n}
\end{array}\right)_{n \times 1} \\
&=B_{m \times n} Y_{n \times 1},
\end{aligned}
\end{equation}
where $b_{i,j}$ is equal to 0 or 1. When the grid point of $t_{i}\left(1 \leq i \leq m\right)$ is located in the area $\Gamma_{j}\left(1 \leq j \leq n\right)$, $b_{i,j}$ is equal to 1 and otherwise $b_{i,j}$ is set to be 0. Then, Eq.(\ref{eq_line}) is simplified as
\begin{equation}
\label{eq_constraint}
A \hat{T}=\left(A_{1},-h^{2} B\right)\left(\begin{array}{l}
T \\
Y
\end{array}\right)=C_{1}.
\end{equation}

The constraint of the optimization problem in Eq.(\ref{eq_optim}) can be described by Eq.(\ref{eq_constraint}) and the optimization objective can be written as
\begin{equation}
\label{eq_obs}
\begin{aligned}
\min \left\|\hat{O} \hat{T}-C_{2}\right\|_{2}^{2} &=\min \left\|\left(\begin{array}{ll}
O & 0
\end{array}\right)\left(\begin{array}{l}
T \\
Y
\end{array}\right)-C_{2}\right\|_{2}^{2},
\end{aligned}
\end{equation}
where $O$ is a matrix describing positions of observations, and $C_{2}$ is a matrix containing $T_{obs}^{i}\left(1 \leq i \leq m\right)$. The matrix $O$ has only $1$ in each row and the rest are 0. For example, there are three observations in grid node 1, 2 and 5. $\hat{O} \hat{T}-C_{2}$ can be expressed as

\begin{equation}
\left(\begin{array}{ccccccc}
1 & 0 & 0 & 0 & 0 & \cdots & 0 \\
0 & 1 & 0 & 0 & 0 & \cdots & 0 \\
0 & 0 & 0 & 0 & 1 & \cdots & 0
\end{array}\right)_{3 \times m} \hat{T}_{m \times 1}-\left(\begin{array}{c}
T_{o b s}^{1} \\
T_{o b s}^{2} \\
0 \\
0 \\
T_{obs}^{3} \\
\vdots \\
0
\end{array}\right)_{m \times 1}.
\end{equation}
Therefore, the optimization problem of the TFI-HSS task is formulated by
\begin{equation}
\label{eq_penalty}
\begin{array}{l}
\min \left\|\hat{O} \hat{T}-C_{2}\right\|_{2}^{2}, \\
\text { s.t. } A \hat{T}=C_{1},
\end{array}
\end{equation}
which can be transformed into an unconstrained problem by the penalty method. It is written by a penalty weight parameter $\lambda^{2}$ as 
\begin{equation}
\min \left\|\hat{O} \hat{T}-C_{2}\right\|_{2}^{2}+\lambda^{2}\left\|A \hat{T}-C_{1}\right\|_{2}^{2},
\end{equation}
which is equivalent to
\begin{equation}
\label{un_opti_problem}
\begin{aligned}
\min \|\hat{A} \hat{T}-\hat{C}\|_{2}^{2} &=\min \left\|\begin{array}{c}
|\lambda|\left(A \hat{T}-C_{1}\right) \\
\hat{O} \hat{T}-C_{2}
\end{array}\right\|_{2}^{2}
\end{aligned}
\end{equation}
where $\hat{A}=(|\lambda| A, \hat{O})^{T}$, $\hat{T}=(T, Y)^{T}$ and $\hat{C}=\left(\lambda C_{1}, C_{2}\right)^{T}$.

As shown in Eq.(\ref{un_opti_problem}), the constrained optimization problem of the TFI-HSS task is transformed into an unconstrained optimization problem. 
In real-life applications, observations are inevitably perturbed by noises. For the unconstrained optimization problem in Eq.(\ref{un_opti_problem}), according to \textbf{Theorem 3.1}, we have proven that when observation $C$ has a perturbation $\delta C$, the relative error of the TFI-HSS task has a upper bound, which is related to the condition number of $\hat{A}$. For
\begin{equation}
\hat{A}=\left(\begin{array}{c}
\lambda A \\
\hat{O}
\end{array}\right),
\end{equation}
if boundary conditions and positions of components are confirmed, the matrix $A$ can be uniquely determined. The matrix $\hat{O}$ is determined by positions of observations. Choosing positions reasonably can find a matrix $\hat{A}$ with a smaller condition number, which results in a smaller relative error upper bound of the TFI-HSS task. 
It is more possible to obtain a robust temperature field by the PINN-TFI method with these positions under the same number of noise observations. 
Therefore, the condition number of $\hat{A}$ becomes the principle to evaluate positions of observations. 

In a summary, the CMCN-PSO method can be briefly described as follows. Assume that there are $L$ sets of positions $\{\left(x_{1}, y_{1}\right),\ldots$ $, \left(x_{m}, y_{m}\right)\}$, $\ldots$,  $\{(x_{L m+1}, y_{L m+1}),$ $\ldots,\left(x_{(L+1) m}, y_{(L+1) m}\right)\}$. They can be converted to matrix form ${\hat{O}_{1}, \cdots, \hat{O}_{L}}$. Then, according to Eq.(\ref{eq_constraint}), coefficient matrix $A$ can be calculated. After that, combing ${O_{1}, \cdots, O_{L}}$ and $A$, the condition number $\kappa_{i}, (i=1, \cdots, L)$ corresponding to different positions of observations $O_{i}$ can be obtained. The minimal condition number is selected as the optimal condition number $\kappa_{optimal}$. Finally, the optimal positions $\hat{O}_{optimal}$ corresponding to $\kappa_{optimal}$ can be obtained. The optimal positions of observations is used for the PINN-TFI method can reconstruct a more robust temperature field.

\begin{figure}[htbp]
	\centering
	\includegraphics[width=0.4\linewidth]{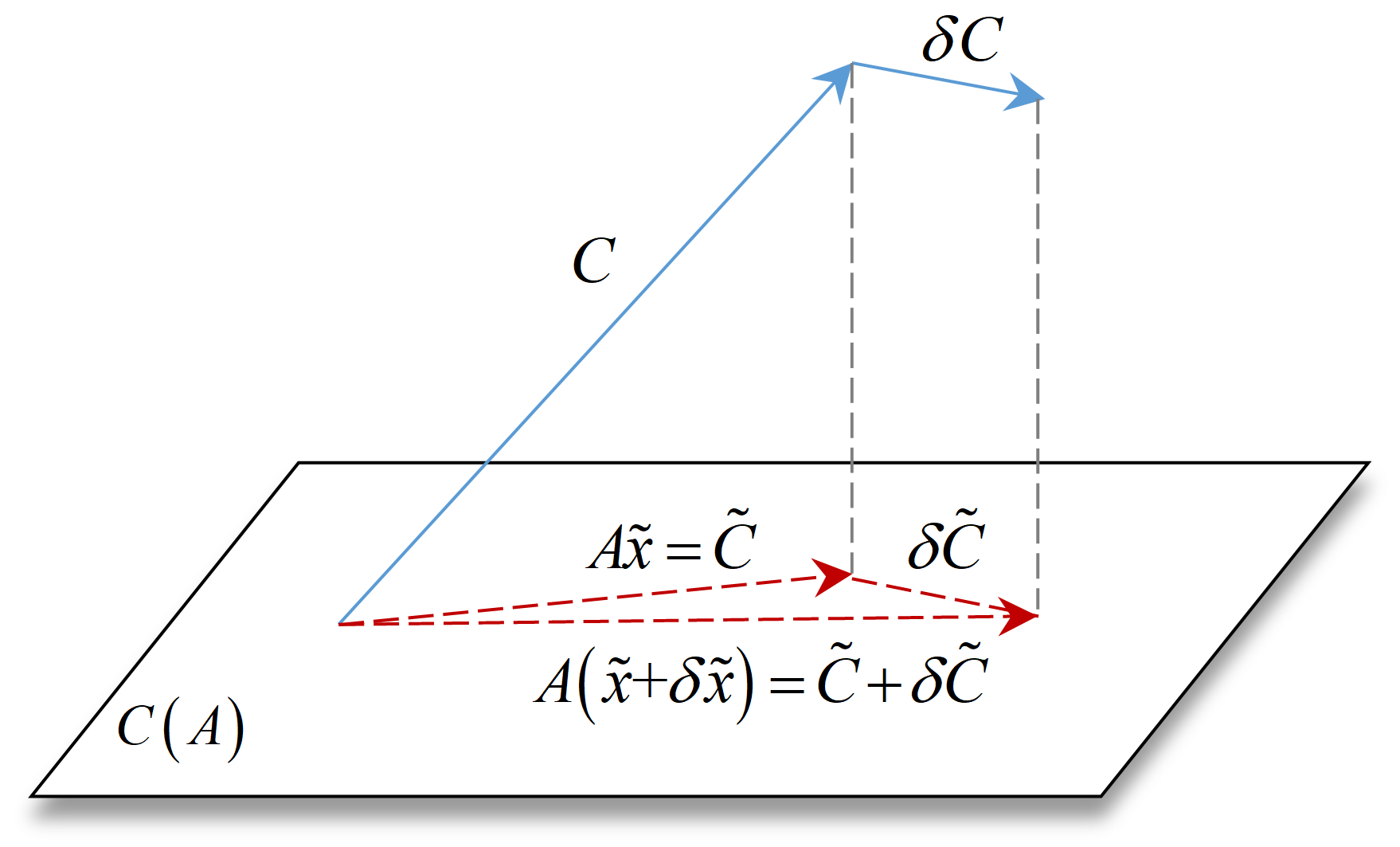}
	\caption{The diagram of the vectors.}
	\label{fig:prove}
\end{figure}
\noindent\textbf{Theorem 3.1} (Relative error upper bound). Suppose that A is m-by-n with $m>n$ and has full remark. Suppose that $x$ minimizes $\|Ax-C\|_{2}$ and $\tilde{x}$ is the solution. There is a perturbation $\delta C$ in $C$. Let $x+\delta x$ minimizes $\|A(x+\delta x)-(C+\delta C)\|_{2}$ and $\tilde{x}+\delta \tilde{x}$ is the solution. Then,
\begin{equation}
\frac{\|\delta \hat{x}\|_{2}}{\|\hat{x}\|_{2}} \leq \frac{\kappa(A)}{M} \frac{\|\delta C\|_{2}}{\|C\|_{2}},
\end{equation}
where $\kappa(A)=\|A\|_{2}\left\|\left(A^{H} A\right)^{-1} A^{H}\right\|_{2}$ and $M$ is a constant.

\noindent\textbf{Sketch of Proof of Theorem 3.1.}

The area $C(A)$ is shown in Fig. \ref{fig:prove}, which represents the column space of A. The vector $C$ typically will not be in $C(A)$. $\tilde{C}$ is the projection of $C$ onto $C(A)$ and $\tilde{x}$ takes the correct linear combinations of the matrix $A$ to obtain $\tilde{C}$, which can be written as 
\begin{equation}
\tilde{C}=A \tilde{x}.
\end{equation}
If there is a change $\delta C$ to $C$, there is a small change $\delta \tilde{C}$ in the projection of $C+\delta C$. Because projection is multiplication by the projection matrix, we can obtain
\begin{equation}
A(\tilde{x}+\delta \tilde{x})=\tilde{C}+\delta \tilde{C}.
\end{equation} 
For 
\begin{equation}
\cos \theta=\frac{\|\tilde{C}\|_{2}}{\|C\|_{2}},
\end{equation}
it can be written as
\begin{equation}
\label{costheta}
\cos \theta\|C\|_{2}=\|\tilde{C}\|_{2}=\|A \tilde{x}\|_{2} \leq\|A\|_{2}\|\tilde{x}\|_{2}.
\end{equation}
Because $\delta \tilde{x}$ is equivalent to $A^{\dagger}\delta \tilde{C}$, $\|\delta \tilde{x}\|_{2}$ can be computed by
\begin{equation}
\label{deltax}
\|\delta \tilde{x}\|_{2}=\left\|A^{\dagger} \delta \tilde{C}\right\| \leq\left\|A^{\dagger}\right\|_{2}\|\delta \tilde{C}\|_{2},
\end{equation}
where $A^{\dagger}$ represents $\left(A^{H} A\right)^{-1} A^{H}$. Combing Eq.(\ref{costheta}) and Eq.(\ref{deltax}), the relative error can be formulated by
\begin{equation}
\begin{aligned}
\frac{\|\delta \tilde{x}\|_{2}}{\|\tilde{x}\|_{2}} & \leq \frac{\|A\|_{2}\left\|A^{\dagger}\right\|_{2}}{\cos \theta} \frac{\|\delta \tilde{C}\|_{2}}{\|\delta C\|_{2}} \frac{\|\delta C\|_{2}}{\|\tilde{C}\|_{2}} \leq \kappa(A) \frac{\cos \tilde{\theta}}{\cos \theta} \frac{\|\delta C\|_{2}}{\|\tilde{C}\|_{2}} \\
& \leq \frac{\kappa(A)}{M} \frac{\|\delta C\|_{2}}{\|\tilde{C}\|_{2}},
\end{aligned}
\end{equation}
where $\kappa(A)$ is equivalent to $\|A\|_{2}\left\|\left(A^{H} A\right)^{-1} A^{H}\right\|_{2}$ and constant ($M>0$) is equivalent to $sup(cos\theta)$.

\section{Experiment}
\label{sec:4}
In this section, we mainly investigate the effectiveness performance of the PINN-TFI method and the CMCN-PSO method. 
Based on three typical cases, experiments are mainly to solve two tasks, namely TFI-HSS with noiseless observations and TFI-HSS with noise observations. (i) TFI-HSS with noiseless observations is designed to validate the effectiveness of the PINN-TFI method, where three sampling methods are used to select observations. (ii) For TFI-HSS with noise observations, the CMCN-PSO method is used to select positions of observations. Then, the PINN-TFI method is utilized to evaluate whether a more robust temperature field can be reconstructed with the optimal positions.
\subsection{Experiment setups}
\label{sec:4.1}
\textbf{Configurations of cases:} In this section, three typical cases are introduced to test performances of the PINN-TFI method and the CMCN-PSO method for the TFI-HSS task. The two-dimensional plane is modeled a square domain with the size $10cm\times10cm$. All cases with different boundaries are shown as follows.

\begin{figure}[htbp]
	\centering
	\subfigure[Case 1]{
		\centering
		\includegraphics[width=0.16\textheight]{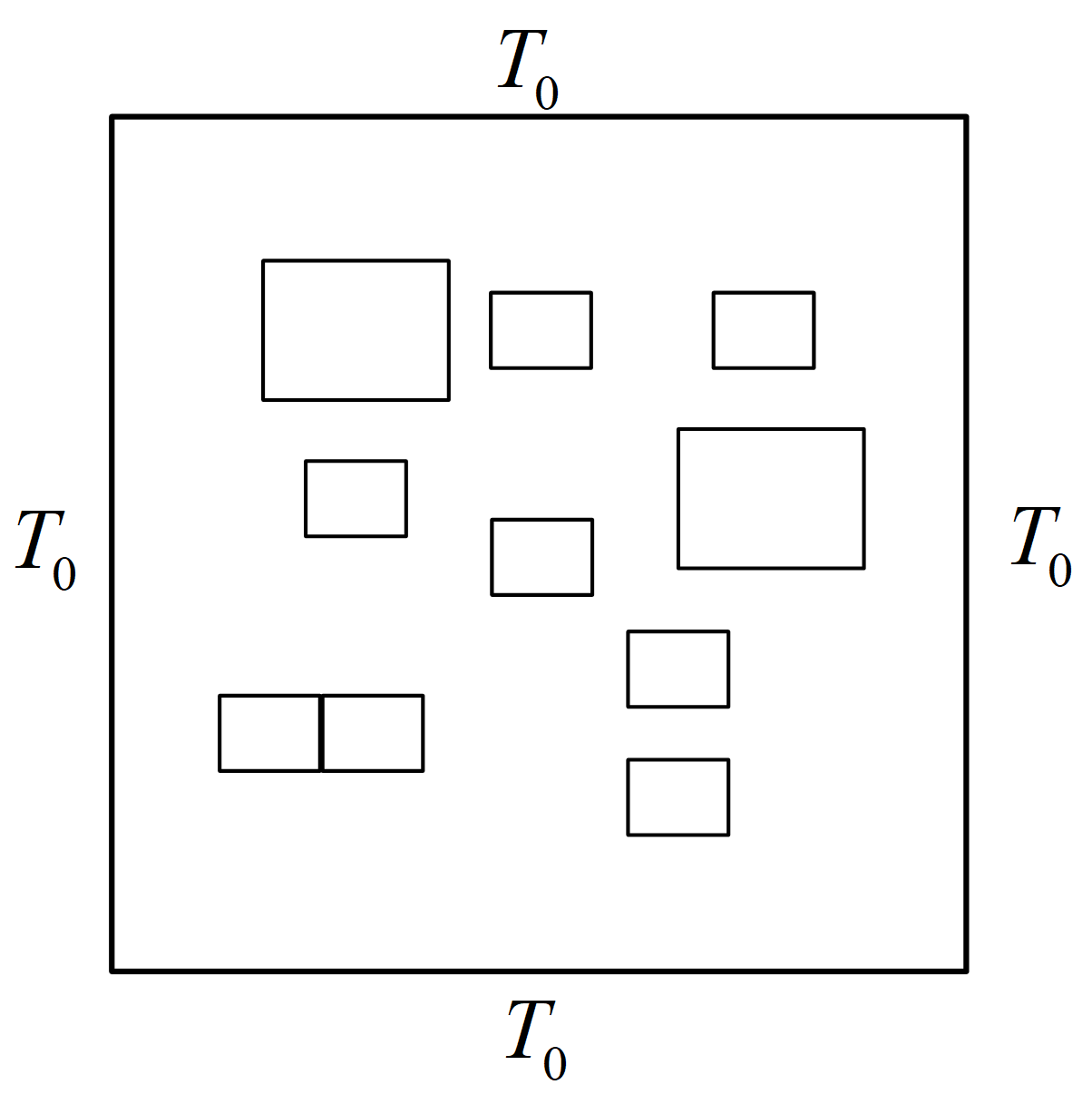}
		\label{fig:case1}
	}
	\subfigure[Case 2]{
		\centering
		\includegraphics[width=0.15\textheight]{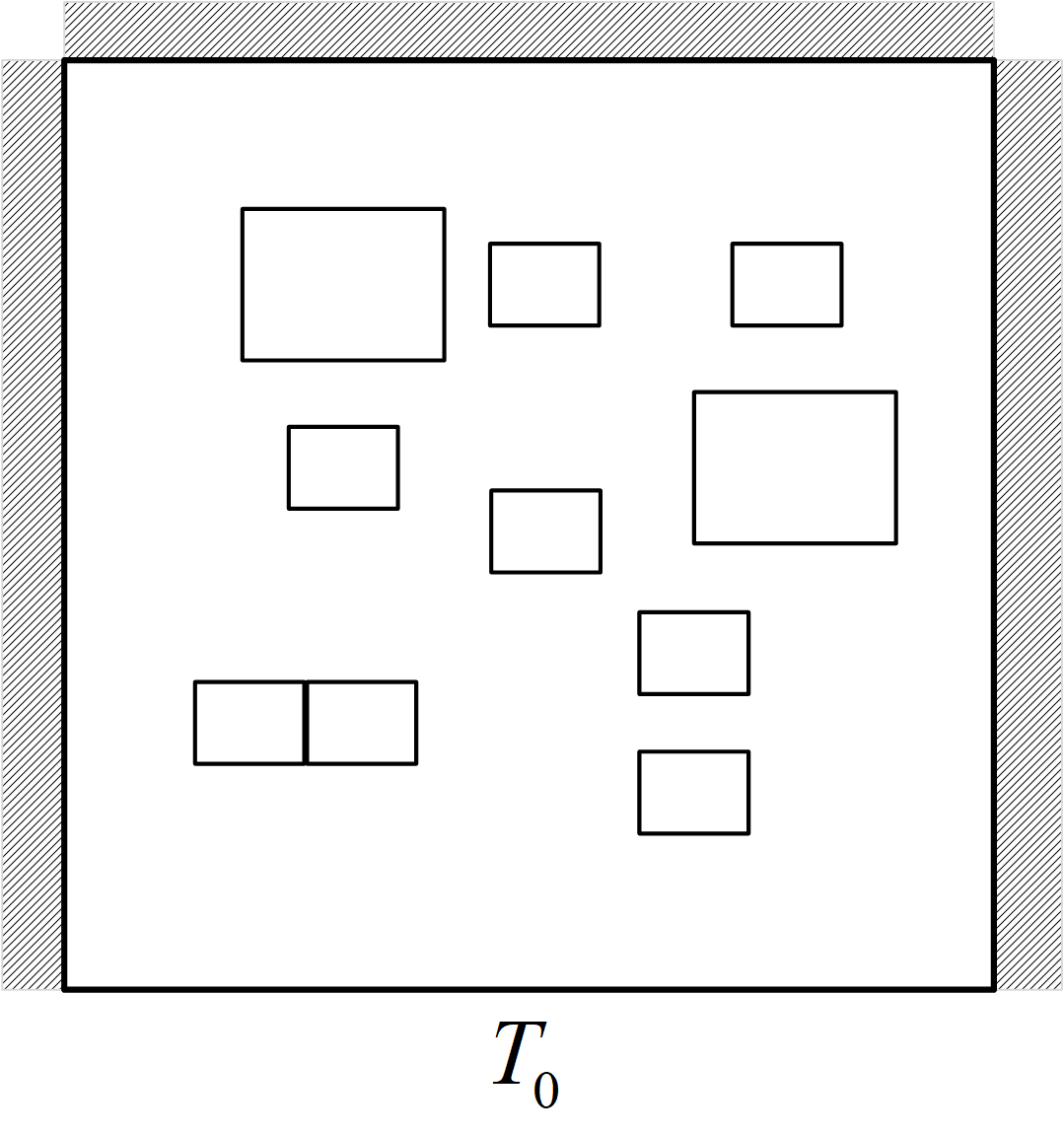}
		\label{fig:case2}
	}
	\subfigure[Case 3]{
		\centering
		\includegraphics[width=0.15\textheight]{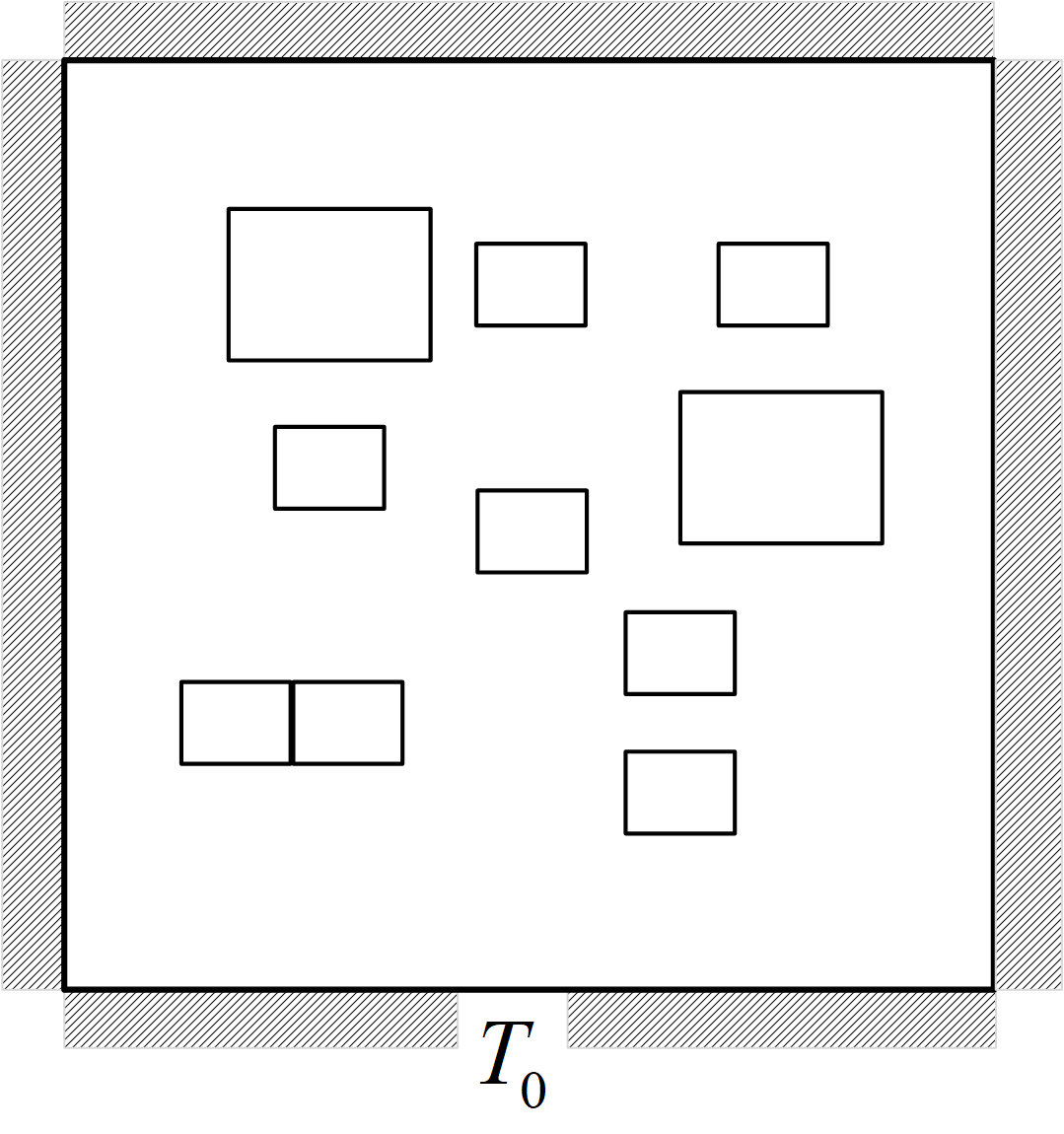}
		\label{fig:case3}
	}
	\caption{The illustration of VP and VB problem in a square domain.}
	\label{fig:case}
\end{figure}

For all the data (observations and exact temperature filed), FEniCS \citep{alnaes2015fenics} is used to generate thermal simulation results.
\begin{itemize}
	\item[$\bullet$] \textbf{Case 1}: VB problem with identical boundary conditions for four boundaries. All boundaries are isothermal with constant temperature valued 298K (see Fig. \ref{fig:case}(a) for better understanding). 
	\item[$\bullet$] \textbf{Case 2}: VB problem with different boundary conditions. The bottom boundary is isothermal with constant temperature valued 298K and other boundaries are adiabatic. Fig. \ref{fig:case}(b) describes this type of VB problem.
	\item[$\bullet$] \textbf{Case 3}: VP problem with heat sink of width $1cm$. A finite heat-generating volume cooled by a small patch of the common heat sink with the temperature of $298K$ located on the middle of the bottom boundary. The rest boundaries are adiabatic. Fig. \ref{fig:case}(c) shows this type of VP problem.
\end{itemize}

\textbf{Three sampling methods:} This work considers Latin hypercube sampling, low discrepancy sampling, and grid-based sampling to select observations. The main motivation is to generate sets of points, which can uniformly coverage of the space. 
\begin{itemize}
	\item[$\bullet$] Latin hypercube sampling (LHS) is to generate a quasi-random sampling distribution. Suppose $n$ samples are sampled from the two-dimensional space. Each dimension is divided into $n$ intervals that do not overlap each other, and each interval has the same probability (it usually considers a uniform distribution so that the interval length is the same). One point then is sampled from each interval in each dimension. Finally, we randomly extract the points sampled in the previous step from each dimension and compose them into a vector. 
	\item[$\bullet$] Low discrepancy sampling (LDS) is to generate the low discrepancy sequence. Assume that the point set $X$ is sampled from the two-dimensional domain. Given a set $\mathrm{P}=\left\{x_{1}, \ldots, x_{n}\right\}$, the discrepancy of $\mathrm{P}$ is defined as follows:
	\begin{equation}
	\mathcal{D}\left(\mathrm{P}\right)=\sup _{B \in J}\left|\frac{\mathcal{C}\left(B, \mathrm{P}\right)}{N}-\lambda(B)\right|,
	\end{equation}
	where $\lambda$ is the Lebesgue measurement of $B$, $A(B,P)$ is the number of points in P that falls into B, and J is the set of two-dimensional intervals. The purpose of the low-discrepancy sequence is to make the discrepancy of the resulting points as small as possible. The detailed method can refer to \citep{cervellera2014low}.
	
	\item[$\bullet$] Grid-based sampling (GS) is to select observations based on mesh points. The most used way is to place as many observations as possible at important positions (such as positions of components). Therefore, except for some pre-defined observations (positions of components and other positions placed by engineers), the whole domain is meshed as a $K \times K$ grid system, and observations are added through the grid to achieve the overall uniformity of observations. Concretely, one observation is placed at components(center positions of components in our work) and some observations are placed by engineers' experiences. Then, the two-dimensional domain is meshed as a $N \times N$ grid system to determine whether there are observations inside grid areas. If not, observations are placed inside the area (grid center position). In particular, when the pred-defined observations are located on the gridlines, the points are considered to belong to the upper-right grid.
\end{itemize}

\textbf{Evaluation Metrics:} This work uses four metrics to evaluate performances of the reconstructed temperature field as follows:
\begin{itemize}
	\item[$\bullet$]\textbf{Mean Absolute Error (MAE)}. MAE is the mean absolute error of the whole reconstructed temperature field, which can be written as
	\begin{equation}
	E_{MAE}=\frac{1}{N^{2}} \sum_{i=1}^{N} \sum_{j=1}^{N}\left|T_{\theta}\left(x_{i}, y_{j}\right)-T_{s}\left(x_{i}, y_{j}\right)\right|,
	\end{equation}
	where $T_{\theta}\left(x_{i}, y_{j}\right)$ and $T_{s}\left(x_{i}, y_{j}\right)$ represent the predicted temperature and the ground-truth temperature by numerical simulations, respectively.
	
	\item[$\bullet$]\textbf{Component-constrained MAE (CMAE)}. CMAE is defined to calculate the mean absolute error of the area that heat sources cover, which can be formulated as
	\begin{equation}
	E_{CMAE}=\frac{1}{\left|\Omega_{com}\right|} \sum_{\left(x_{i}, y_{j}\right) \in \Omega_{com}}\left|T_{\theta}\left(x_{i}, y_{j}\right)-T_{s}\left(x_{i}, y_{j}\right)\right|,
	\end{equation}
	where $\Omega_{com}$ represents the area of heat sources cover.
	
	\item[$\bullet$] \textbf{Boundary-constrained MAE (BMAE)}. BMAE is defined to calculate the mean absolute error of the boundary area, which can be calculated as
	\begin{equation}
	E_{B M A E}=\frac{1}{\left|\Omega_{b}\right|} \sum_{\left(x_{i}, y_{j}\right) \in \Omega_{b}}\left|T_{\theta}\left(x_{i}, y_{j}\right)-T_{s}\left(x_{i}, y_{j}\right)\right|,
	\end{equation}
	where $\Omega_{b}$ represents the boundary area.
	
	\item[$\bullet$]\textbf{Maximum of Component-constrained Absolute Error (M-CAE)}. M-CAE is defined to calculate the maximum error of the temperature field, which be computed as
	\begin{equation}
	E_{M-C A E}=\max _{\left(x_{i}, y_{j}\right) \in \Omega}\left|T_{\theta}\left(x_{i}, y_{j}\right)-T_{s}\left(x_{i}, y_{j}\right)\right|,
	\end{equation}
	where $\Omega_{com}$ represents the whole domain.
\end{itemize}

\textbf{Optimization and hyperparameters:} By default, Adam solver \citep{kingma2014adam} is used with an initial learning rate of $1e-3$. Four hidden layers with 50 neurons in each layer are used. The weights $w_{PDE}, w_{BC}$ and $w_{data}$ are set to $1$, $1$, and $1e4$, respectively. The training iteration is set to be $5,000$. 

\textbf{Compute Infrastructure: }
A very common machine with a 2.8-GHz Intel(R) Xeon(R) Gold 6242 CPU, 128-GB memory, and NVIDIA GeForce RTX 3090 GPU was used to test the performance of proposed methods.

All implementations for the TFI-HSS task are based on an open source deep learning framework Pytorch.

\subsection{TFI-HSS with noiseless observations}
To investigate the effectiveness of the PINN-TFI method, we first consider solving the TFI-HSS task with noiseless observations. Three sampling methods are used to generate positions of observations. Following we mainly present the performance of the PINN-TFI method with the different number of observations, the performance of the PINN-TFI method with and without transfer strategy, the performance of the PINN-TFI method with different width and depth of the NN, the performance of the PINN-TFI method with different weights, and the comparisons with other methods.

\subsubsection{Performance with the different number of observations}
\begin{figure*}[!htbp]
	\centering
	\subfigure[layout]{
		\includegraphics[width=0.19\linewidth]{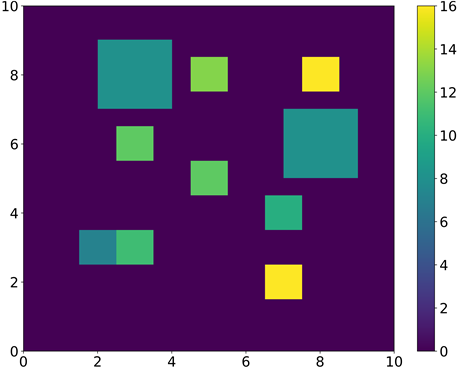}
	}
	\subfigure[case 1]{
		\includegraphics[width=0.2\linewidth]{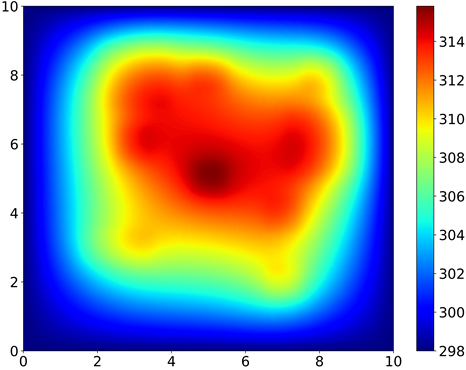}
	}
	\subfigure[case 2]{
		\includegraphics[width=0.2\linewidth]{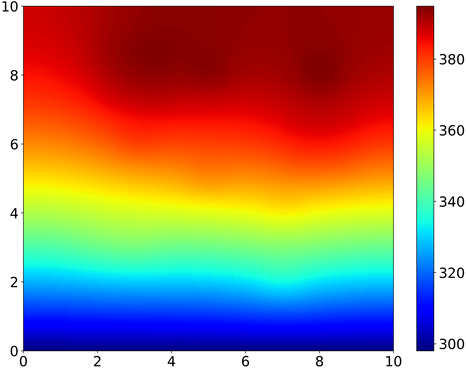}
	}
	\subfigure[case 3]{
		\includegraphics[width=0.2\linewidth]{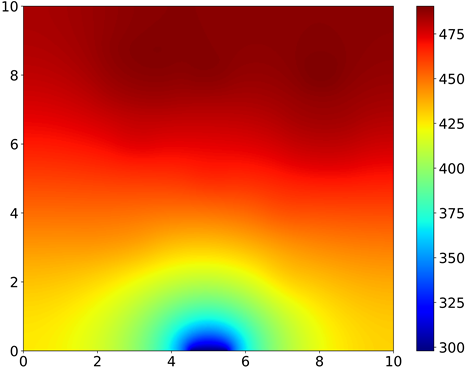}
	}
	\caption{The component layout and true temperature fileds for three cases.}
	\label{fig:true}
\end{figure*}

In this part, we conduct experiments under different numbers of observations from Latin hypercube sampling (LHS), low discrepancy sampling (LDS) and grid-based sampling (GS). The number of observations is chosen from $\{42, 68, 104, 148\}$. Fig. \ref{fig:true} shows positions and intensities of components as well as true temperature fields for three cases.

\begin{table*}[!bp]
	\caption{Performance of the PINN-TFI method with different numbers of observations from Latin hypercube sampling (LHS), low discrepancy sampling (LDS) and grid-based sampling (GS). The best results under different numbers of observations are highlight.}
	\label{tab:1}
	\centering
	\scalebox{0.78}{
		\begin{tabular}{cccccclcccclcccc}
			\hline
			\multirow{2}{*}{Num} & \multirow{2}{*}{Position} & \multicolumn{4}{c}{Case 1}                                            &           & \multicolumn{4}{c}{Case 2}                                            &  & \multicolumn{4}{c}{Case 3}                                            \\ \cline{3-6} \cline{8-11} \cline{13-16} 
			&                           & MAE             & CMAE            & BMAE            & M-CAE           &           & MAE             & CMAE            & BMAE            & M-CAE           &  & MAE             & CMAE            & BMAE            & M-CAE           \\ \hline
			& LHS                       & 0.3515          & 0.4588          & 0.1545          & 2.1604          &           & 0.7120          & 0.4889          & 0.8295          & 2.7160          &  & 0.8488          & 0.5006          & 2.0459          & 2.2240          \\
			42                   & LDS                       & \textbf{0.3058} & \textbf{0.2804} & \textbf{0.1249} & \textbf{1.7357} & \textbf{} & \textbf{0.5353} & \textbf{0.3821} & 0.5939          & 2.2917          &  & 0.7587          & 0.4689          & 1.2950          & 1.8336          \\
			& GS                        & 0.4396          & 0.3287          & 0.2531          & 2.1168          &           & 0.6817          & 0.3958          & \textbf{0.5664} & \textbf{1.6842} &  & \textbf{0.6488} & \textbf{0.3809} & \textbf{1.3150} & \textbf{1.7395} \\ \hline
			& LHS                       & 0.1946          & 0.1345          & 0.1272          & 0.9913          &           & 0.2956          & 0.3047          & 0.6487          & 1.0054          &  & \textbf{0.3794} & 0.3504          & \textbf{0.9033} & 1.5122          \\
			68                   & LDS                       & 0.1938          & 0.1446          & \textbf{0.1245} & 0.8330          &           & 0.2859          & 0.2564          & \textbf{0.5440} & \textbf{0.8572} &  & 0.4078          & 0.3471          & 1.2371          & 1.4958          \\
			& GS                      & \textbf{0.1816} & \textbf{0.1109} & 0.1892          & \textbf{0.7215} &           & \textbf{0.2472} & 0.1913          & 0.5506          & 0.9930          &  & 0.4262          & \textbf{0.2573} & 1.1508          & \textbf{1.4526} \\ \hline
			& LHS                       & 0.1789          & 0.1976          & 0.1097          & 1.2450          &           & 0.2201          & 0.1760          & 0.3241          & 0.9592          &  & 0.4490          & 0.4701          & 1.6813          & 1.3689          \\
			104                  & LDS                       & \textbf{0.1127} & \textbf{0.1178} & \textbf{0.1250} & \textbf{0.6170} &           & \textbf{0.1850} & 0.2151          & \textbf{0.2111} & 1.1666          &  & 0.2927          & 0.2825          & \textbf{0.7925} & 1.4032          \\
			& GS                        & 0.1785          & 0.1214          & 0.1961          & 0.6443          &           & 0.2163          & \textbf{0.1675} & 0.4997          & \textbf{0.8172} &  & \textbf{0.2833} & \textbf{0.2230} & 0.9241          & \textbf{0.7633} \\ \hline
			& LHS                     & 0.1157          & 0.0934          & 0.1207          & 0.5232          &           & 0.2309          & 0.1449          & 0.4662          & 0.9938          &  & 0.3823          & 0.3942          & 1.2509          & 1.7600          \\
			148                  & LDS                       & \textbf{0.0904} & 0.1138          & \textbf{0.0924} & 0.7660          &           & 0.1275          & 0.1652          & \textbf{0.1545} & \textbf{0.6356} &  & 0.2562          & 0.3160          & \textbf{0.6972} & \textbf{1.0443} \\
			& GS                        & 0.0918          & \textbf{0.0693} & 0.1896          & \textbf{0.3086} &           & \textbf{0.1070} & \textbf{0.1095} & 0.2512          & 0.6675          &  & \textbf{0.2115} & \textbf{0.2750} & 0.8019          & 1.1034          \\ \hline
		\end{tabular}
	}
\end{table*}

\begin{figure*}[!htbp]
	\centering
	\subfigure[Case 1 with LHS (125)]{
		\includegraphics[width=0.31\linewidth]{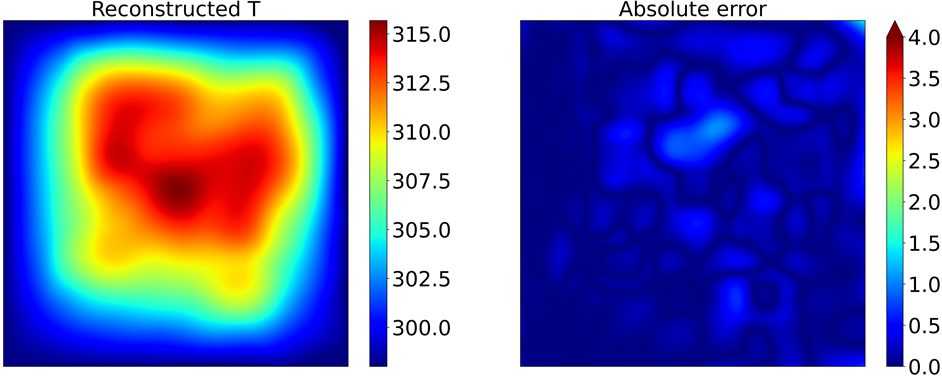}
	}
	\subfigure[Case 1 with LDS (125)]{
		\includegraphics[width=0.31\linewidth]{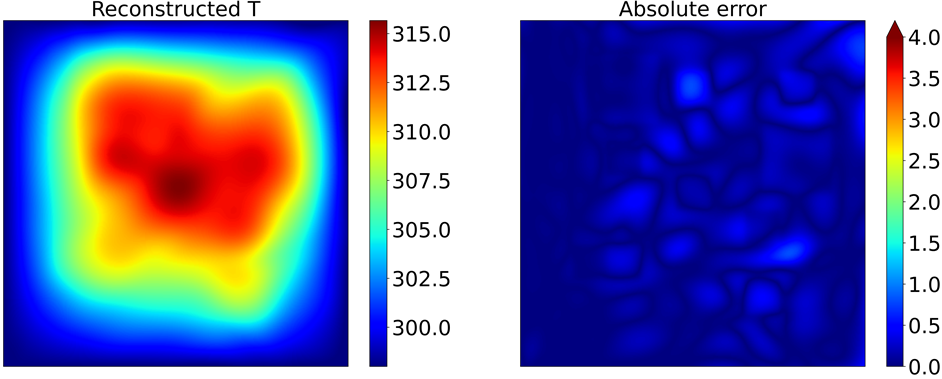}
	}
	\subfigure[Case 1 with GS (125)]{
		\includegraphics[width=0.31\linewidth]{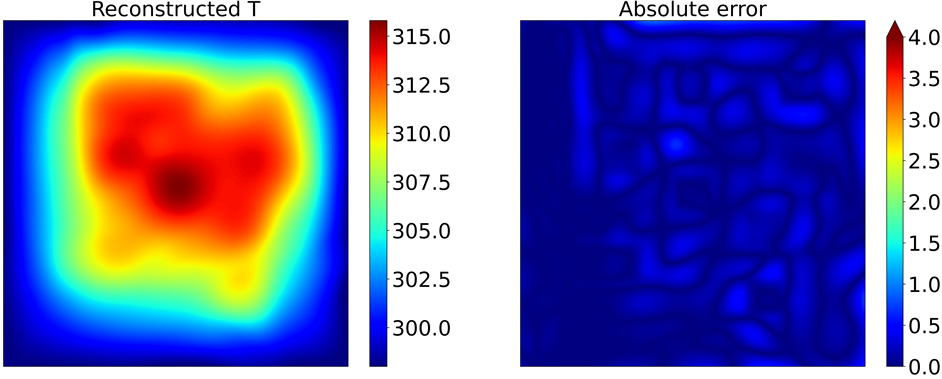}
	}
	\quad
	\subfigure[Case 2 with LHS (125)]{
		\includegraphics[width=0.31\linewidth]{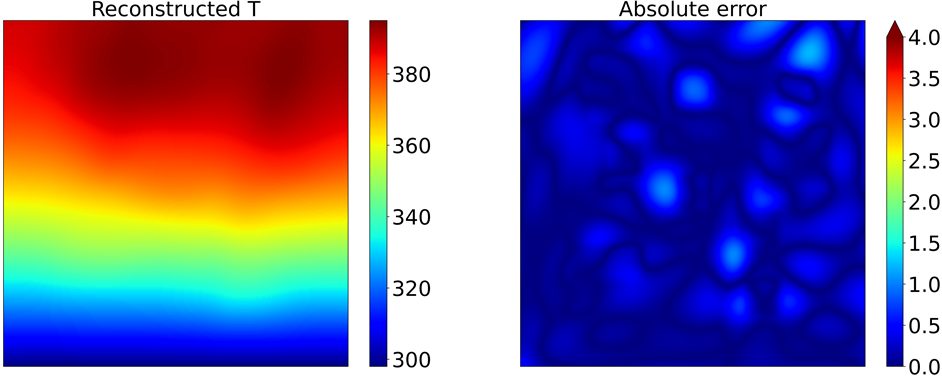}
	}
	\subfigure[Case 2 with LDS (125)]{
		\includegraphics[width=0.31\linewidth]{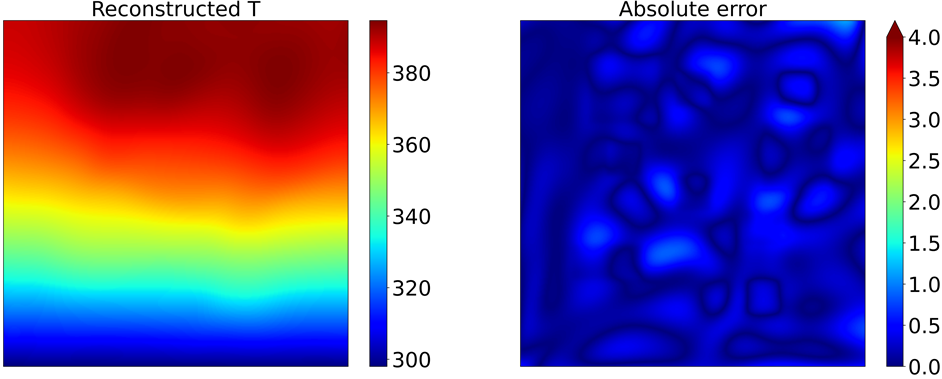}
	}
	\subfigure[Case 2 with GS (125)]{
		\includegraphics[width=0.31\linewidth]{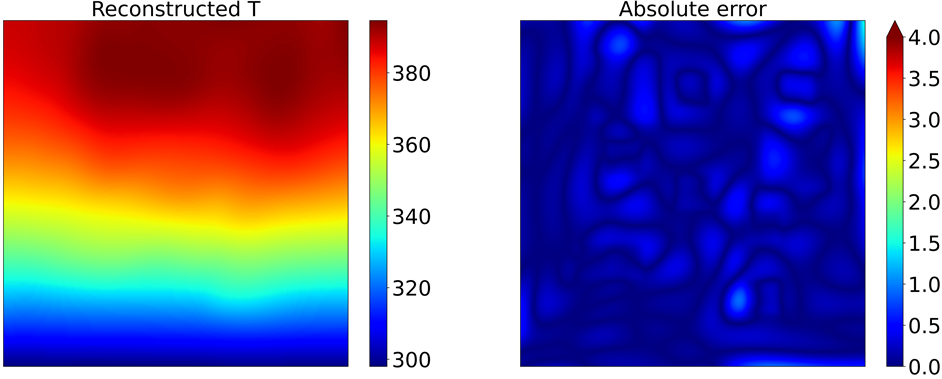}
	}
	\quad
	\subfigure[Case 3 with LHS (125)]{
		\includegraphics[width=0.31\linewidth]{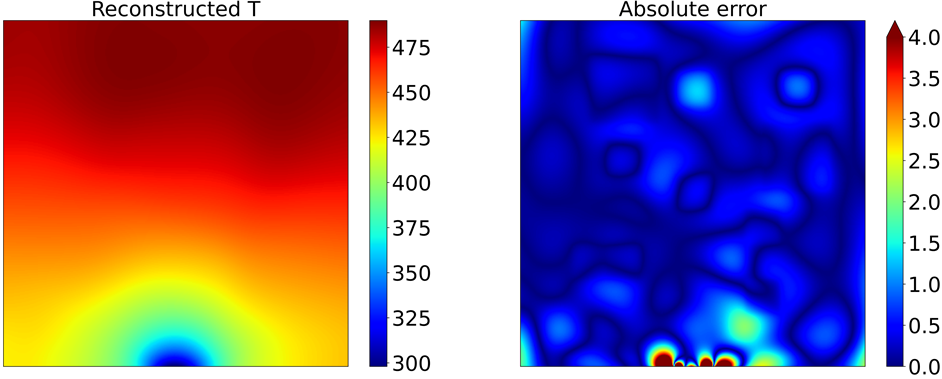}
	}
	\subfigure[Case 3 with LDS (125)]{
		\includegraphics[width=0.31\linewidth]{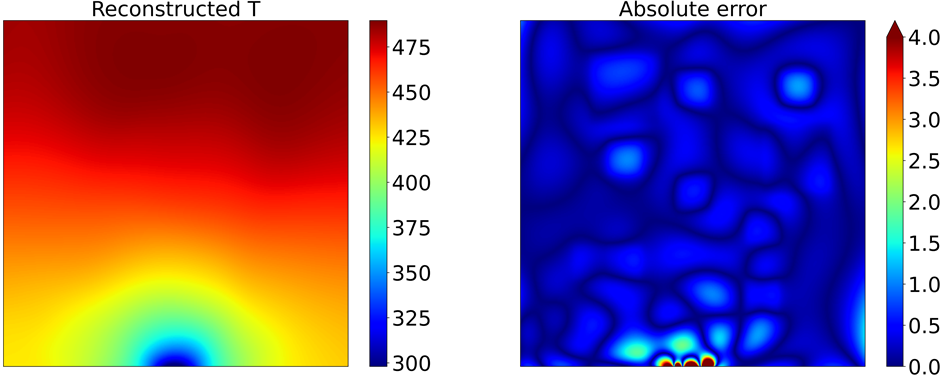}
	}
	\subfigure[Case 3 with GS (125)]{
		\includegraphics[width=0.31\linewidth]{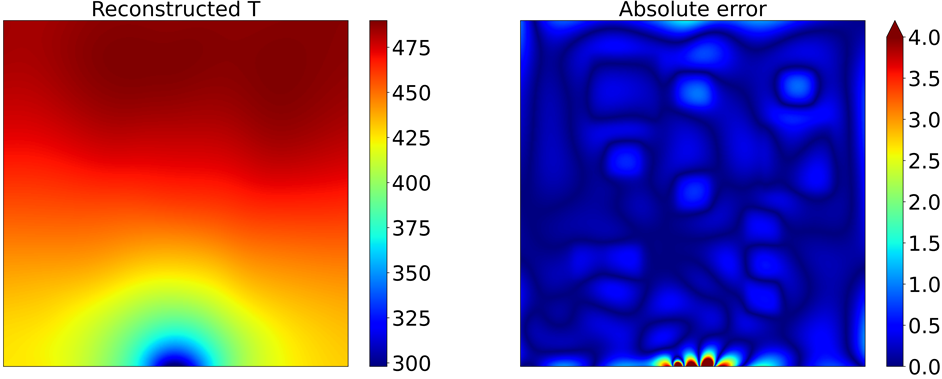}
	}
	\caption{Examples of the PINN-TFI method with 125 observations for TFI-HSS task.}
	\label{fig:TFI}
\end{figure*}

Table \ref{tab:1} lists the performance of the PINN-TFI method with 42, 68, 104, and 148 observations, where LHS, LDS, and GS are used to generate positions of observations. MAEs over three cases are decreasing with the increase of observations and are less than 0.5K under just 42 observations. CMAEs and the BMAEs are also less than 0.6K under just 42 observations. The prediction accuracy is acceptable even when the number of observations is extremely small. This also indicates the PINN-TFI method can work well in a small data setting. Except for case 3 with 68 observations, LDS and GS can find better positions for the PINN-TFI method to achieve smaller errors than LHS under the same number of observations. For case 1, the performance of the PINN-TFI method with 42 and 148 observations from LDS is better than that from GS. Fig. \ref{fig:TFI} shows the examples of the PINN-TFI method with 125 observations from LHS, LDS, and GS, respectively.

\begin{figure*}[!htbp]
	\centering
	\subfigure[MAE in case 1]{
		\includegraphics[width=0.2\linewidth]{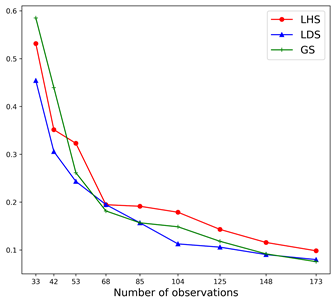}
	}
	\subfigure[CMAE in case 1]{
		\includegraphics[width=0.2\linewidth]{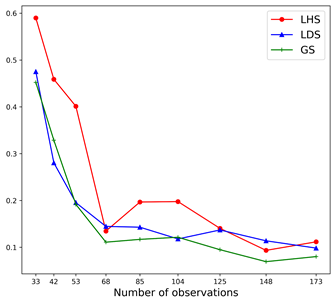}
	}
	\subfigure[BMAE in case 1]{
		\includegraphics[width=0.21\linewidth]{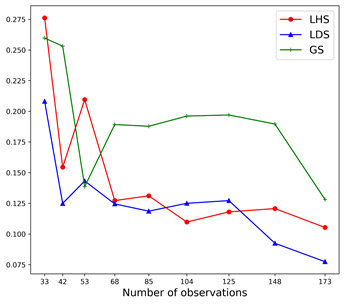}
	}
	\subfigure[M-CAE in case 1]{
		\includegraphics[width=0.2\linewidth]{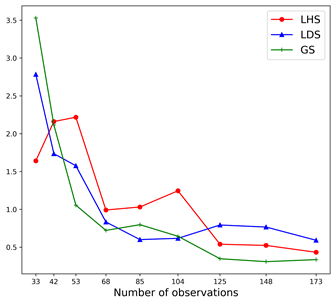}
	}
	\quad
	\subfigure[MAE in case 2]{
		\includegraphics[width=0.2\linewidth]{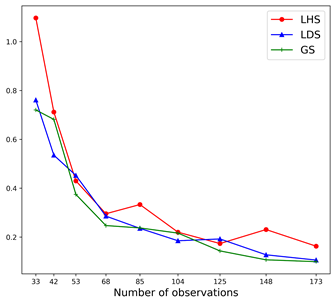}
	}
	\subfigure[CMAE in case 2]{
		\includegraphics[width=0.2\linewidth]{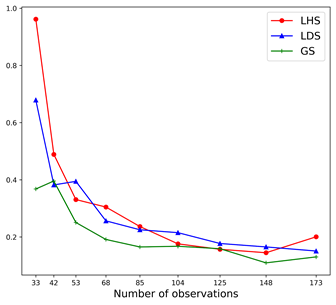}
	}
	\subfigure[BMAE in case 2]{
		\includegraphics[width=0.2\linewidth]{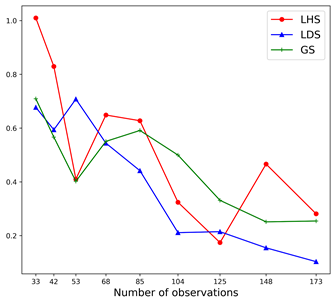}
	}
	\subfigure[M-CAE in case 2]{
		\includegraphics[width=0.2\linewidth]{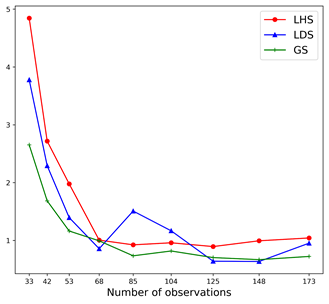}
	}
	\quad
	\subfigure[MAE in case 3]{
		\includegraphics[width=0.2\linewidth]{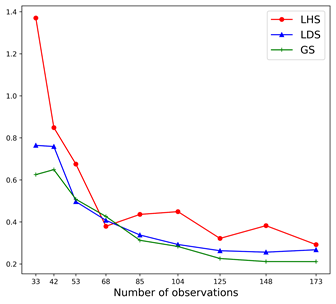}
	}
	\subfigure[CMAE in case 3]{
		\includegraphics[width=0.2\linewidth]{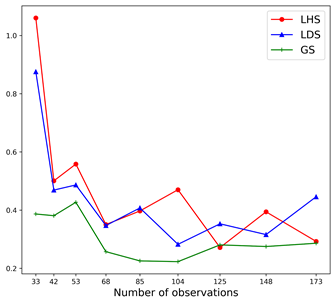}
	}
	\subfigure[BMAE in case 3]{
		\includegraphics[width=0.2\linewidth]{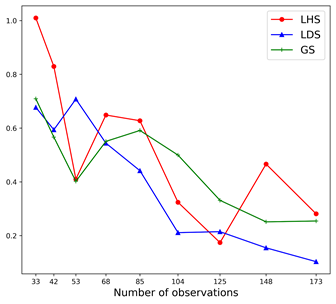}
	}
	\subfigure[M-CAE in case 3]{
		\includegraphics[width=0.2\linewidth]{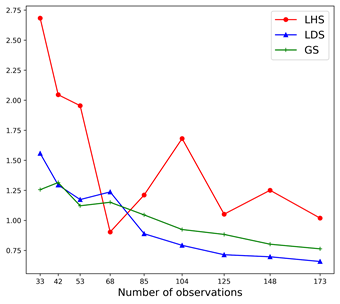}
	}
	\caption{Performance of the PINN-TFI method with different number of observations.}
	\label{fig:example}
\end{figure*}

It is more intuitive to get the above conclusion from Fig. \ref{fig:example}. As the figure shows, LDS and GS are better than LHS. The trends of MAEs of the PINN-TFI method with observations from LDS and GS are close. The MAEs of LDS and GS are decreasing with the increase of observations except for case 2 under 125 observations. Obviously, when the number of observations differs greatly, the performance of the PINN-TFI method with more observations is better than fewer ones whatever observations are from LHS, LDS, or GS. But when the number of observations is close, the performance of the PINN-TFI method with more observations from LHS might not better than fewer ones. It also can be seen that when the number of observations increases to some level, the performance improves slightly.

\subsubsection{Performance with and without transfer learning strategy}
To use transfer learning strategy for the acceleration, the PINN-TFI method designs the model initialization part, where observations are not required. Transfer learning strategy is used to initialize the NN model. In addition, Xavier initialization is a widespread initialization method used by many works \citep{raissi2019physics,raissi2020hidden}. To show the acceleration effect, this work compares the performance of the PINN-TFI method with transfer learning strategy and Xavier initialization.
\begin{figure*}[!bp]
	\centering
	\subfigure[Case 1 with LHS]{
		\includegraphics[width=0.31\linewidth]{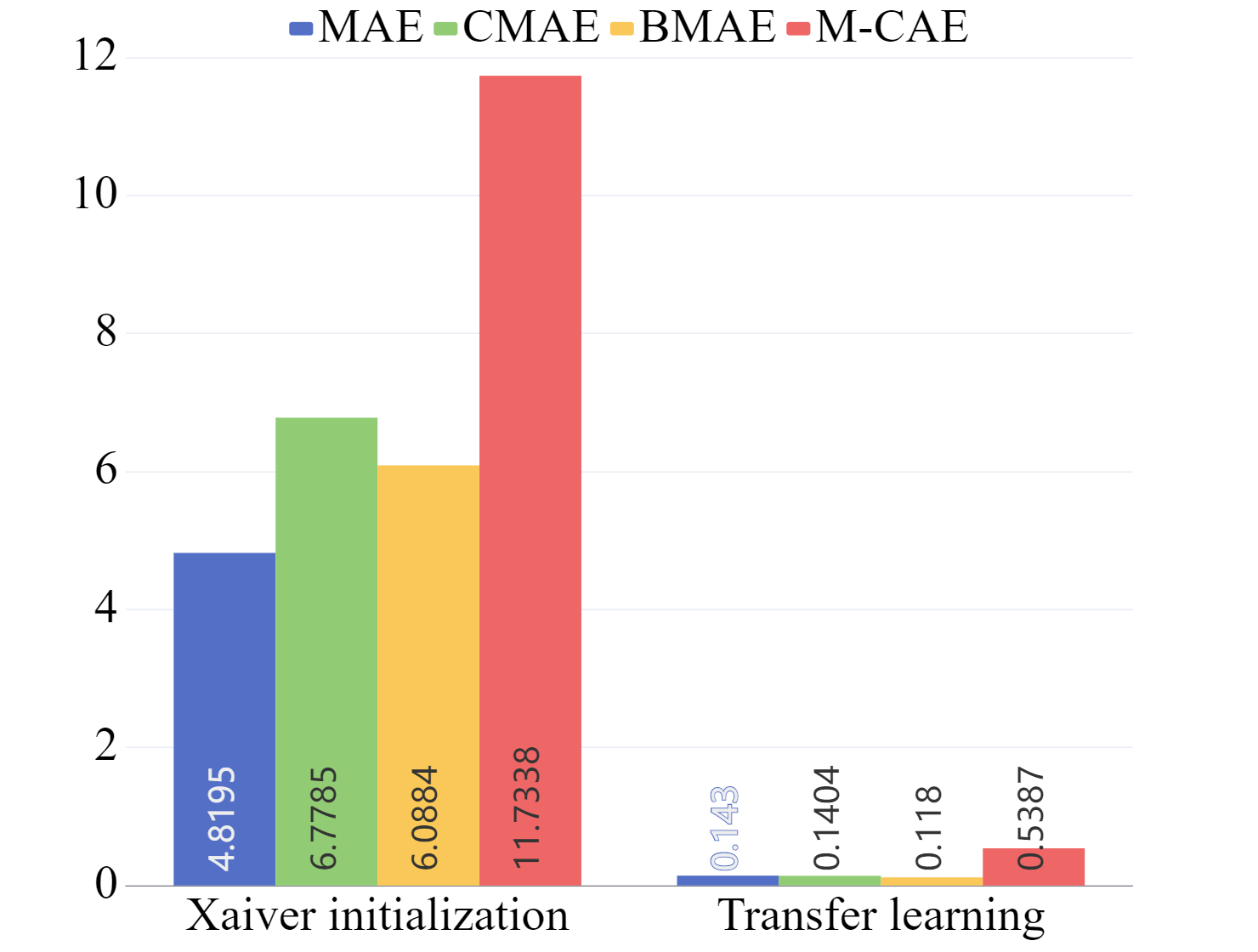}
	}
	\subfigure[Case 1 with LDS]{
		\includegraphics[width=0.31\linewidth]{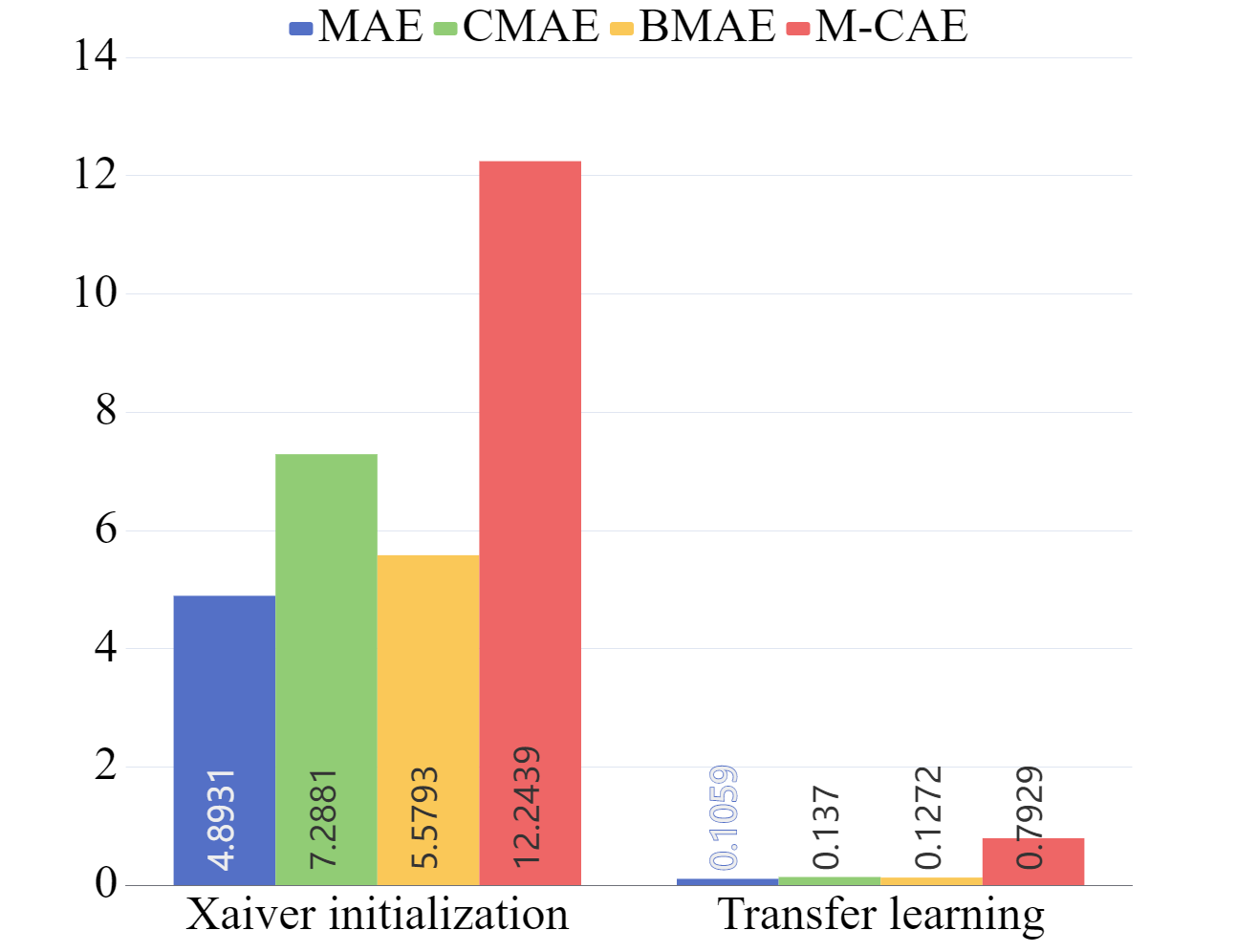}
	}
	\subfigure[Case 1 with GS]{
		\includegraphics[width=0.31\linewidth]{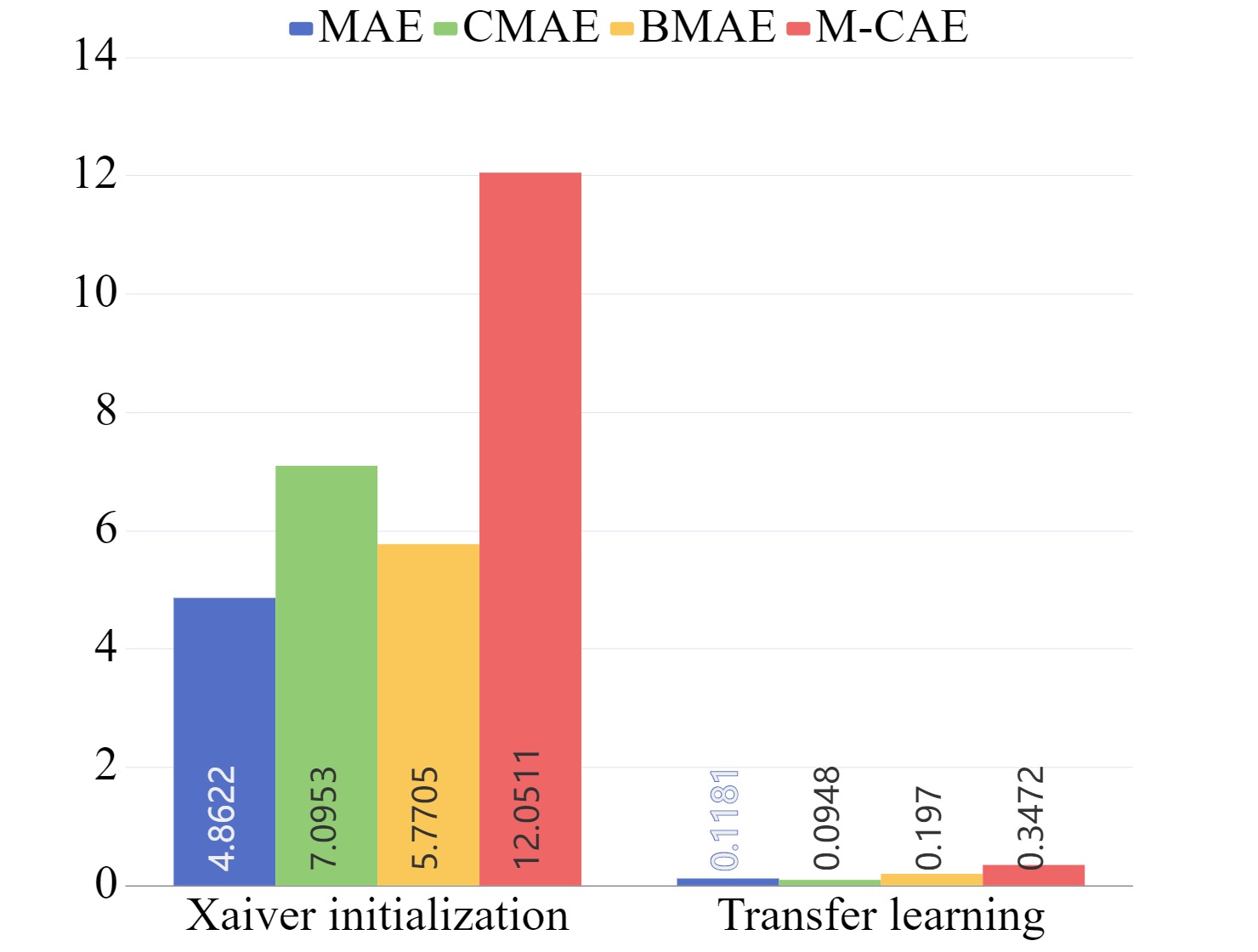}
	}
	\quad
	\subfigure[Case 2 with LHS]{
		\includegraphics[width=0.31\linewidth]{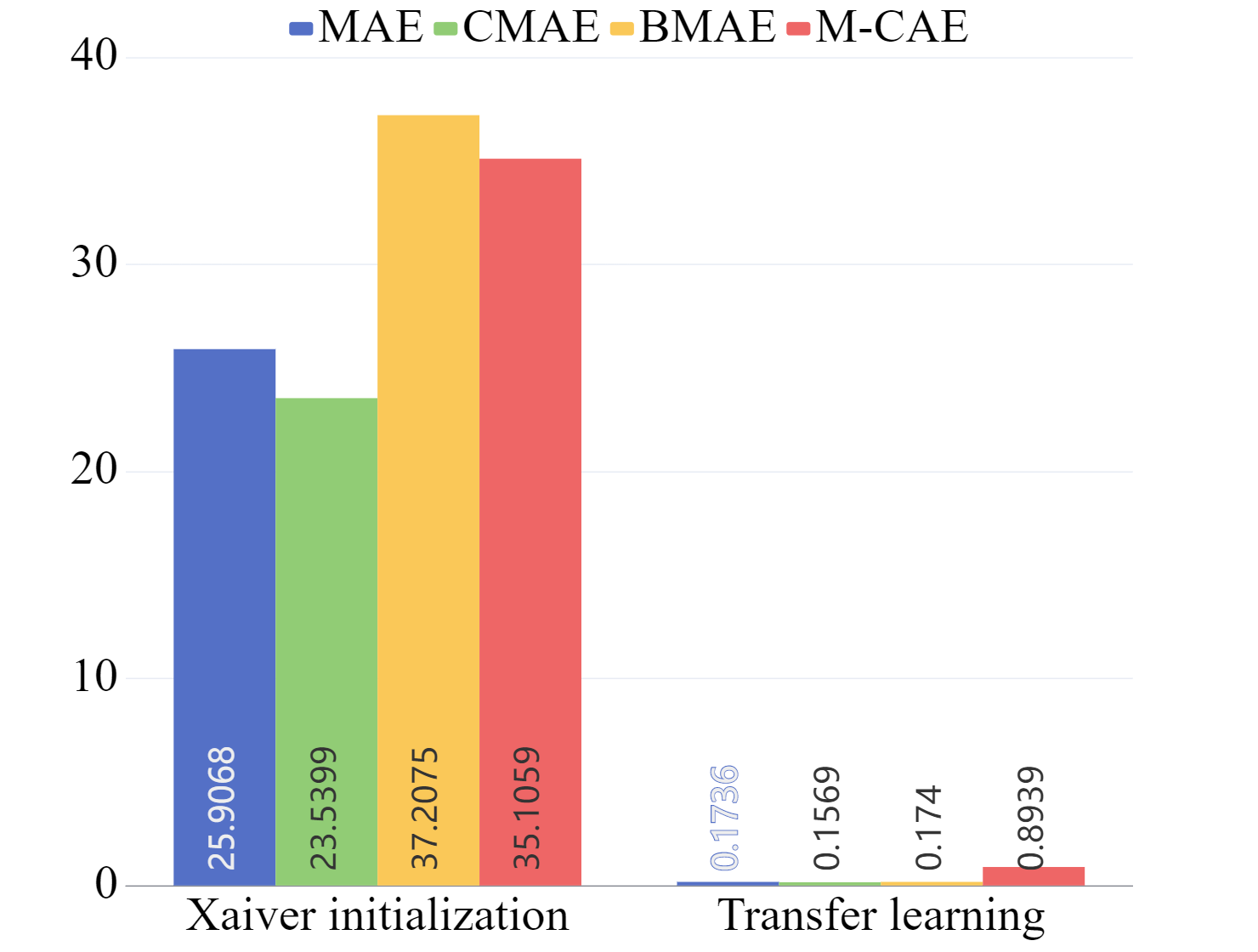}
	}
	\subfigure[Case 2 with LDS]{
		\includegraphics[width=0.31\linewidth]{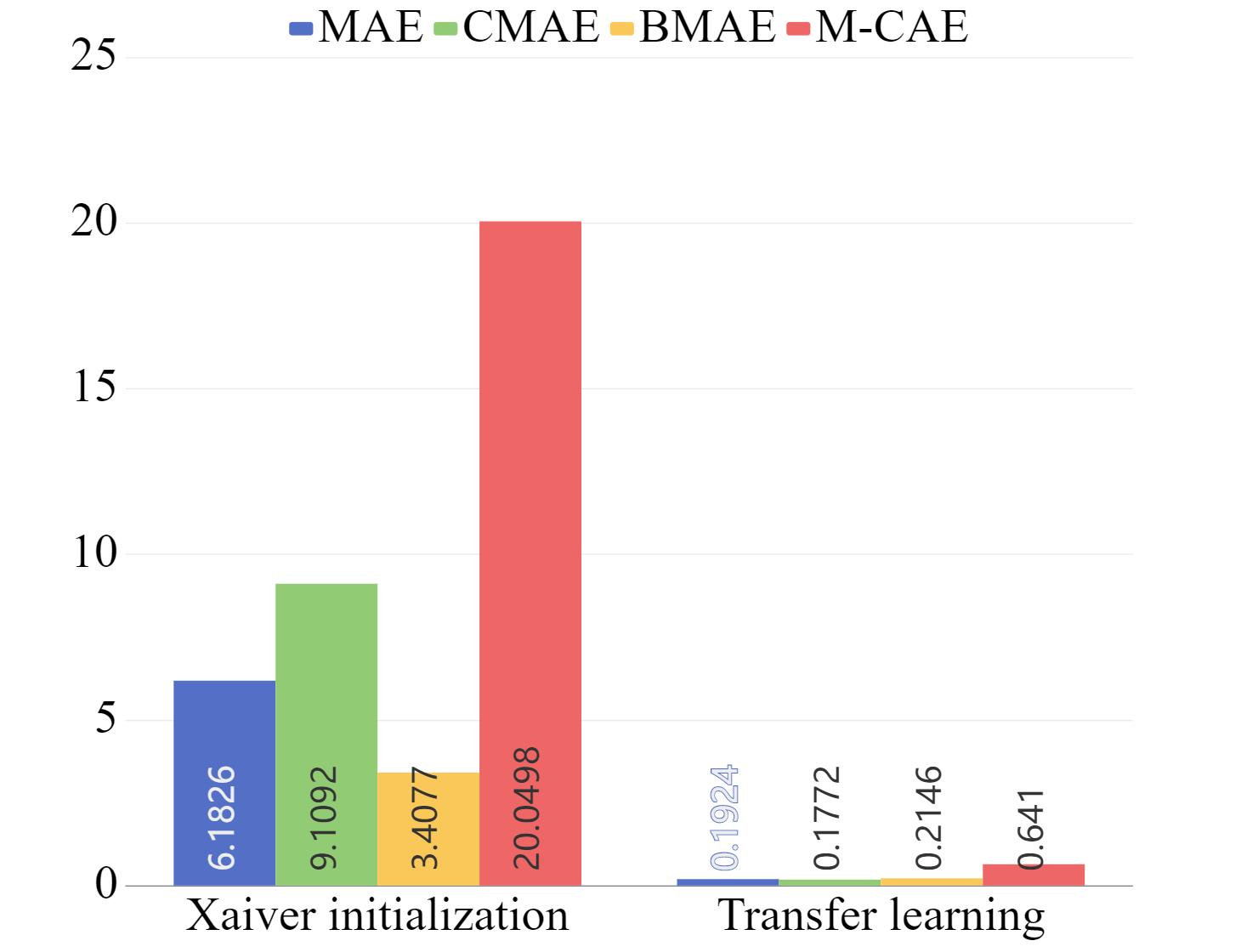}
	}
	\subfigure[Case 2 with GS]{
		\includegraphics[width=0.31\linewidth]{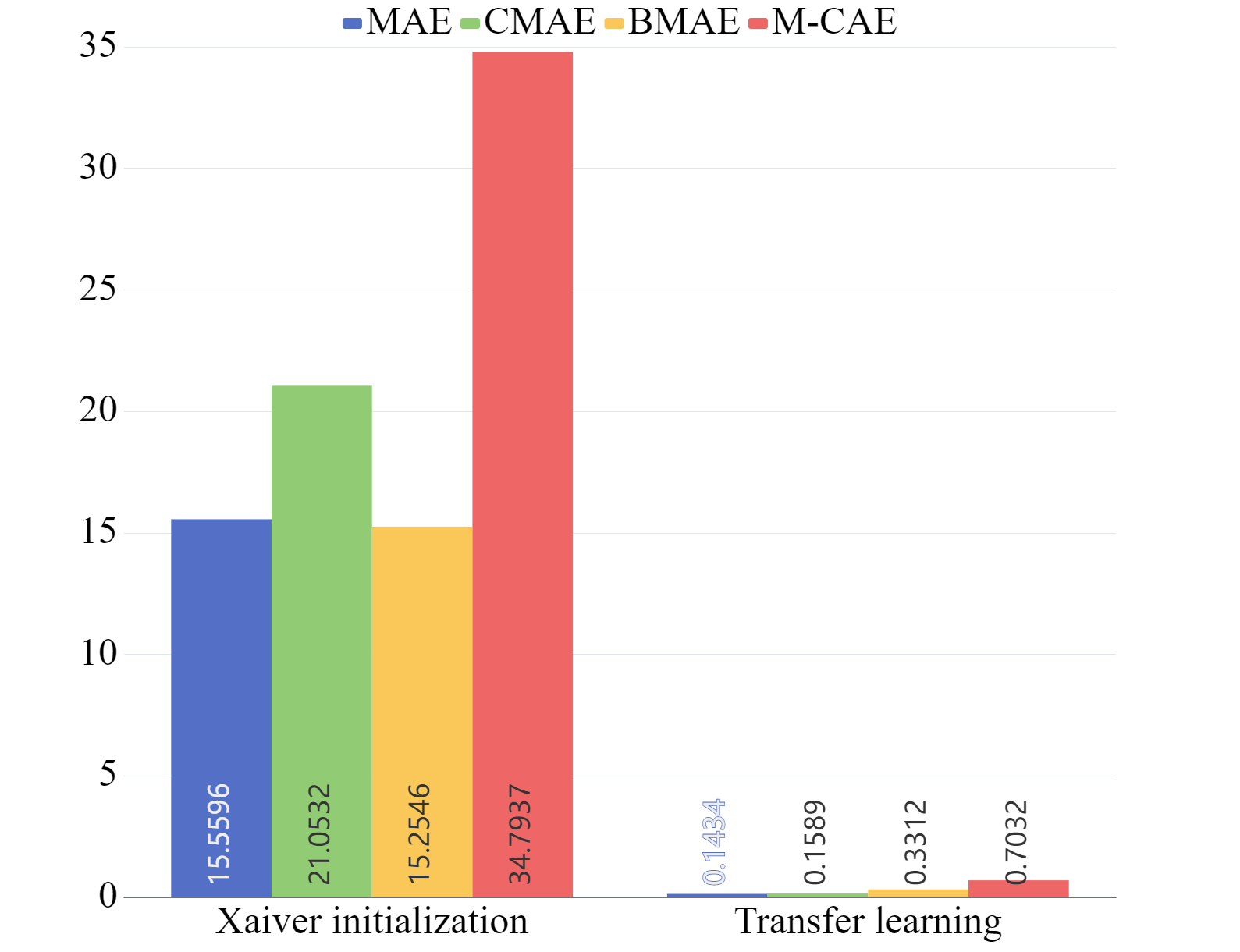}
	}
	\quad
	\subfigure[Case 3 with LHS]{
		\includegraphics[width=0.31\linewidth]{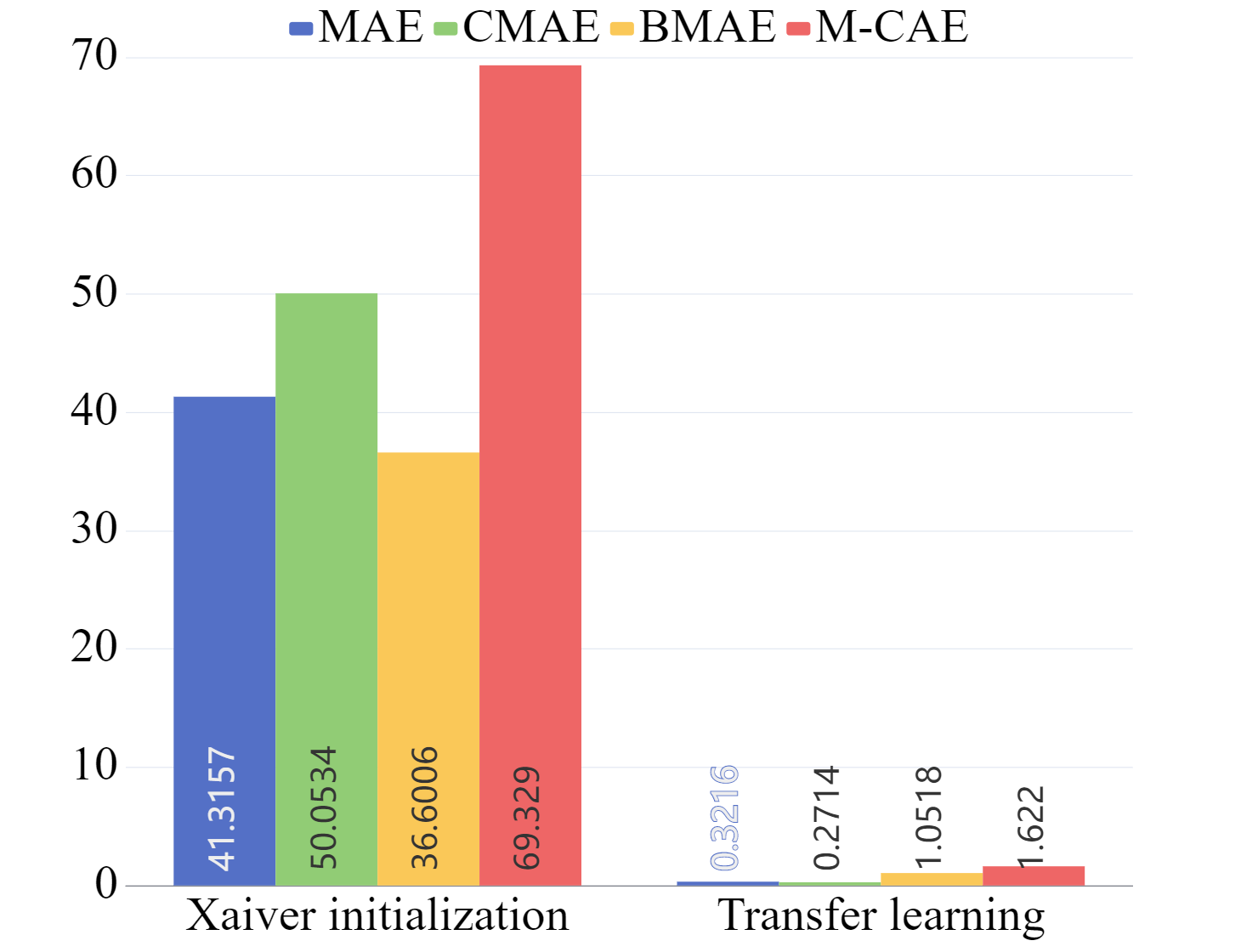}
	}
	\subfigure[Case 3 with LDS]{
		\includegraphics[width=0.31\linewidth]{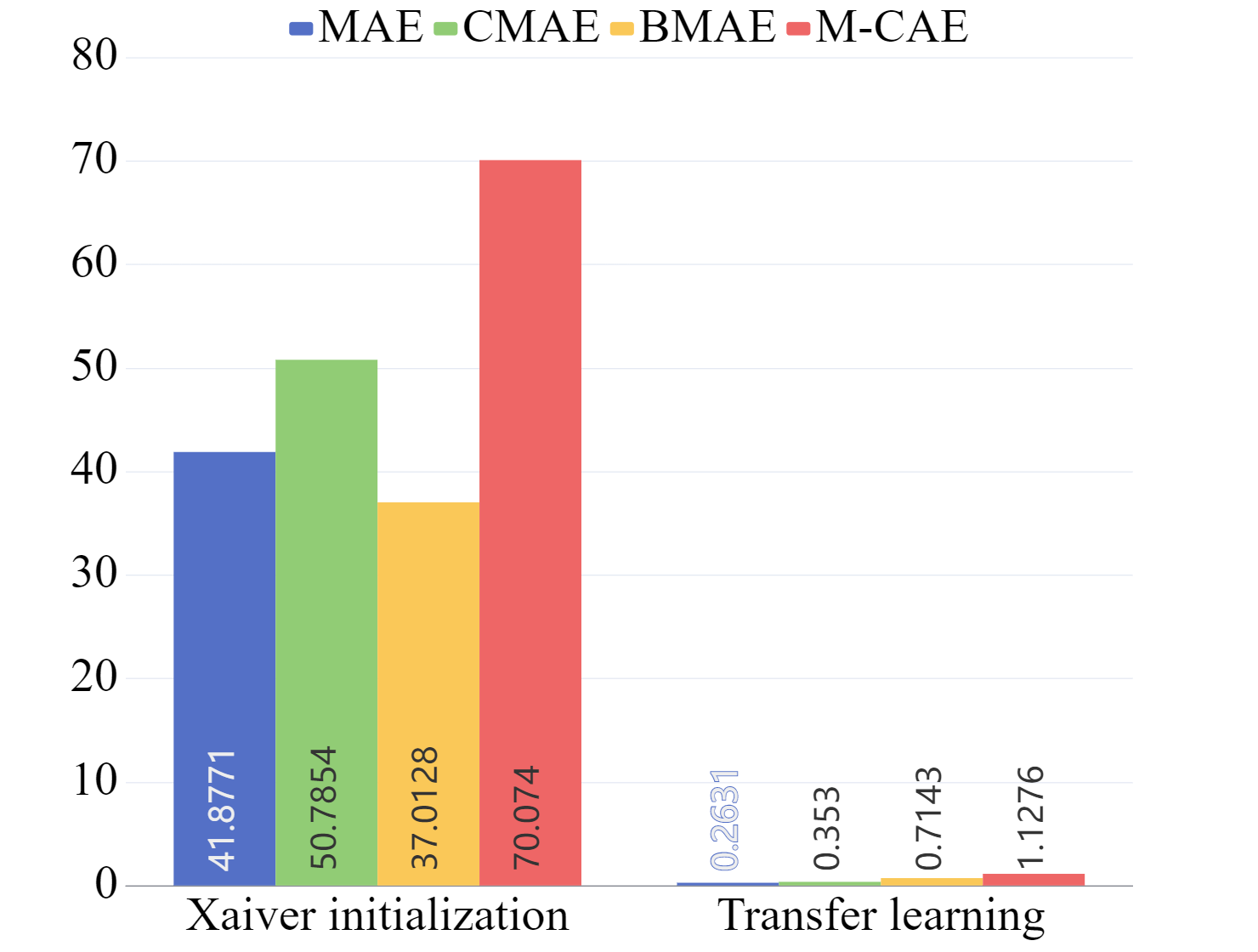}
	}
	\subfigure[Case 3 with GS]{
		\includegraphics[width=0.31\linewidth]{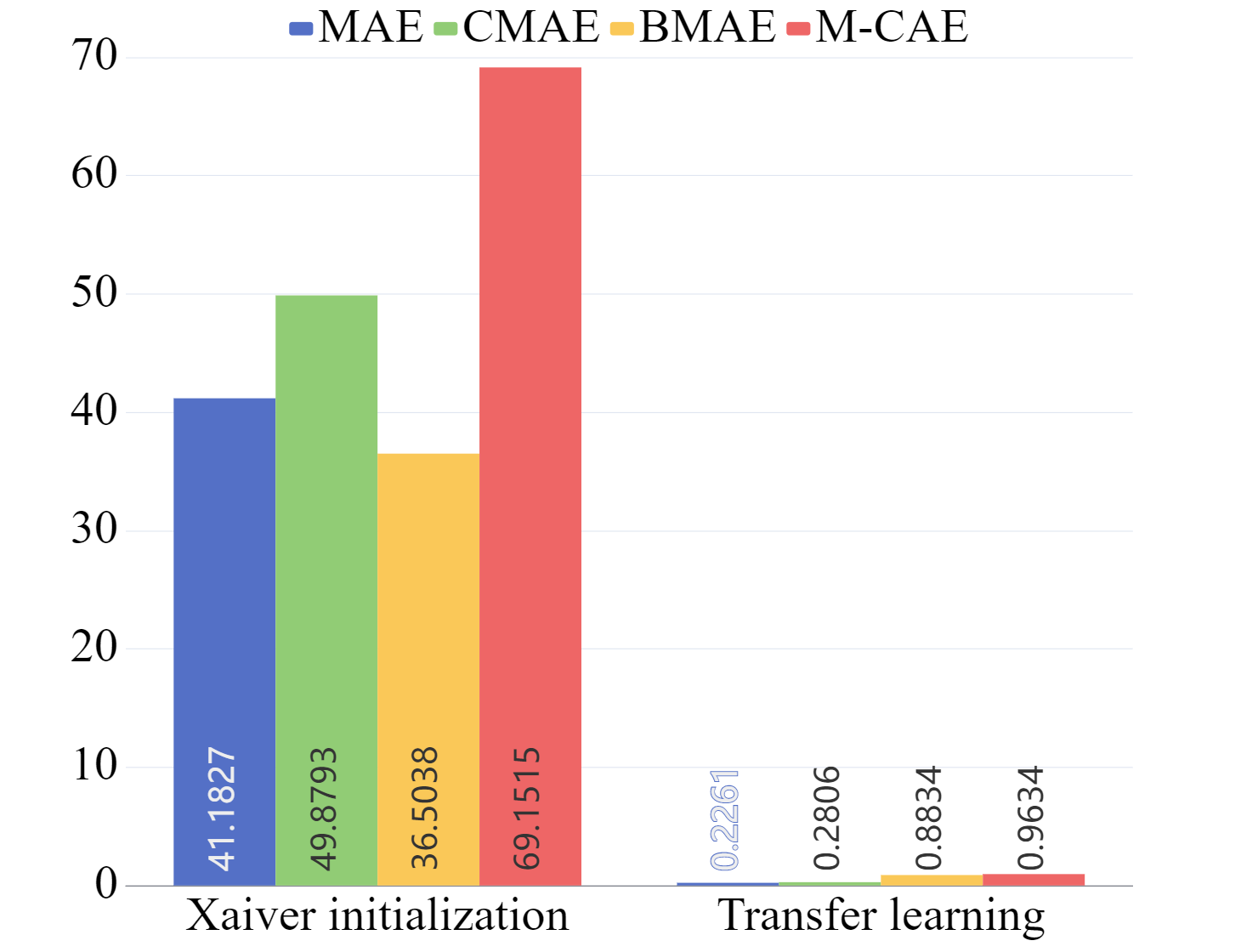}
	}
	\caption{Performance of the PINN-TFI method with Xaiver initialization and transfer learning.}
	\label{fig:ini}
\end{figure*}

\begin{figure*}[!htbp]
	\centering
	\subfigure[Case 1]{
		\centering
		\includegraphics[width=0.21\textheight]{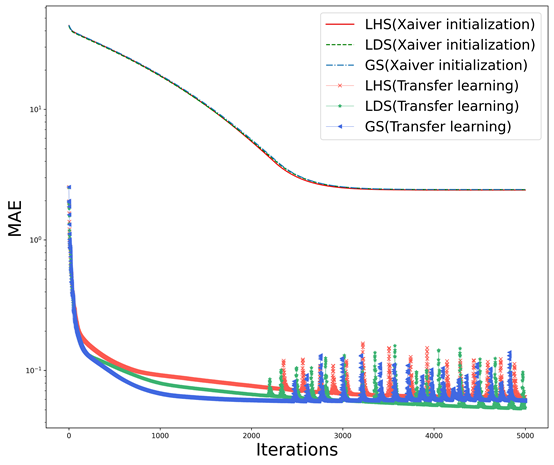}
		\label{fig:case1_ini}
	}
	\subfigure[Case 2]{
		\centering
		\includegraphics[width=0.2058\textheight]{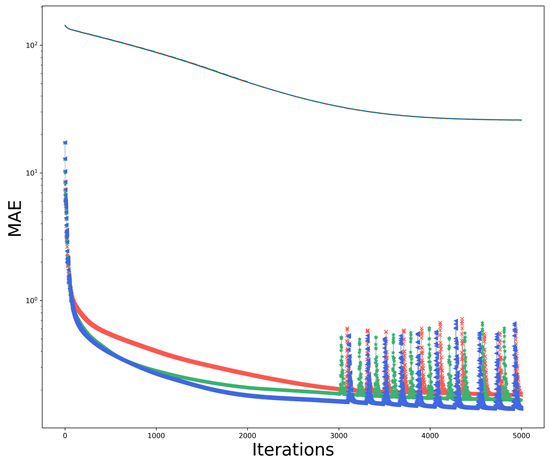}
		\label{fig:case2_ini}
	}
	\subfigure[Case 3]{
		\centering
		\includegraphics[width=0.2058\textheight]{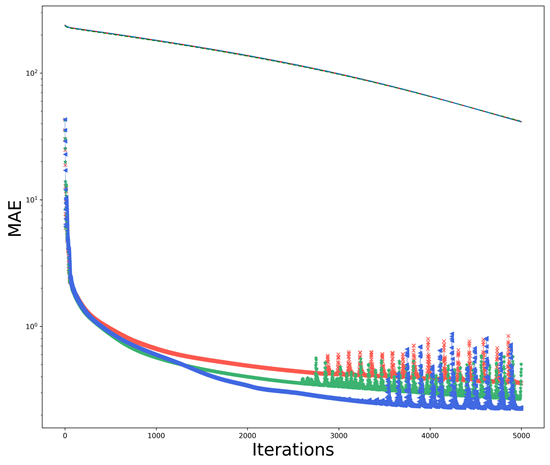}
		\label{fig:case3_ini}
	}
	\caption{The historical MAE of PINN-TFI method over three cases with transfer learning and Xaiver initialization.}
	\label{fig:case_ini}
\end{figure*}

Fig. \ref{fig:ini} shows the performance of the PINN-TFI method over three cases with transfer learning and Xaiver initialization under 5 000 iterations. MAEs, CMAEs, BMAEs, and M-CAEs of the PINN-TFI method with transfer learning are significantly less than that of the PINN-TFI method with Xaiver initialization whatever observations are from LHS, LDS, or GS. For three cases, MAEs, CMAEs, BMAEs, and M-CAEs of the PINN-TFI method with Xaiver initialization are close whatever observations are from LHS, LDS, or GS. Fig. \ref{fig:case_ini} shows the historical MAEs of the PINN-TFI method over three cases with transfer learning and Xaiver initialization. More obviously, the curves of MAEs with Xaiver initialization are almost close for observations from LHS, LDS, and GS. It means that, in the early stage of the training process, positions of observations have little influence on the NN with Xaiver initialization. For observations from LHS, LDS, and GS, curves of MAEs with transfer learning are close within about 100 iterations. The curves of MAEs with transfer learning are going down very fast and MAEs with transfer learning converge to different values. Transfer learning strategy has embedded physical informations learned in the model initialization part into the model for temperature field inversion. It means that transfer learning strategy endows the model for temperature field inversion with a good start, which makes the model efficiently learning from observations.

\begin{figure*}[!htbp]
	\centering
	\subfigure[MAE in case 1]{
		\centering
		\includegraphics[width=0.21\textheight]{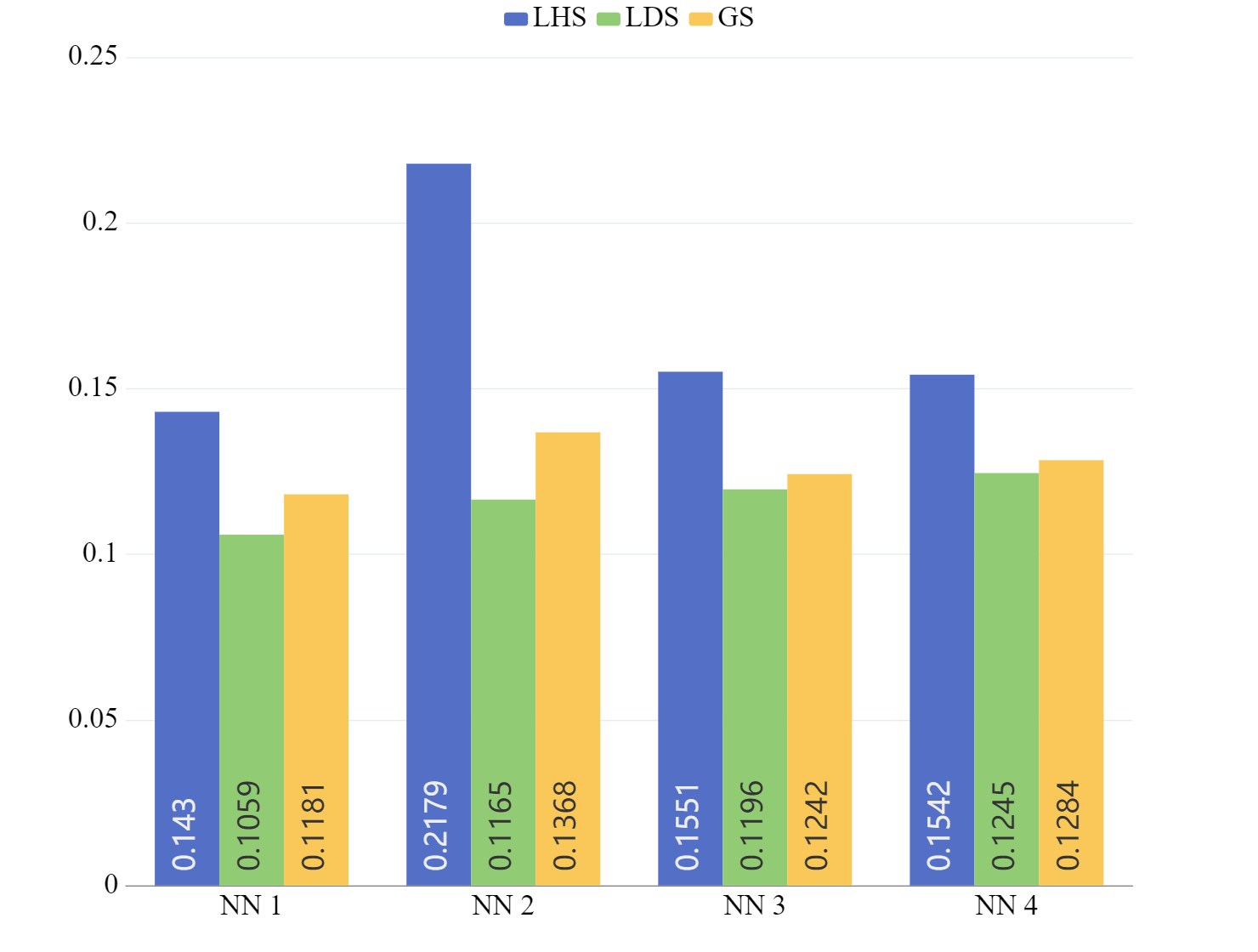}
		\label{fig:case1_NN}
	}
	\subfigure[MAE in case 2]{
		\centering
		\includegraphics[width=0.23\textheight]{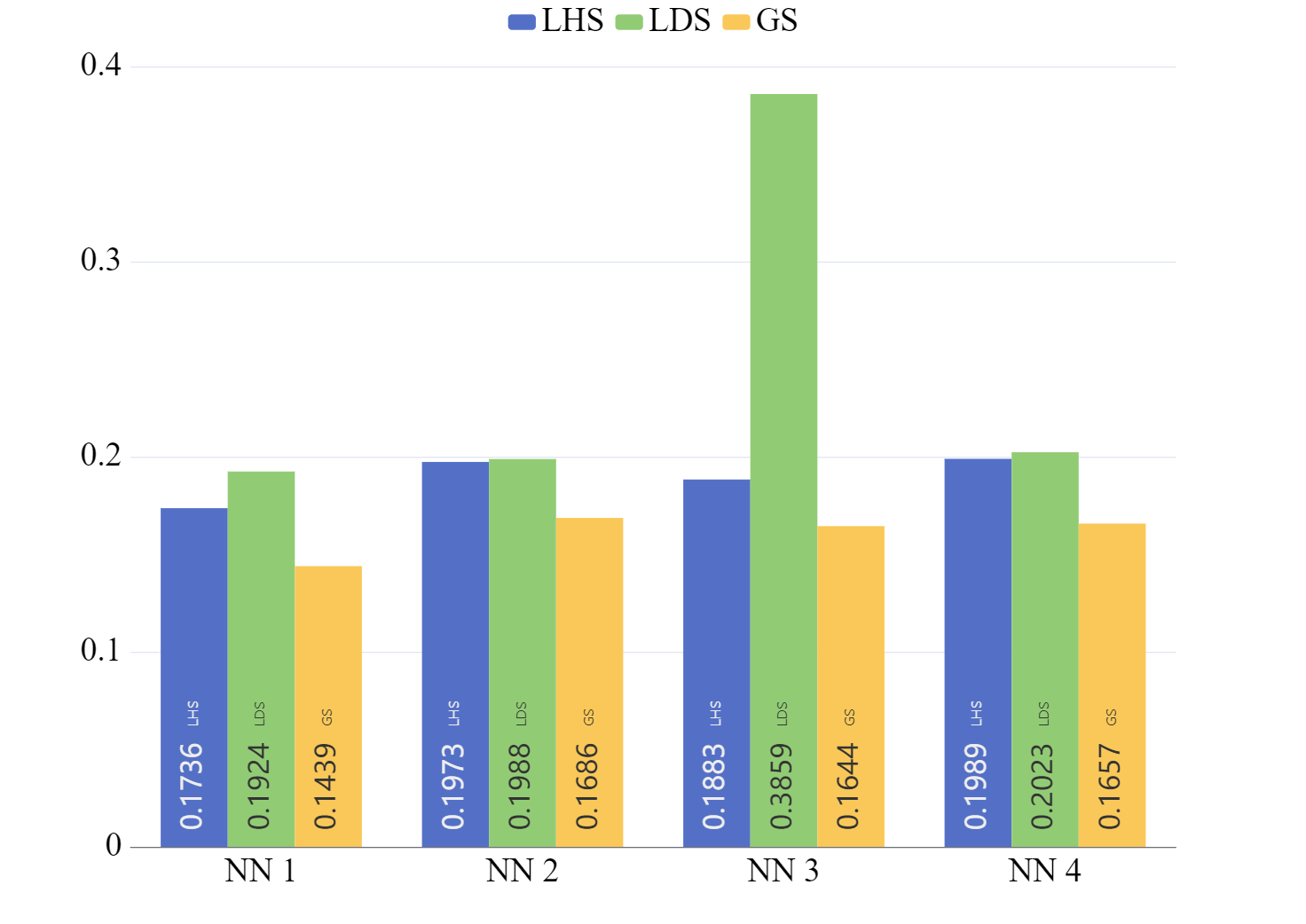}
		\label{fig:case2_NN}
	}
	\subfigure[MAE in case 3]{
		\centering
		\includegraphics[width=0.23\textheight]{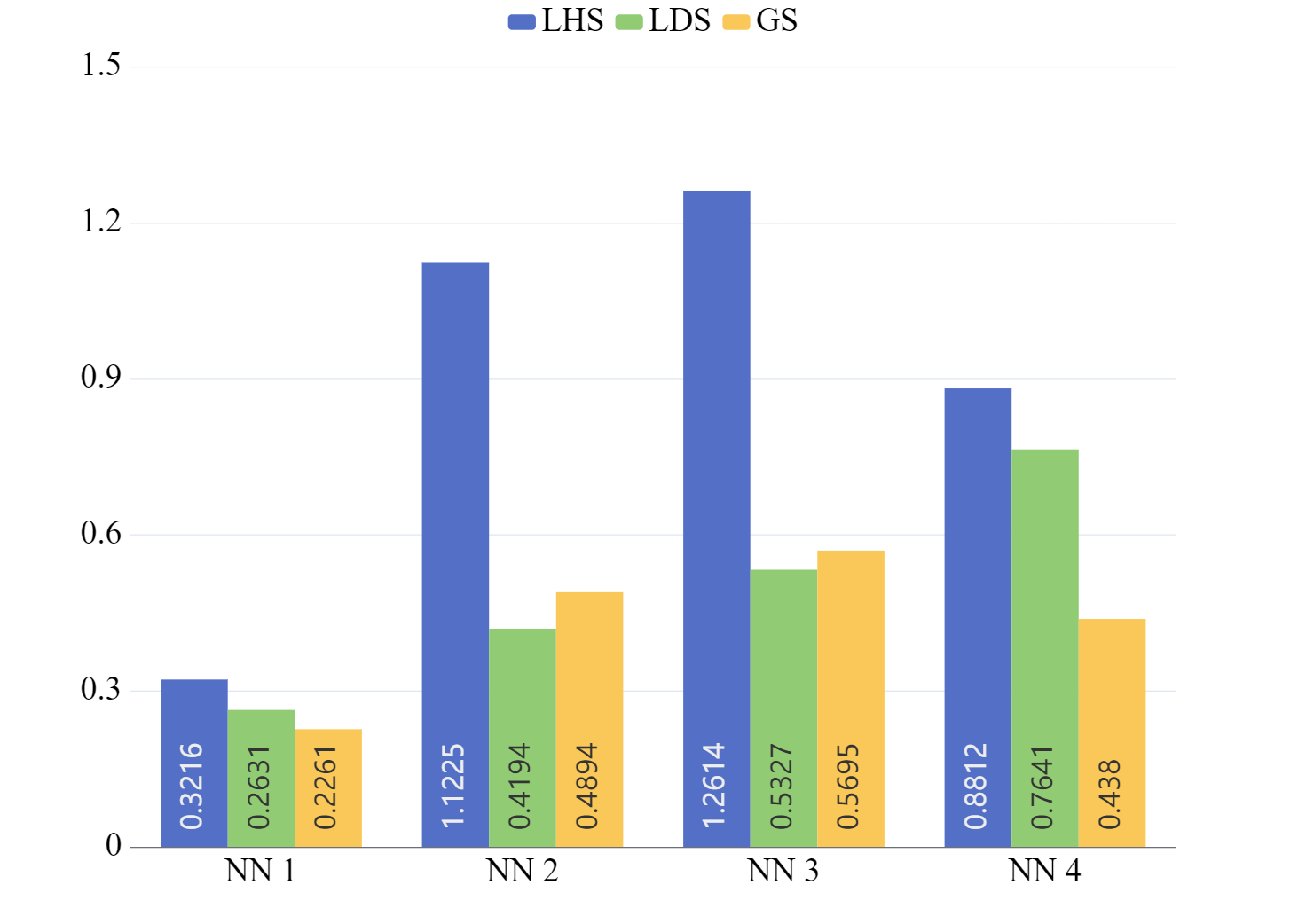}
		\label{fig:case3_NN}
	}
	\caption{MAEs of the PINN-TFI method with observations from LHS, LDS, and GS under different NNs.}
	\label{fig:case_NN}
\end{figure*}

\subsubsection{Performance with different width and depth of the NN}
We conduct experiments with different NN architectures. The NN 1 has four hidden layers with 50 neurons in each layer, which is the default NN architecture setting in this work. The NN 2 has four layers with 100 neurons in each layer. The NN 3 has five layers with 50 neurons in each layer and the NN 4 has three layers with 50 neurons in each layer. 

\begin{table*}[!htbp]
	\caption{Performance of the PINN-TFI method with observations from LHS, LDS, and GS under different widths and depths of NNs. The best results under different NN architectures are highlight.}
	\label{tab:NN_architectures}
	\centering
	\scalebox{0.74}
	{
		\begin{tabular}{cccccclcccclcccc}
			\hline
			\multirow{2}{*}{NN} & \multirow{2}{*}{Position} & \multicolumn{4}{c}{Case 1}                                            &  & \multicolumn{4}{c}{Case 2}                                   &  & \multicolumn{4}{c}{Case 3}                                            \\ \cline{3-6} \cline{8-11} \cline{13-16} 
			&                           & MAE             & CMAE            & BMAE            & M-CAE           &  & MAE             & CMAE   & BMAE            & M-CAE           &  & MAE             & CMAE            & BMAE            & M-CAE           \\ \hline
			& LHS                       & 0.1430          & 0.1404          & 0.1180          & 0.5387          &  & 0.1736          & 0.1569 & \textbf{0.1740} & 0.8939          &  & 0.3216          & \textbf{0.2714} & 1.0518          & 1.622           \\
			NN 1                & LDS                       & \textbf{0.1059} & 0.1370          & 0.1272          & 0.7929          &  & 0.1924          & 0.1772 & 0.2146          & \textbf{0.6410} &  & 0.2631          & 0.353           & \textbf{0.7143} & 1.1276          \\
			& GS                        & 0.1181          & \textbf{0.0948} & 0.1970          & \textbf{0.3472} &  & \textbf{0.1434} & 0.1589 & 0.3312          & 0.7032          &  & \textbf{0.2261} & 0.2806          & 0.8834          & \textbf{0.9634} \\ \hline
			& LHS                       & 0.2179          & 0.2179          & 0.2178          & 0.9259          &  & 0.1973          & 0.1838 & 0.2687          & 0.9562          &  & 1.1225          & 0.5746          & 5.2919          & 2.1522          \\
			NN 2                & LDS                       & 0.1165          & 0.0959          & \textbf{0.0882} & 0.5925          &  & 0.1988          & 0.1826 & 0.2062          & 0.8449          &  & 0.4194          & 0.4402          & 1.7399          & 1.5644          \\
			& GS                        & 0.1368          & 0.0952          & 0.1981          & 0.5274          &  & 0.1696          & 0.0957 & 0.3098          & 0.6655          &  & 0.4894          & 0.3042          & 2.3088          & 1.7257          \\ \hline
			& LHS                       & 0.1551          & 0.1540          & 0.1788          & 0.7427          &  & 0.1883          & 0.1528 & 0.3366          & 0.9703          &  & 1.2614          & 1.0379          & 5.0108          & 2.4625          \\
			NN 3                & LDS                       & 0.1196          & 0.1424          & 0.1933          & 0.8760          &  & 0.3859          & 0.4776 & 0.3539          & 1.3347          &  & 0.5327          & 0.8412          & 1.7858          & 2.1799          \\
			& GS                        & 0.1242          & 0.1083          & 0.2176          & 0.5793          &  & 0.1644          & 0.1242 & 0.3532          & 0.7579          &  & 0.5695          & 0.7951          & 1.9217          & 1.7909          \\ \hline
			& LHS                       & 0.1542          & 0.1543          & 0.1176          & 1.0330          &  & 0.1989          & 0.1699 & 0.3240          & 1.0549          &  & 0.8812          & 0.5461          & 3.7753          & 3.6405          \\
			NN 4                & LDS                       & 0.1245          & 0.1610          & 0.1165          & 0.8282          &  & 0.2023          & 0.2661 & 0.2434          & 0.8543          &  & 0.7641          & 0.6060          & 2.0173          & 2.0880          \\
			& GS                        & 0.1284          & 0.1141          & 0.2410          & 0.5655          &  & 0.1657          & 0.1300 & 0.3516          & 0.6657          &  & 0.4380          & 0.4266          & 2.1595          & 2.0787          \\ \hline
		\end{tabular}
	}
\end{table*}

Table \ref{tab:NN_architectures} shows the performance of the PINN-TFI method with different NN architectures, where the best results of MAEs, CMAEs, BMAEs, and M-CAEs can be obtained by NN 1 (default NN architecture) except for the BMAE of case 1 and the CMAE of case 2. Compared to NN 1, NN 2 has the same depth but has more neurons in each layer. As Fig. \ref{fig:case_NN} shows, MAEs of the PINN-TFI method with NN 2 are larger than ones with NN 1. For NN1 and NN3, the more neurons in the hidden, the larger MAEs. Additionally, the MAE of the PINN-TFI method with NN 3 over case 3 tends to be 1.2614K, which is significantly larger than 0.3216K with NN 1. It means that an unusually large number of neurons in the hidden layers will cause overfitting. Compared to NN 1, NN 4 has the same number of neurons in each layer but has the less depth. MAEs of the PINN-TFI method with NN 4 is worse than ones with NN 1. It means that a shallow NN architecture can not learn enough physics information to depict the whole temperature field. In a summary, a NN architecture of appropriate width and depth enable the PINN-TFI method to achieve better performance.


\subsubsection{Performance with different weights}

To encode physics and observation informations into the loss function, the PINN-TFI method constructs three different losses, namely the PDE loss, BC loss, and Data loss. Three losses are combined with weights, which play different roles in the training process. This work designs different weights to discuss the effect of different losses. The default weight setting ($w_{pde}=1$, $w_{bc}=1$ and $w_{data}=1e4$) is used the baseline of the comparison.

\begin{figure*}[!htbp]
	\centering
	\subfigure[Case 1 with LHS]{
		\includegraphics[width=0.31\linewidth]{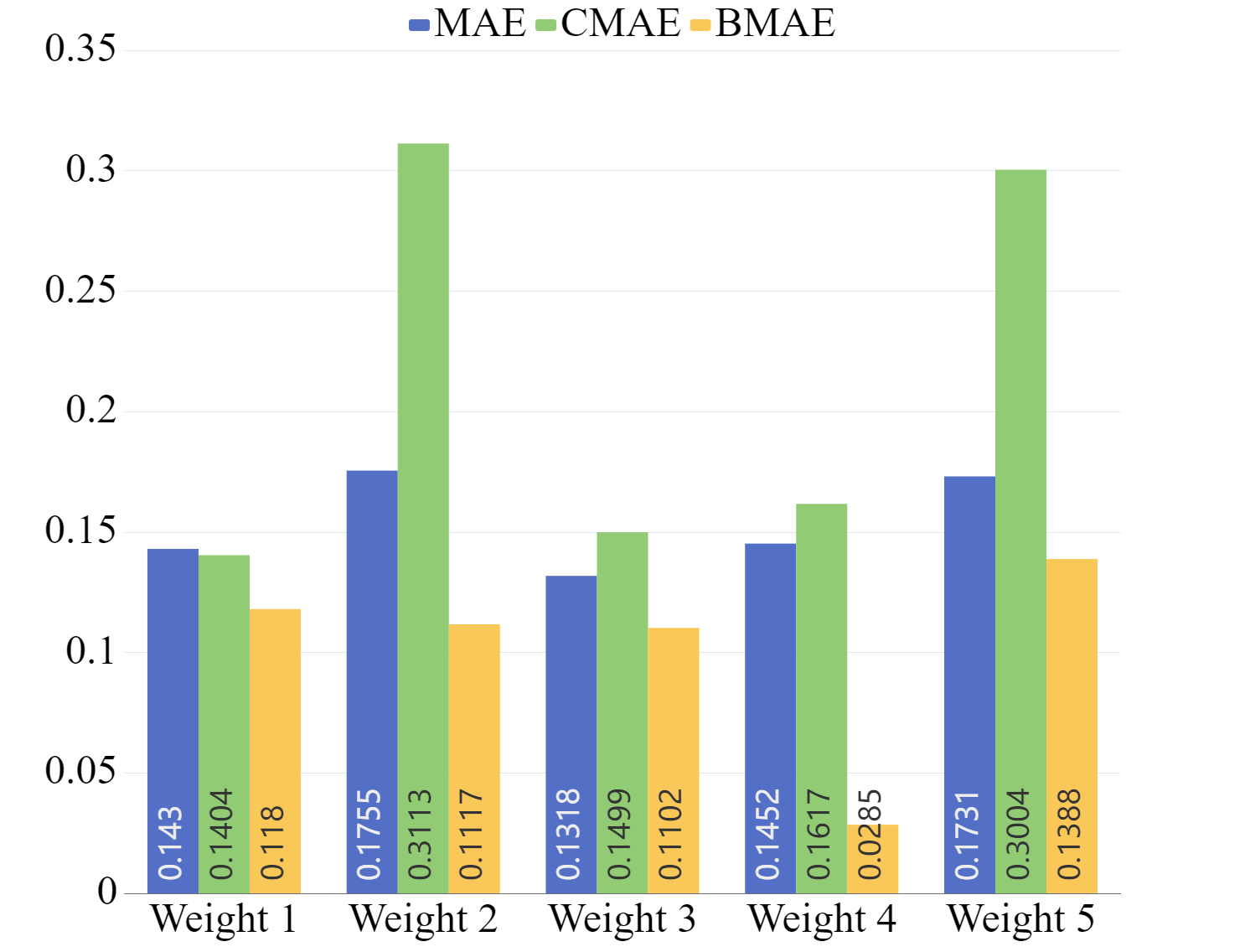}
	}
	\subfigure[Case 1 with LDS]{
		\includegraphics[width=0.31\linewidth]{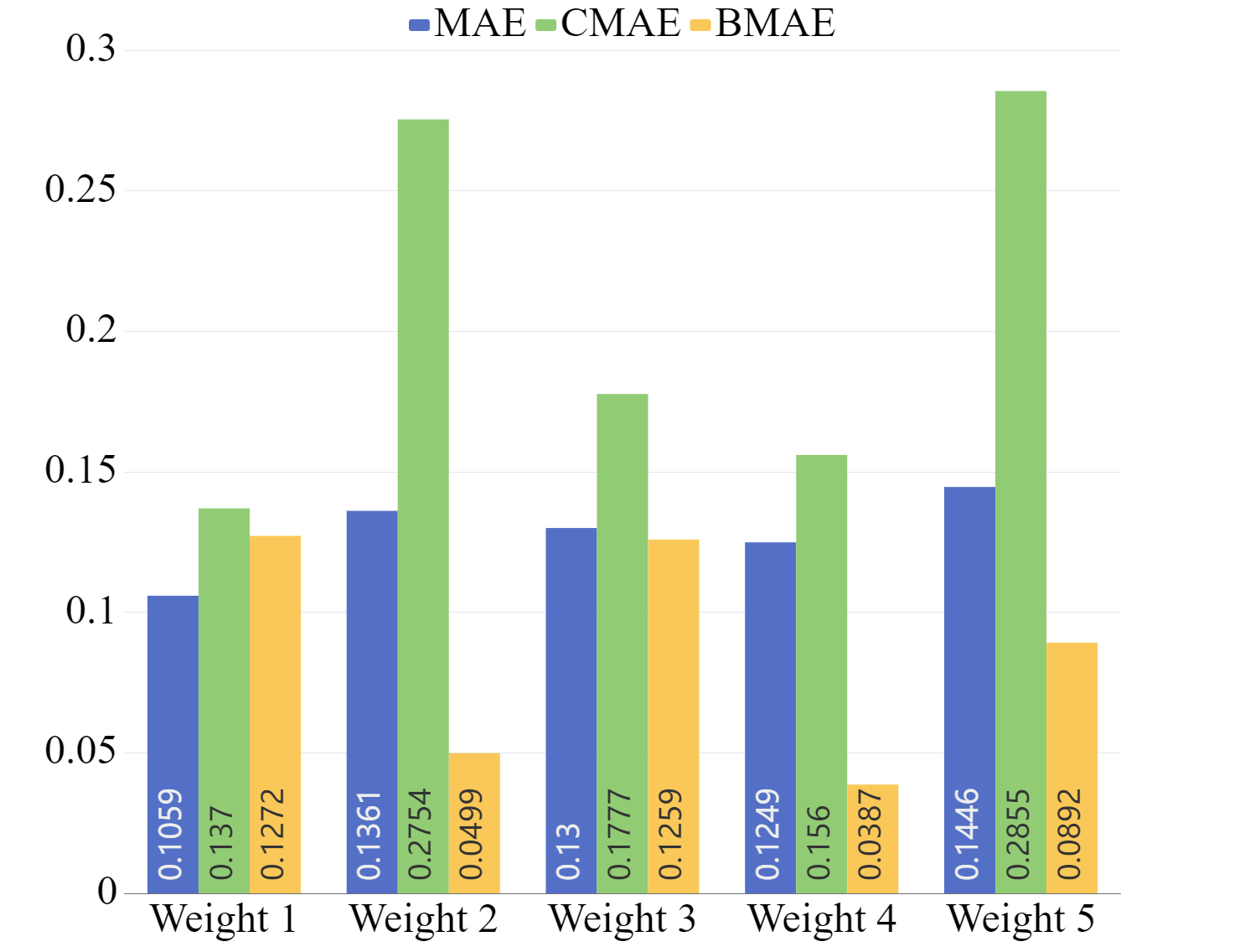}
	}
	\subfigure[Case 1 with GS]{
		\includegraphics[width=0.31\linewidth]{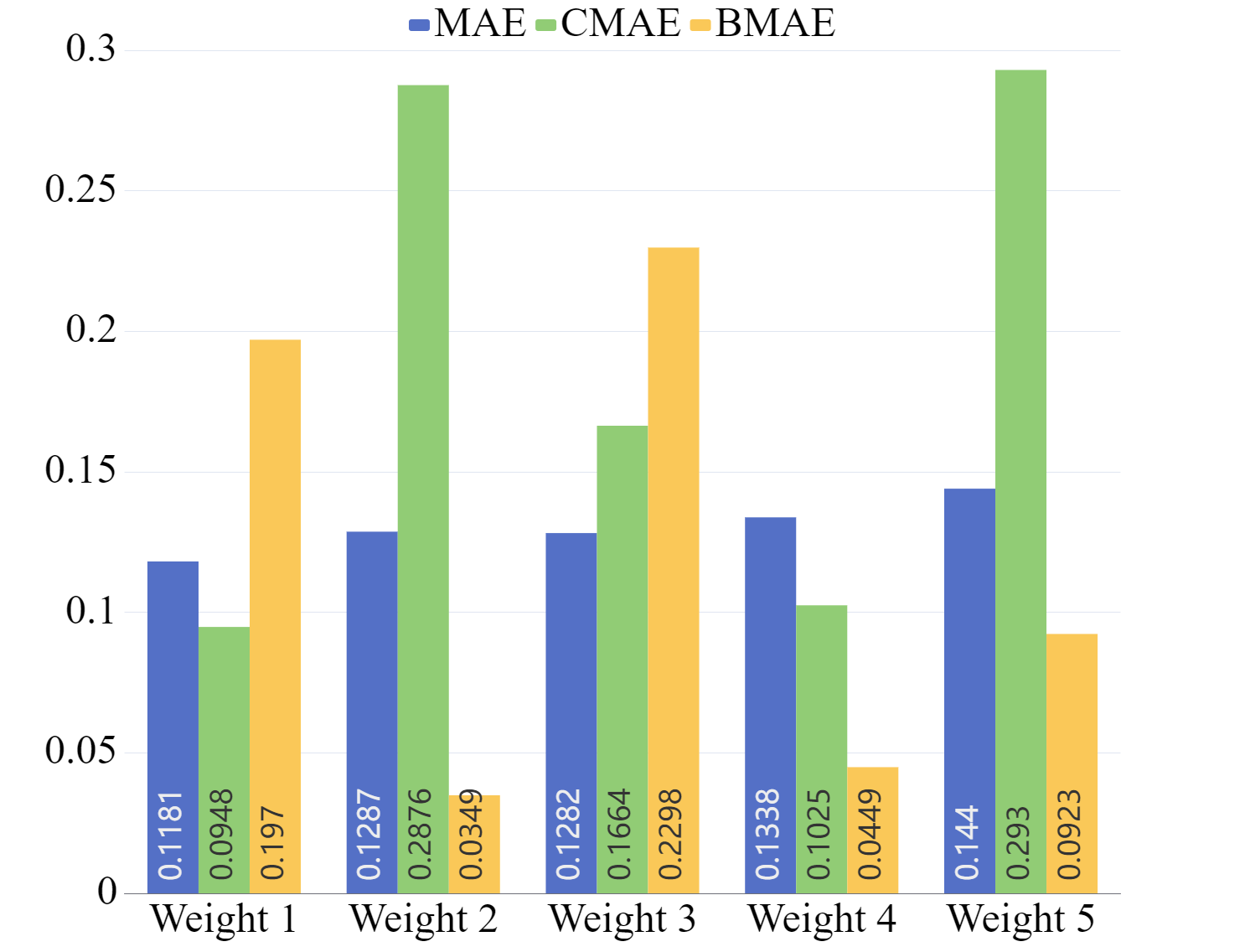}
	}
	\quad
	\subfigure[Case 2 with LHS]{
		\includegraphics[width=0.31\linewidth]{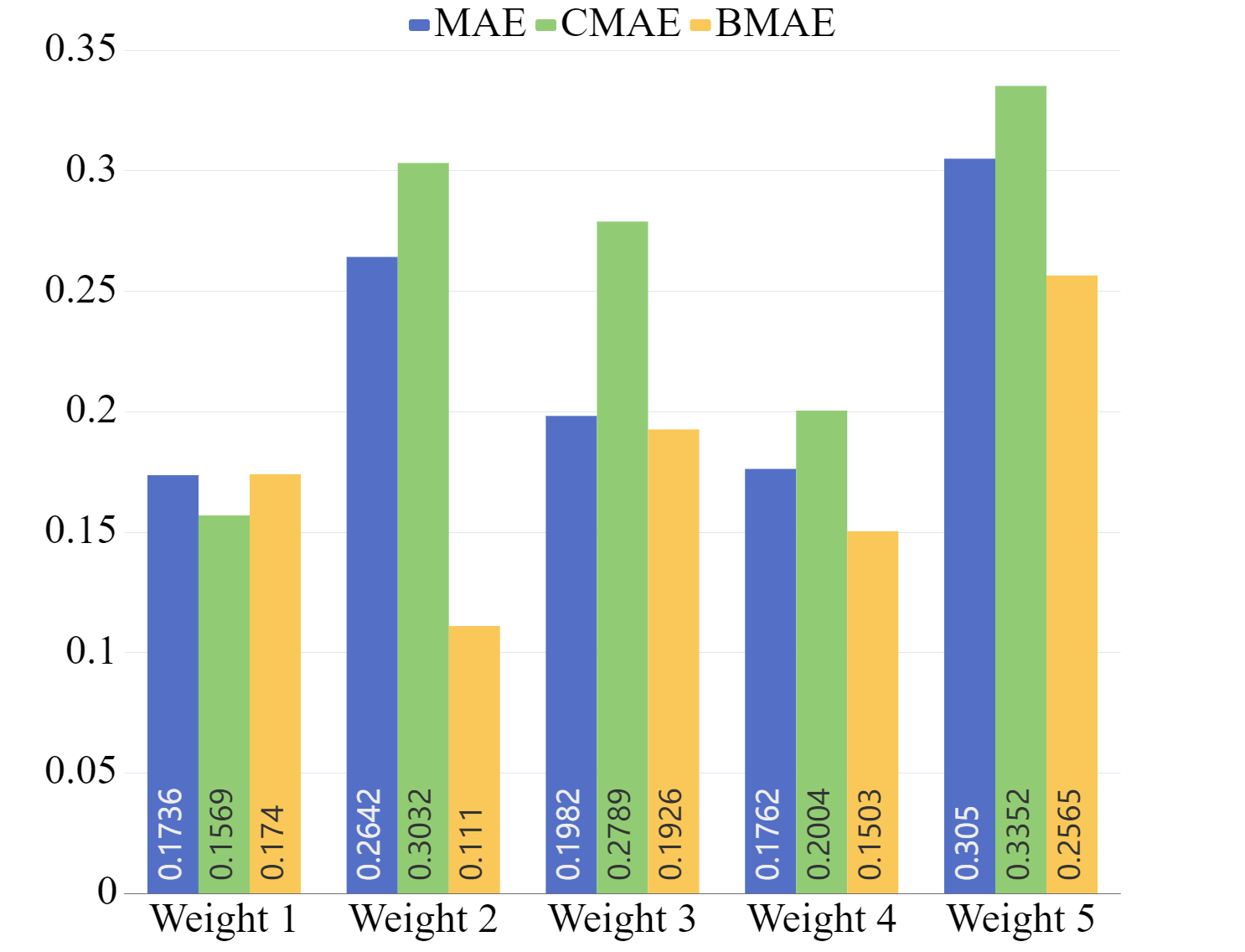}
	}
	\subfigure[Case 2 with LDS]{
		\includegraphics[width=0.31\linewidth]{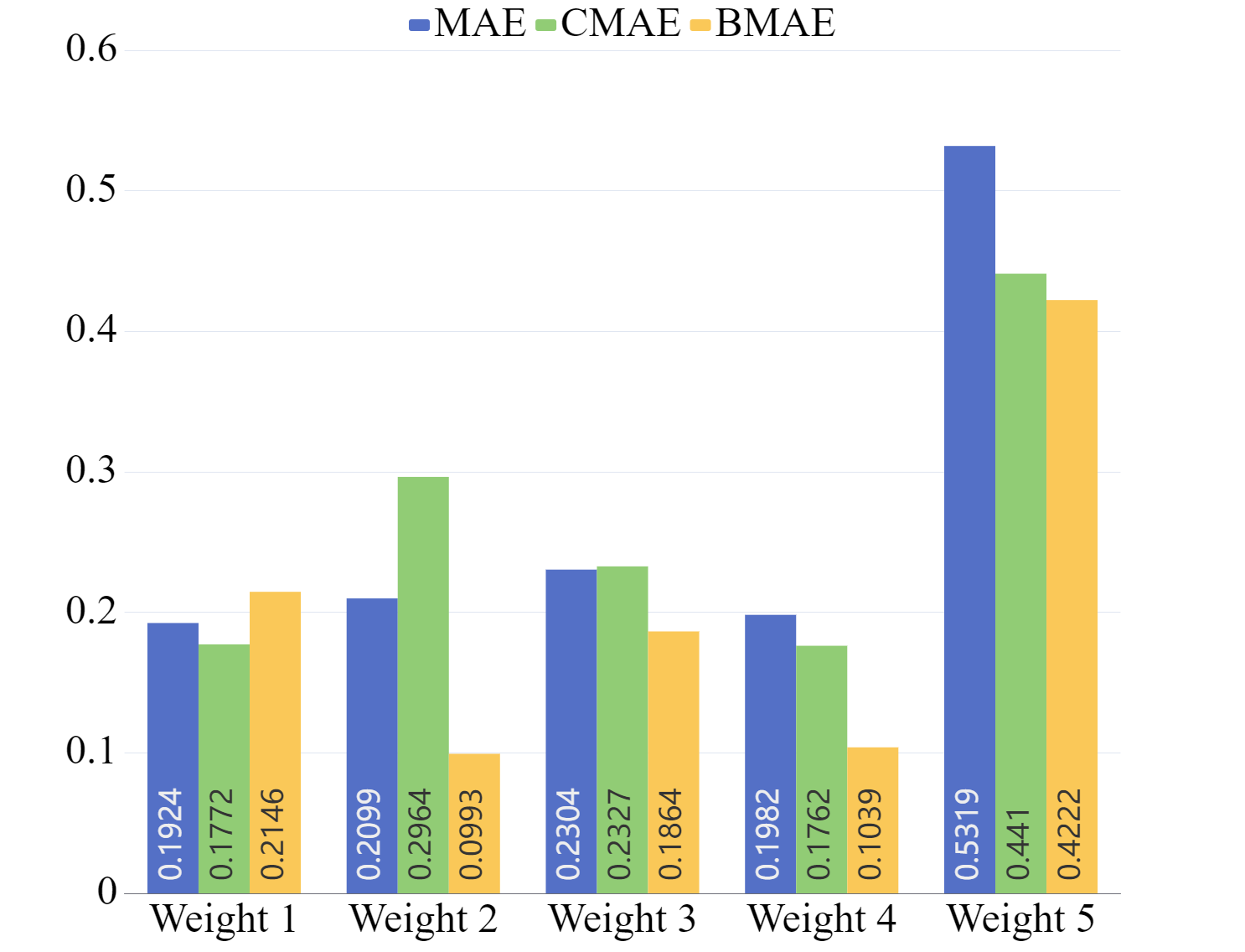}
	}
	\subfigure[Case 2 with GS]{
		\includegraphics[width=0.31\linewidth]{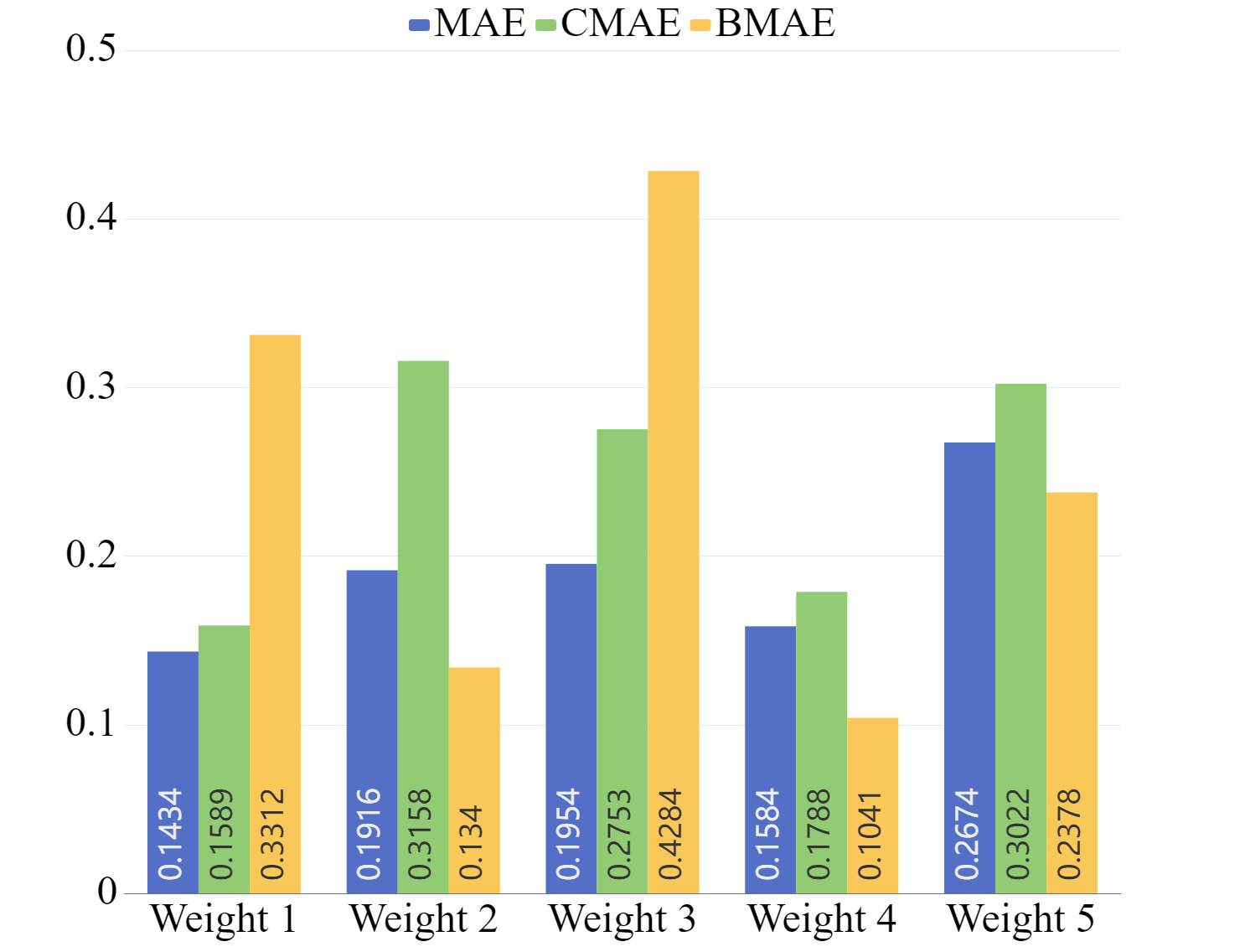}
	}
	\quad
	\subfigure[Case 3 with LHS]{
		\includegraphics[width=0.31\linewidth]{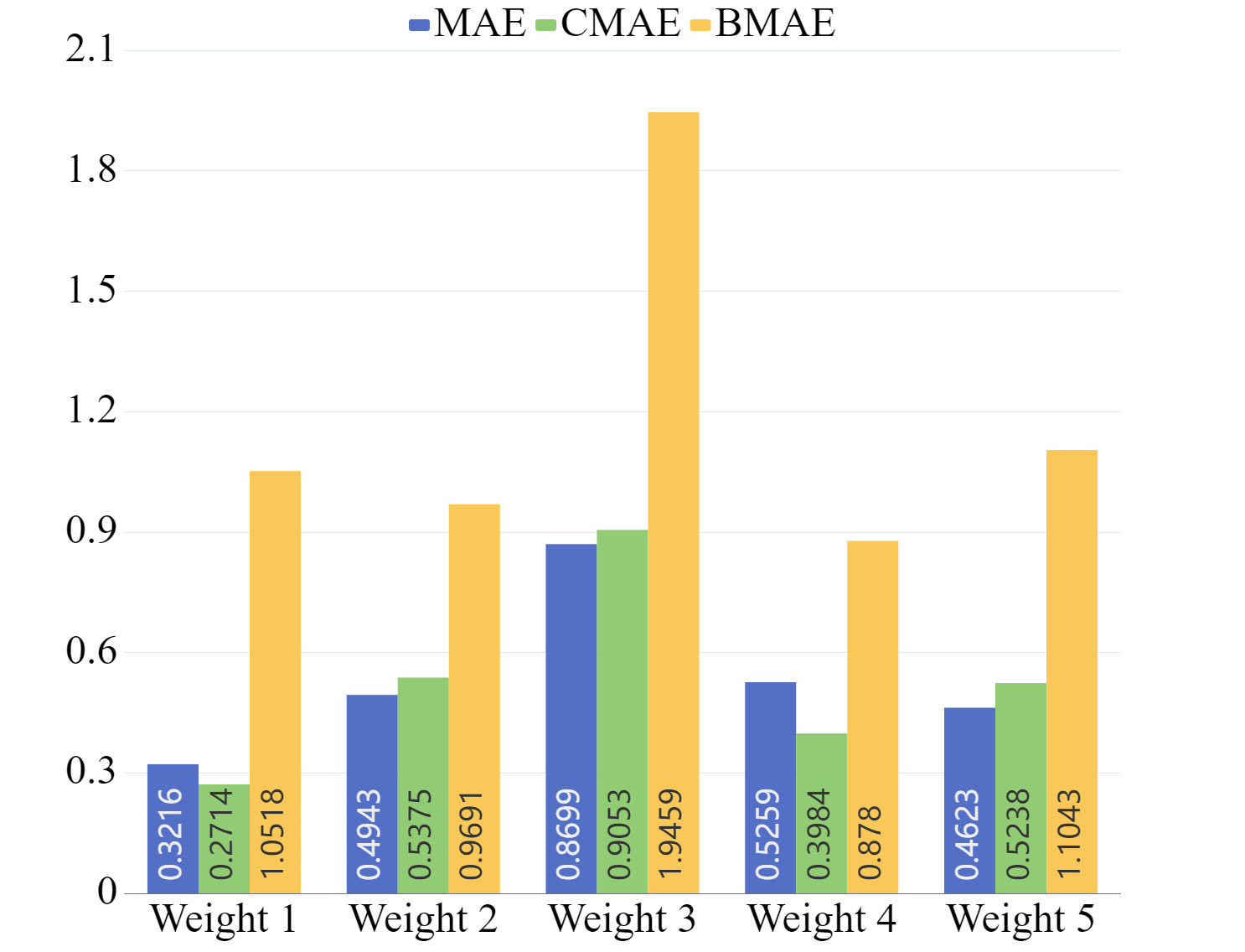}
	}
	\subfigure[Case 3 with LDS]{
		\includegraphics[width=0.31\linewidth]{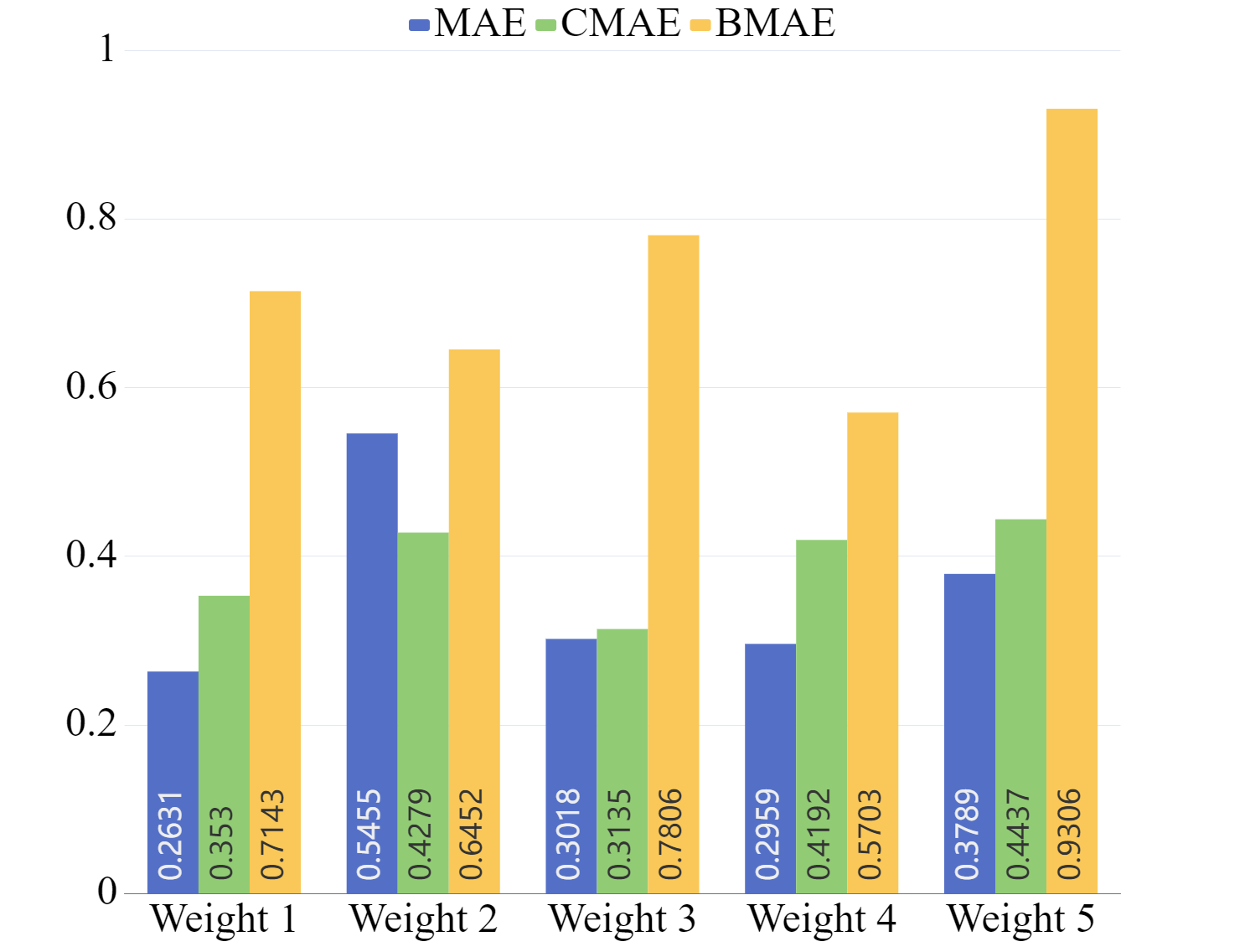}
	}
	\subfigure[Case 3 with GS]{
		\includegraphics[width=0.31\linewidth]{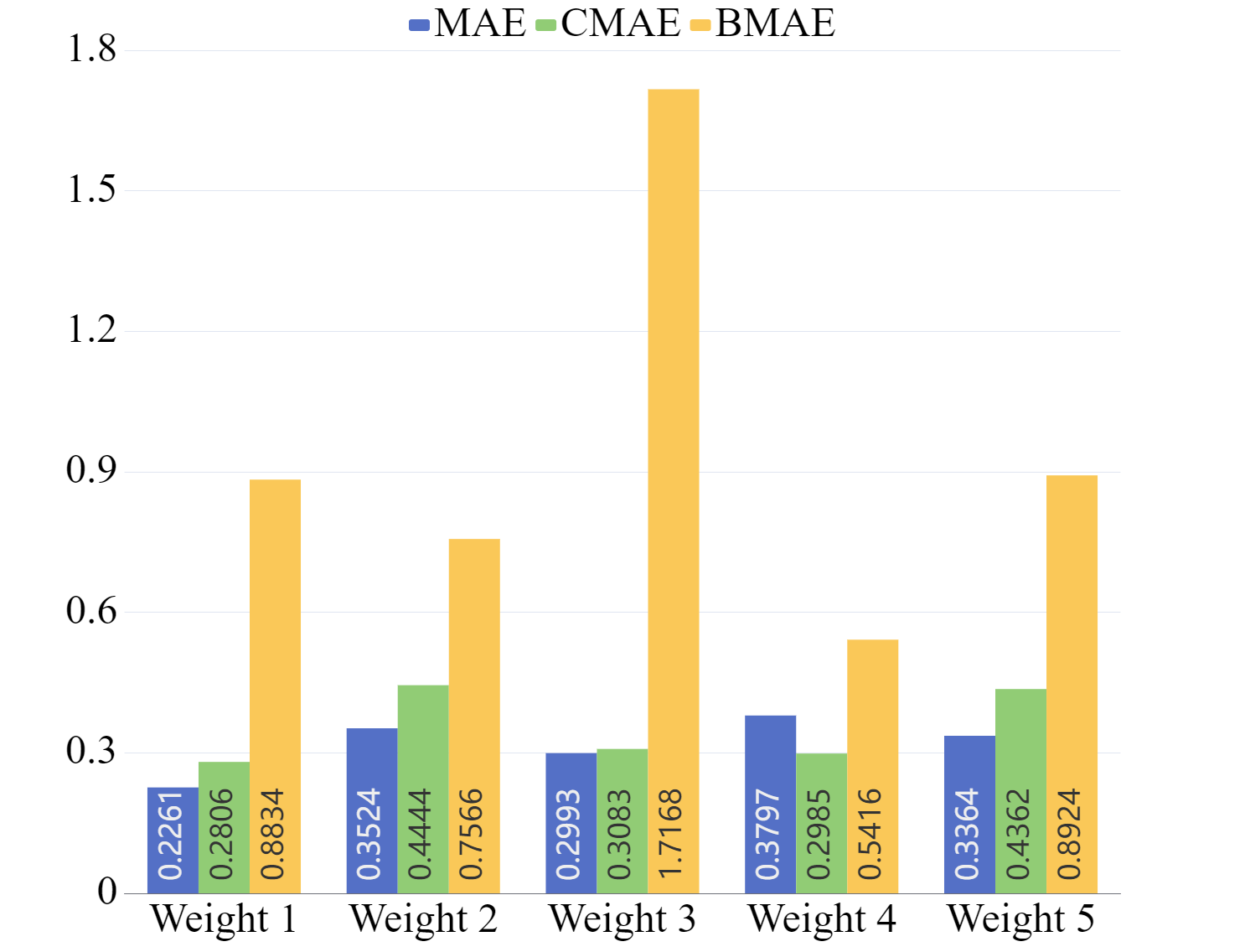}
	}
	\caption{Performance of the PINN-TFI method with observations from LHS, LDS and GS under different weights.}
	\label{fig:weight}
\end{figure*}

Fig. \ref{fig:weight} also shows the performance of the PINN-TFI method under different weights. As figure shows, compared with weight 1, weight 4 decreases $w_{bc}$. Thus, BMAEs under weight 4 are significantly less than that under weight 1. But a larger $w_{bc}$ causes a slight increment in MAEs. BC loss plays an important role for BMAEs. MAEs under weight 5 are larger than that under weight 1. This indicates that large $w_{pde}$ is not suitable for the PINN-TFI method. In addition, for different observations from LHS, LDS, and GS, the influence of those different weight terms are close. 

Table \ref{tab:3} lists results of the PINN-TFI method with observations from LHS, LDS, and GS under different weights. From the table, except for BMAEs, the best results of MAEs, CMAEs, and M-CAE can be obtained under weight 1. For weight 2, MAEs over three cases with GS tend to be $0.1181$K, $0.1434$K, and $0.2261$K, which are less than $0.1287$K, $0.1916$K and $0.3524$K with GS, respectively. This means that larger $w_{data}$ increases the proportion of data loss in the total loss, which results in smaller MAEs. But it can be seen from results under weight 1 and weight 3 that an unusually large $w_{data}$ may result in worse MAEs. Additional, larger $w_{data}$ will also reduce the proportion of BC loss in the total loss, causing worse BMAEs for the PINN-TFI method. In conclusion, Data loss has a great influence on MAEs of the PINN-TFI method.

\begin{table*}[!htbp]
	\caption{Performance of the PINN-TFI method with observations from LHS, LDS, and GS under different weights. The best results under different weights are highlight.}
	\label{tab:3}
	\centering
	\scalebox{0.74}{
		\begin{tabular}{lcccccclcccclcccc}
			\hline
			\multicolumn{2}{c}{\multirow{2}{*}{Weight}} & \multirow{2}{*}{Position} & \multicolumn{4}{c}{Case 1}                                            &  & \multicolumn{4}{c}{Case 2}                                            &  & \multicolumn{4}{c}{Case 3}                                   \\ \cline{4-7} \cline{9-12} \cline{14-17} 
			\multicolumn{2}{c}{}     &                           & MAE             & CMAE            & BMAE            & M-CAE           &  & MAE             & CMAE            & BMAE            & M-CAE           &  & MAE             & CMAE            & BMAE   & M-CAE           \\ \hline
			& $w_{PDE}$=1      & LHS                       & 0.1430          & 0.1404          & 0.1180          & 0.5387          &  & 0.1736          & 0.1569          & 0.1740          & 0.8939          &  & 0.3216          & \textbf{0.2714} & 1.0518 & 1.6220          \\
			1                        & $w_{BC}$=1       & LDS                       & \textbf{0.1059} & 0.1370          & 0.1272          & 0.7929          &  & 0.1924          & 0.1772          & 0.2146          & \textbf{0.6410} &  & 0.2631          & 0.3530          & 0.7143 & 1.1276          \\
			& $w_{BC}$=1e4     & GS                        & 0.1181          & \textbf{0.0948} & 0.1970          & \textbf{0.3472} &  & \textbf{0.1434} & \textbf{0.1589} & 0.3312          & 0.7032          &  & \textbf{0.2261} & 0.2806          & 0.8834 & \textbf{0.9634} \\ \hline
			& $w_{PDE}$=1      & LHS                       & 0.1755          & 0.3113          & 0.1117          & 1.2655          &  & 0.2642          & 0.3032          & 0.1010          & 1.2645          &  & 0.4943          & 0.5375          & 0.9691 & 1.9090          \\
			2                        & $w_{BC}$=1       & LDS                       & 0.1361          & 0.2754          & 0.0499          & 0.9108          &  & 0.2099          & 0.2964          & 0.0993          & 1.1959          &  & 0.5455          & 0.4279          & 0.6452 & 1.2501          \\
			& $w_{BC}$=1e2     & GS                        & 0.1287          & 0.2876          & 0.0349          & 0.9756          &  & 0.1916          & 0.3158          & 0.1340          & 1.1443          &  & 0.3524          & 0.4444          & 0.7566 & 1.3431          \\ \hline
			& $w_{PDE}$=1      & LHS                       & 0.1318          & 0.1499          & 0.1102          & 1.1359          &  & 0.1982          & 0.2789          & 0.1926          & 1.1940          &  & 0.8699          & 0.9053          & 1.9459 & 1.8760          \\
			3                        & $w_{BC}$=1       & LDS                       & 0.1300          & 0.1777          & 0.1259          & 0.7719          &  & 0.2304          & 0.2327          & 0.1864          & 1.3967          &  & 0.3018          & 0.3135          & 0.7806 & 1.1607          \\
			& $w_{BC}$=1e6     & GS                        & 0.1282          & 0.1664          & 0.2298          & 0.7088          &  & 0.1954          & 0.2753          & 0.4284          & 1.3503          &  & 0.2993          & 0.3083          & 1.7168 & 1.0400          \\ \hline
			& $w_{PDE}$=1      & LHS                       & 0.1452          & 0.1617          & \textbf{0.0285} & 0.6933          &  & 0.1762          & 0.2004          & 0.1503          & 1.1180          &  & 0.4259          & 0.3984          & 0.8780 & 1.4880          \\
			4                        & $w_{BC}$=1e2     & LDS                       & 0.1249          & 0.1560          & 0.0387          & 0.8034          &  & 0.1982          & 0.1762          & \textbf{0.1039} & 0.6677          &  & 0.2959          & 0.4192          & 0.5703 & 1.2571          \\
			& $w_{BC}$=1e4     & GS                        & 0.1338          & 0.1025          & 0.0449          & 0.3589          &  & 0.1584          & 0.1788          & 0.1041          & 0.7521          &  & 0.3797          & 0.2985          & 0.5416 & 1.1146          \\ \hline
			& $w_{PDE}$=1e2    & LHS                       & 0.1731          & 0.3004          & 0.1388          & 1.1978          &  & 0.3050          & 0.3352          & 0.2565          & 1.3302          &  & 0.4623          & 0.5238          & 1.1043 & 1.6181          \\
			5                        & $w_{BC}$=1       & LDS                       & 0.1446          & 0.2855          & 0.0892          & 0.8776          &  & 0.5319          & 0.4410          & 0.4222          & 0.9733          &  & 0.3789          & 0.4437          & 0.9306 & 1.4719          \\
			& $w_{BC}$=1e4     & GS                        & 0.1440          & 0.2930          & 0.0923          & 1.0036          &  & 0.2674          & 0.3022          & 0.2378          & 0.9963          &  & 0.3364          & 0.4362          & 0.8924 & 1.4793          \\ \hline
		\end{tabular}
	}
\end{table*}

\subsubsection{Comparisons with other methods}
This work uses Gaussian process regression (GPR) \citep{schulz2018tutorial} and random forest regression (RFR) \citep{segal2004machine} as baselines, which are implemented with Sklearn toolbox \citep{pedregosa2011scikit} (a very powerful machine learning Python library). It should be noted that all the results are obtained by calculating the average value of five independent repeated runs of the training. 
\begin{figure*}[!htbp]
	\centering
	\subfigure[Case 1 with LHS]{
		\includegraphics[width=0.3\linewidth]{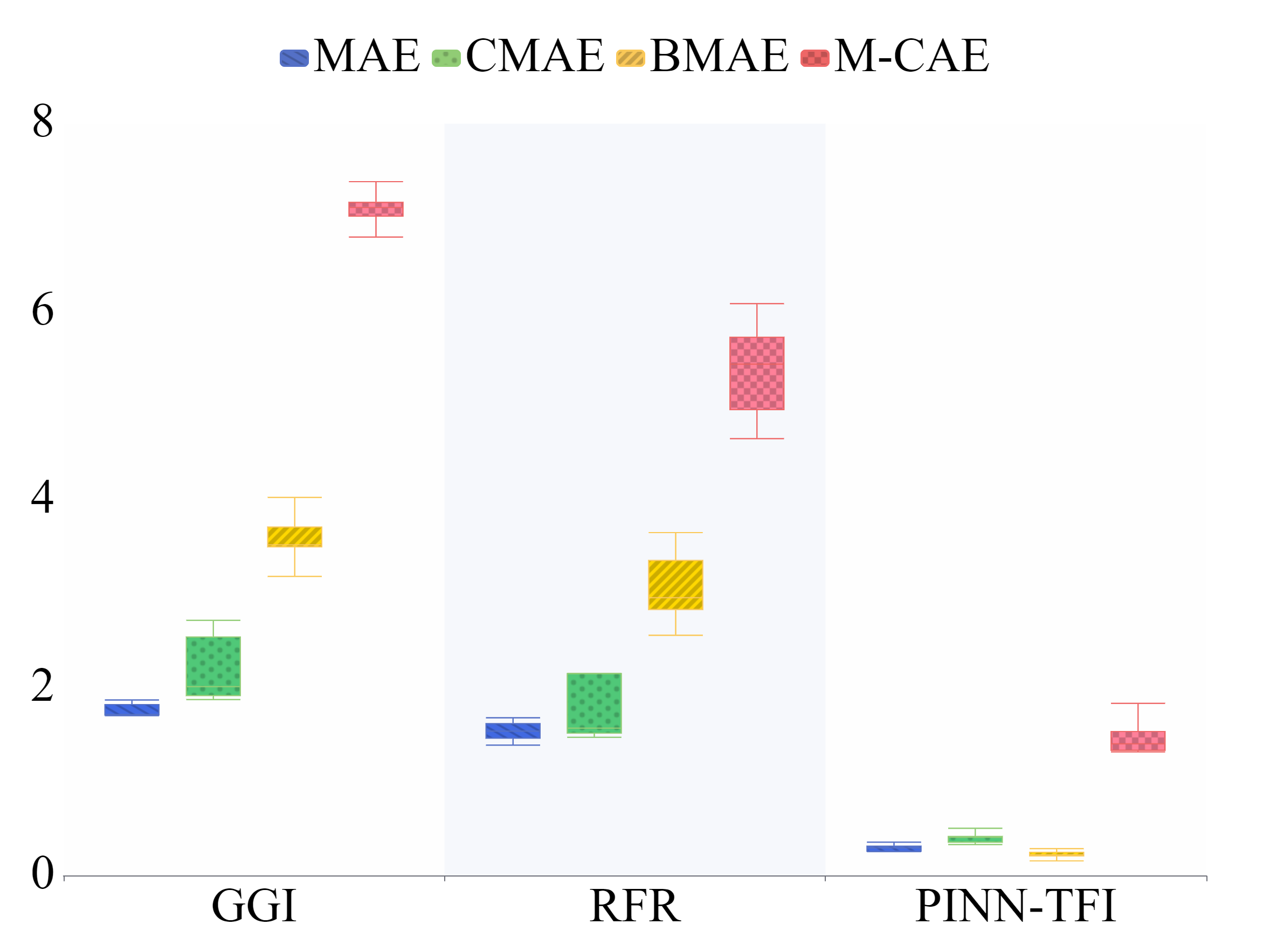}
	}
	\subfigure[Case 1 with LDS]{
		\includegraphics[width=0.3\linewidth]{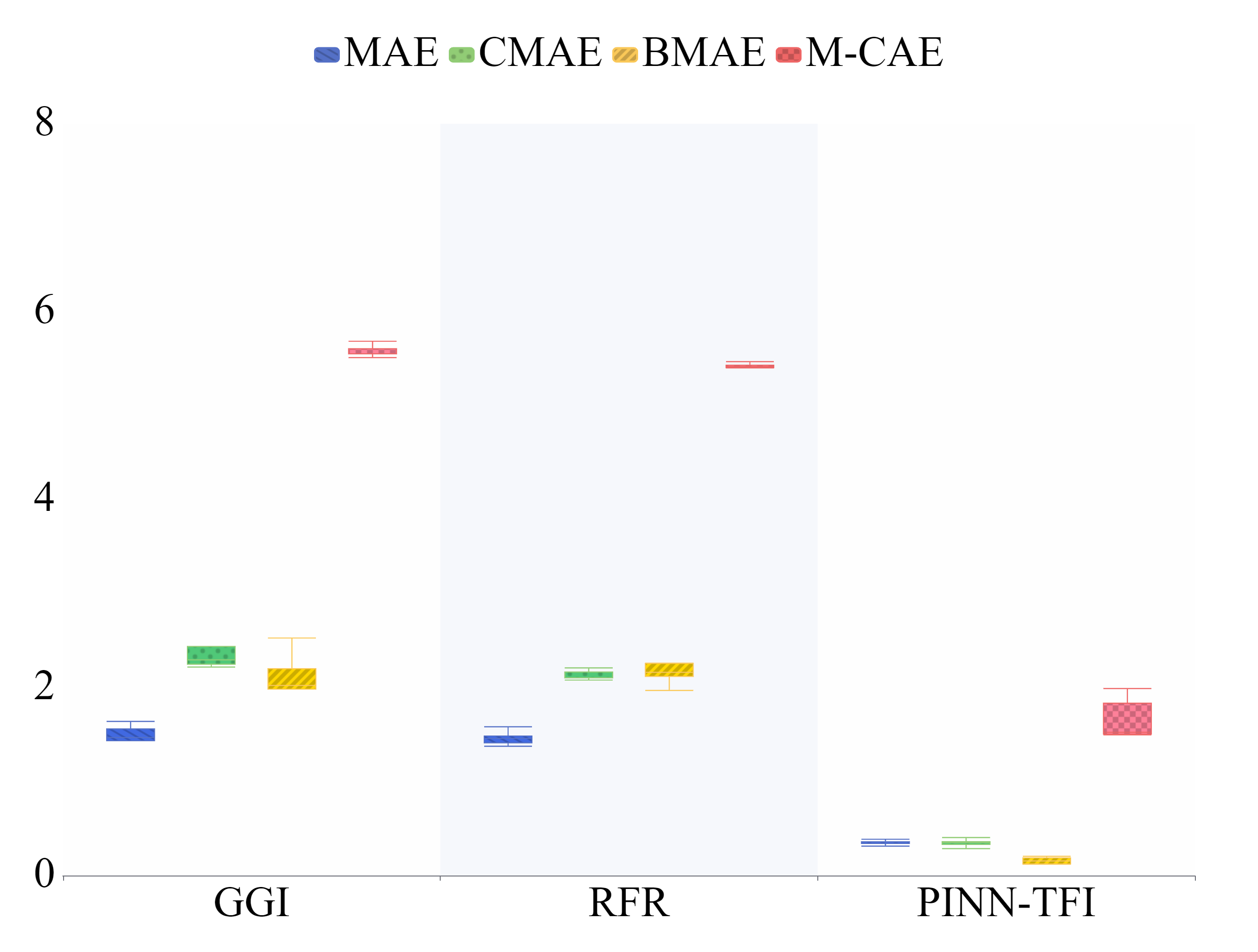}
	}
	\subfigure[Case 1 with GS]{
		\includegraphics[width=0.3\linewidth]{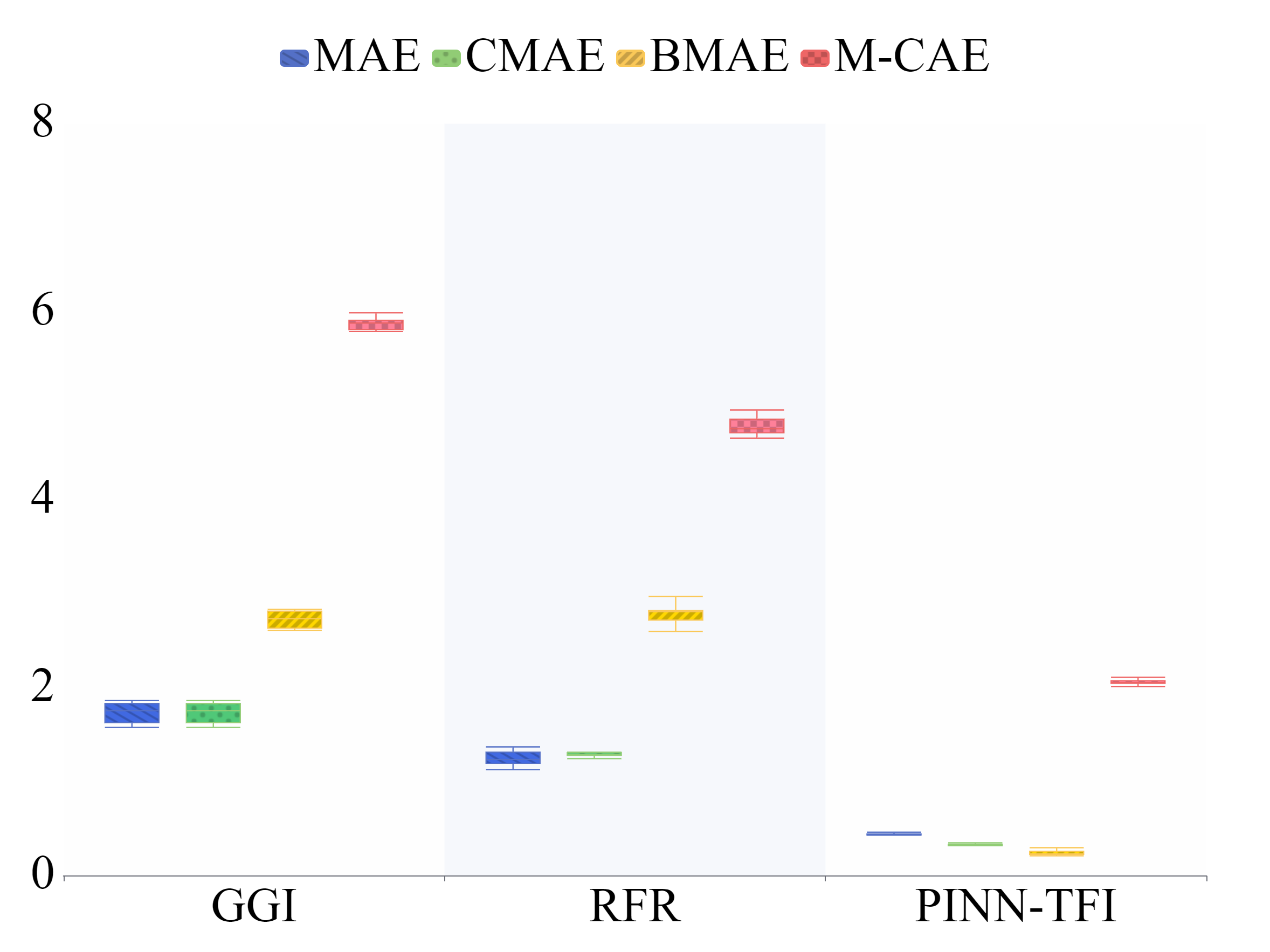}
	}
	\quad
	\subfigure[Case 2 with LHS]{
		\includegraphics[width=0.3\linewidth]{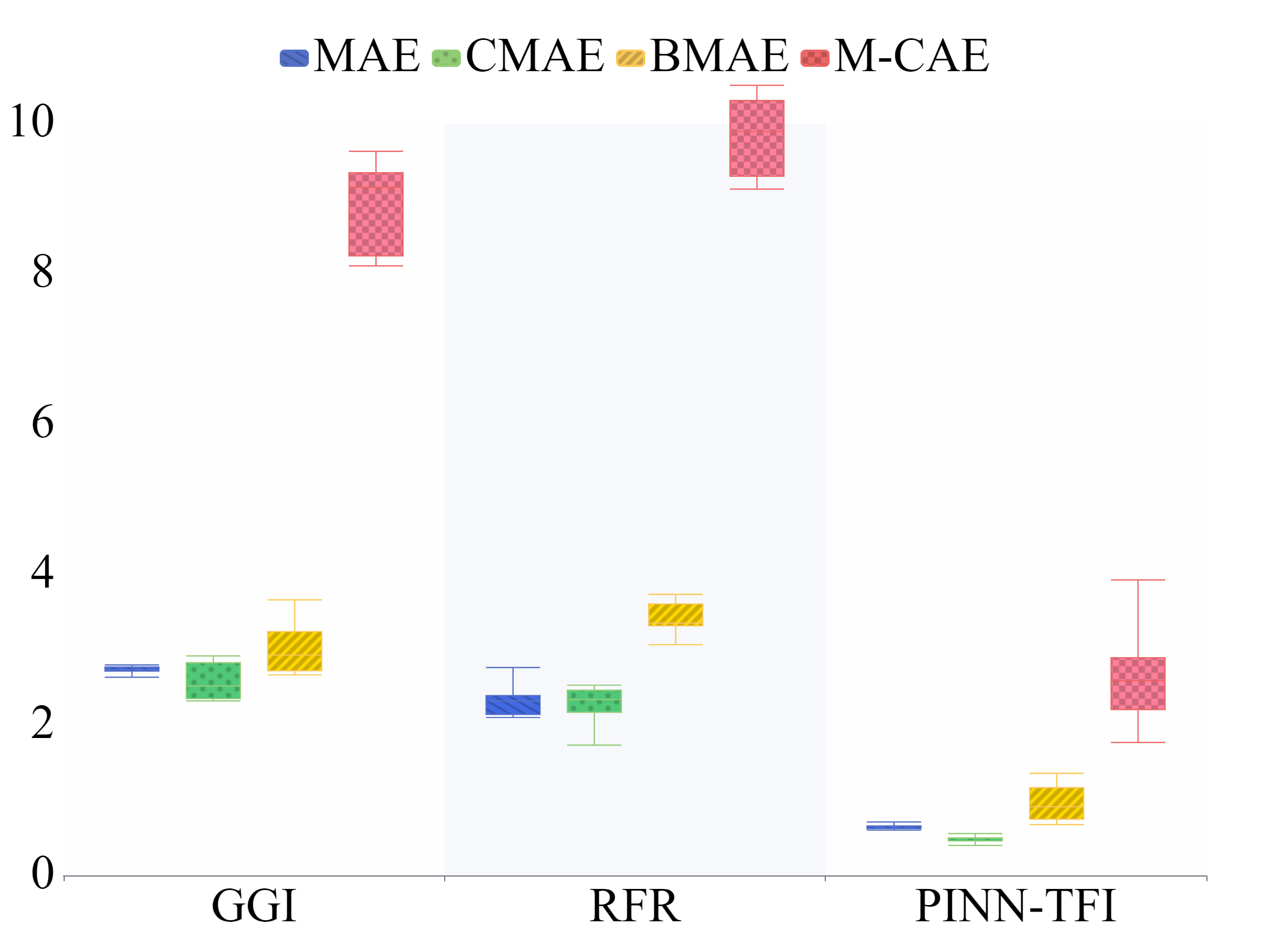}
	}
	\subfigure[Case 2 with LDS]{
		\includegraphics[width=0.3\linewidth]{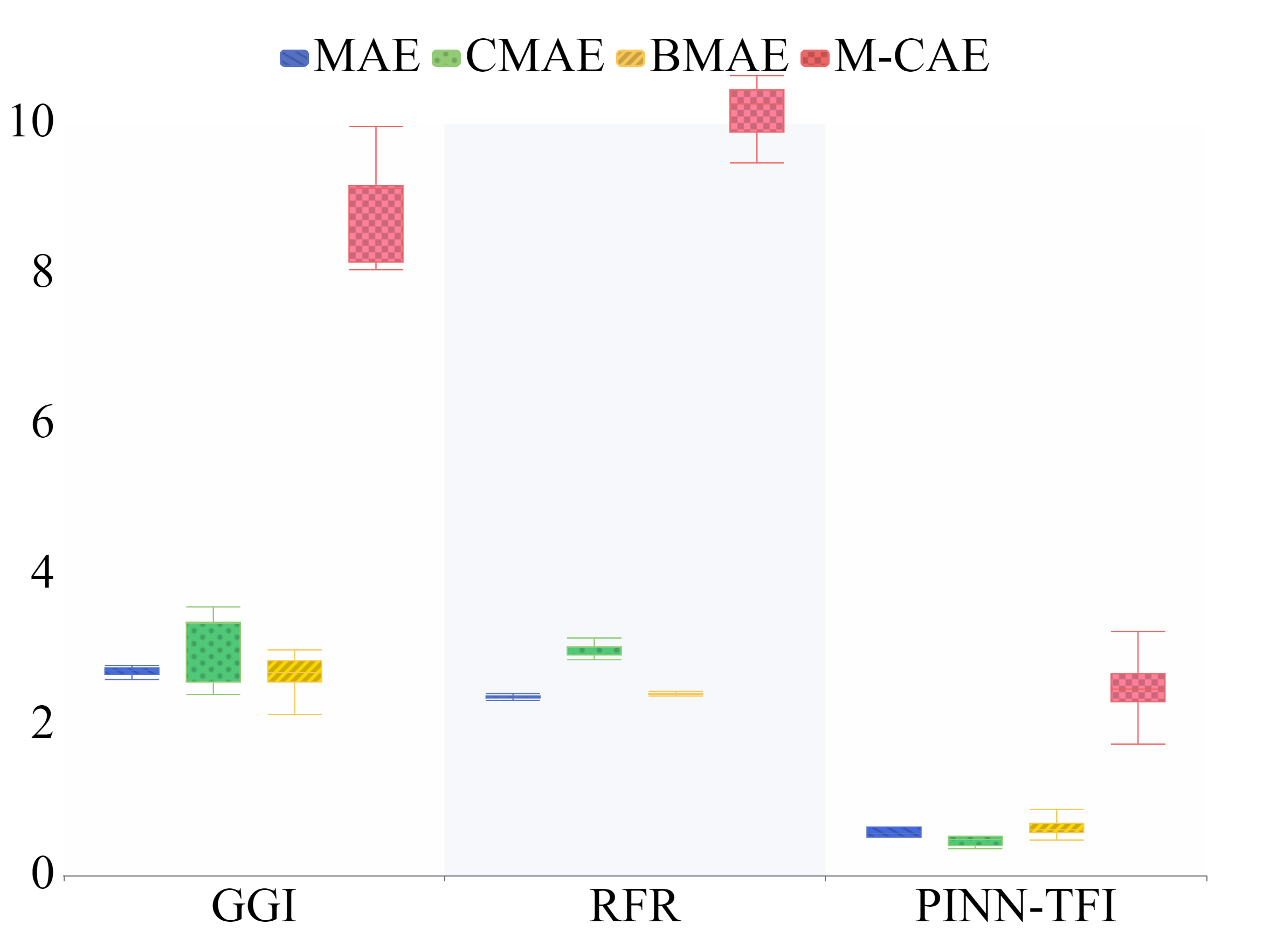}
	}
	\subfigure[Case 2 with GS]{
		\includegraphics[width=0.3\linewidth]{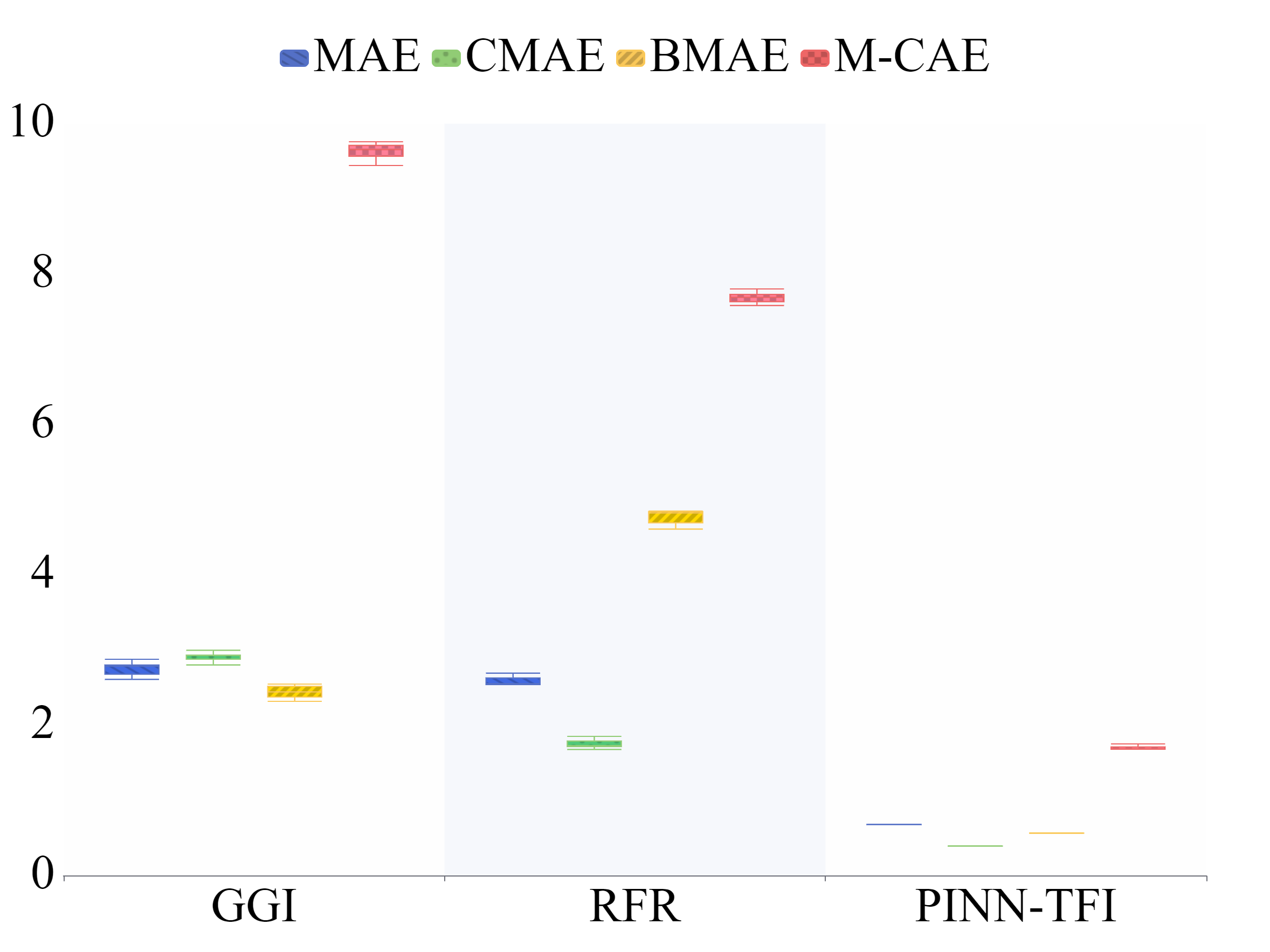}
	}
	\quad
	\subfigure[Case 3 with LHS]{
		\includegraphics[width=0.3\linewidth]{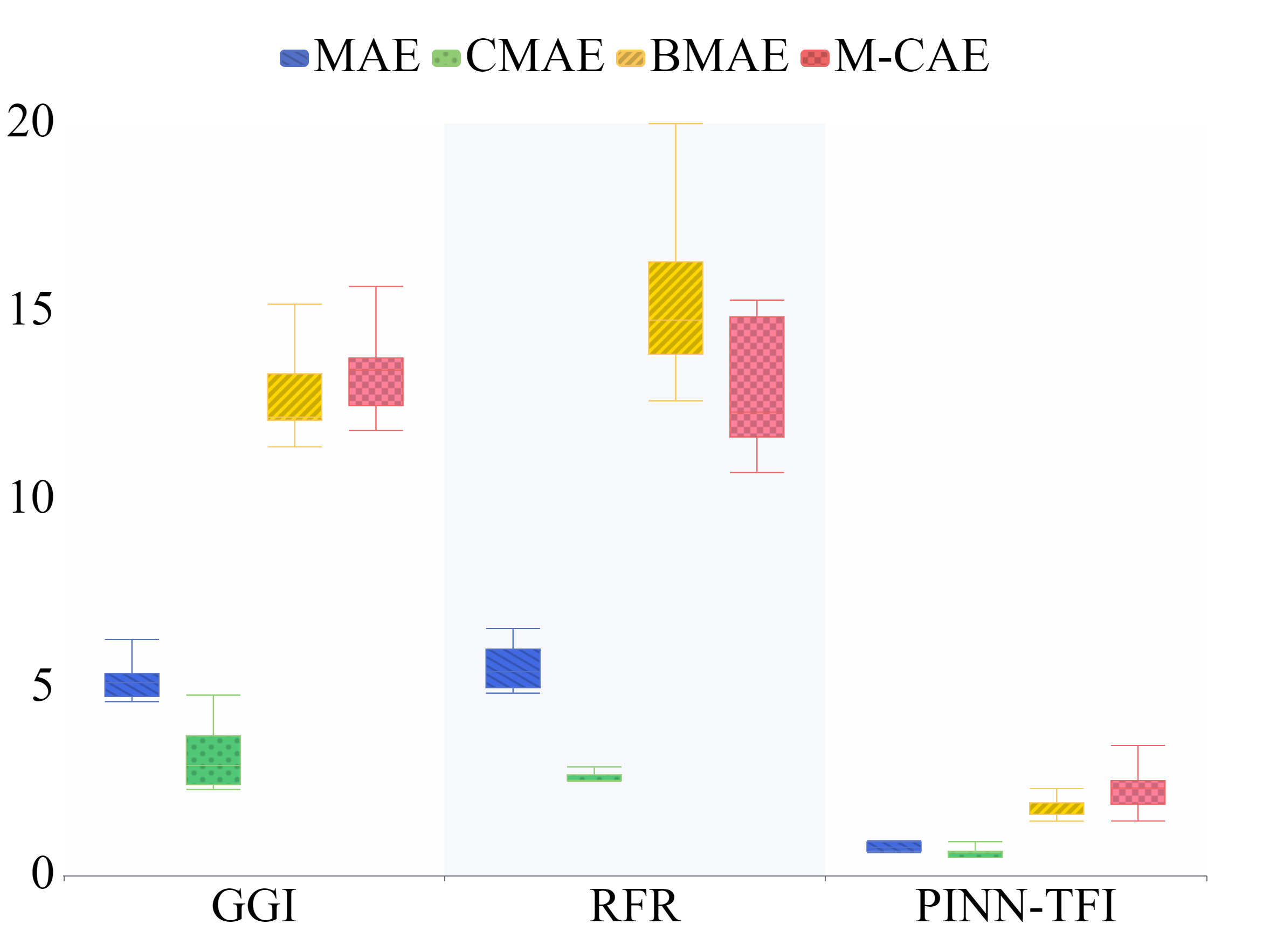}
	}
	\subfigure[Case 3 with LDS]{
		\includegraphics[width=0.3\linewidth]{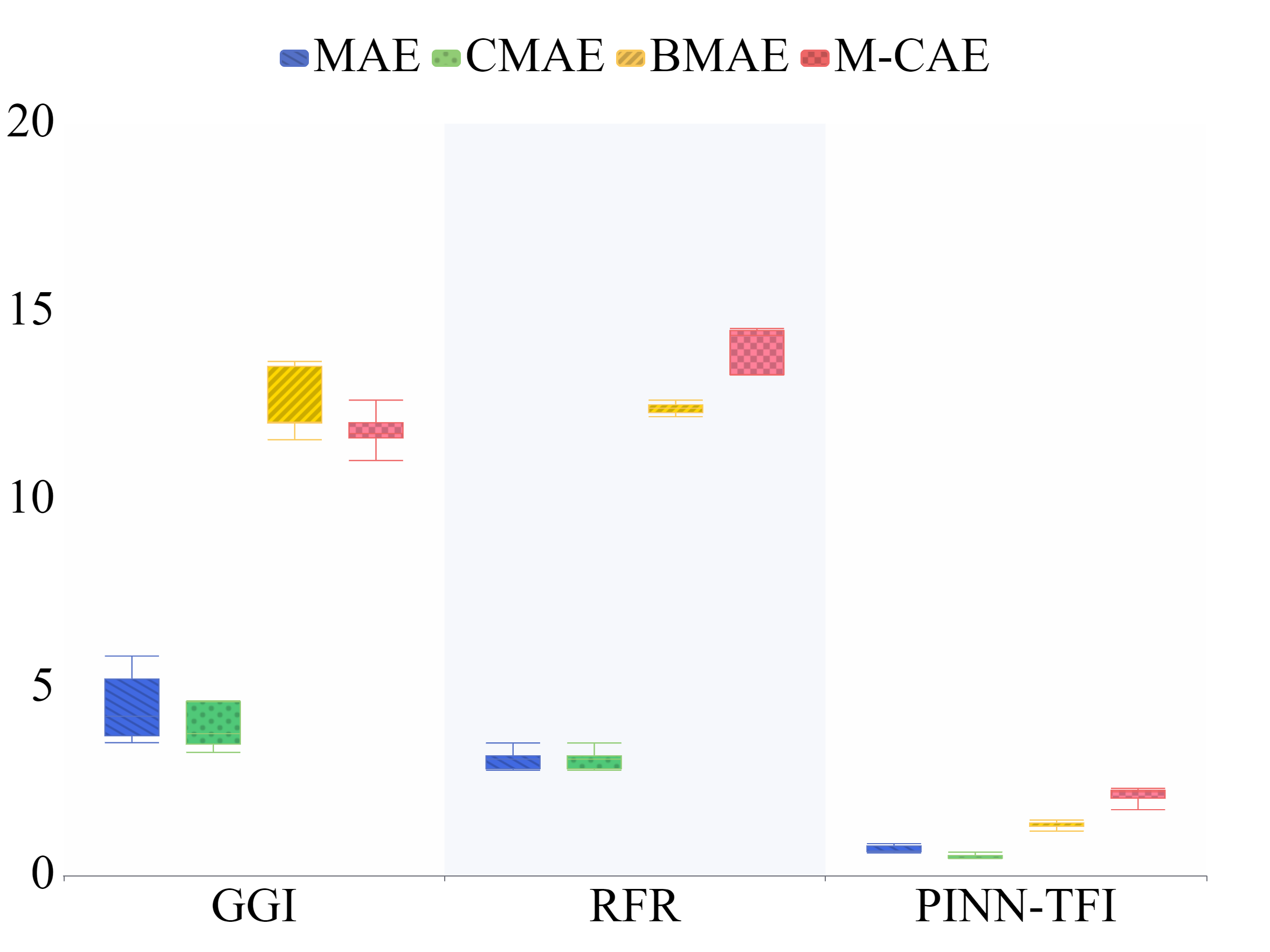}
	}
	\subfigure[Case 3 with GS]{
		\includegraphics[width=0.3\linewidth]{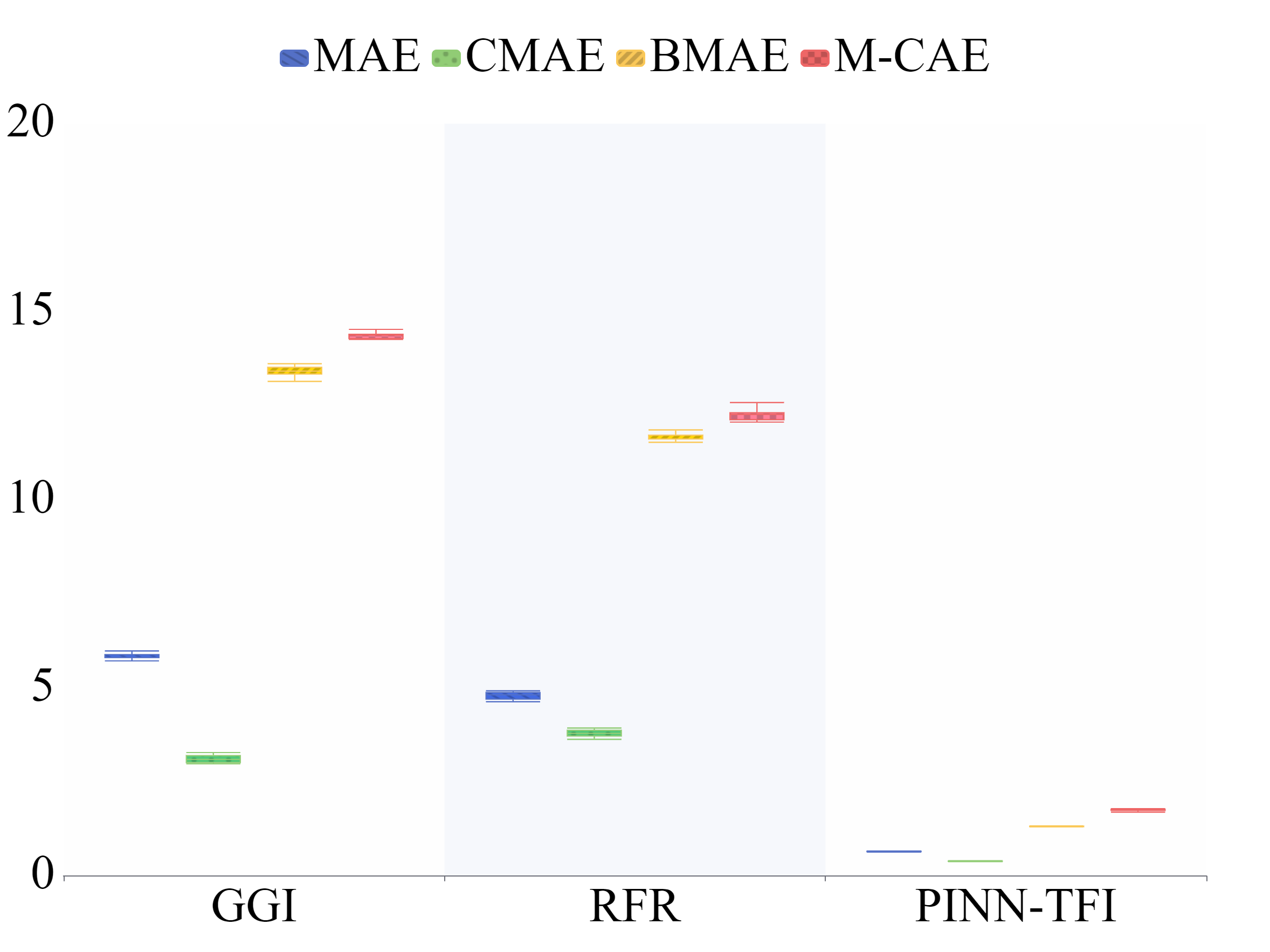}
	}
	\caption{Performance of the PINN-TFI method, Guassian process regession (GPR), and random forest regression (RFR) with 42 observations obtained from LHS, LDS and GS.}
	\label{fig:compare}
\end{figure*}
Fig. \ref{fig:compare} shows the comparison result of these former methods with 42 observations after five independent repeated runs. For three cases, whatever observations are sampled from LHS, LDS, or GS, the PINN-TFI method outperforms GPR and RFR. The PINN-TFI method with 42 observations from LHS can obtain MAEs of 0.2883K, 0.6735K and 0.7735K over three cases. For observations from LDS, the PINN-TFI method obtains MAEs of 0.3538K, 0.5787K, and 0.7132K over three cases. For observations from GS, the PINN-TFI method obtains MAEs of 0.3538K, 0.5787K, and 0.7132K over three cases. Under the same observations, MAEs of GPR and RFR over three cases exceed 1.5K, 2K, and 4.5K, respectively. For different cases, performances of GPR and RFR change dramatically. Especially for the most complicated case 3, B-MAEs and M-CAEs of GPR and RFR exceed 10, which are far more than that of the PINN-TFI method. In addition, similar performances of the PINN-TFI method under repeated runs indicate that the PINN-TFI method is more stable than GPR and RFR. In a word, the PINN-TFI method has a better reconstructed performance and is more suitable for the TFI-HSS task.

\subsection{TFI-HSS with noise observations}
To investigate the effectiveness of the CMCN-PSO method, this work first finds the optimal positions of noise observations and then uses them by the PINN-TFI method to solve the TFI-HSS task with noise observations. Concretely, we assume that there are 150 sets of positions under the same number of observations, where 50 sets are sampled from LHS, 50 sets are sampled from LDS, and 50 set is sampled from GS. The CMCN-PSO method is used to find the optimal positions and the PINN-TFI method with those noise observations is used to reconstruct the temperature field. Following we mainly present optimal positions of CMCN-PSO method with the different number of noise observations, the performance of the CMCN-PSO method with the different number of noise observations and the performance of the CMCN-PSO method with different noise levels.

\subsubsection{Optimal positions of CMCN-PSO method with the different number of noise observations}

This work ranks 50 sets of positions from LHS and LDS, respectively. The CMCN-PSO method is used to find the optimal set of positions and the PINN-TFI method uses optimal positions to reconstruct the temperature field. In the CMCN-PSO method, the two-dimensional domain is meshed as a $50\times50$ grid. To study the effect of additive noise in observations, this work creates noise observations from the true data as follows:
\begin{equation}
T_{obs}^{noise}=T_{obs}^{true}+\epsilon T_{obs}^{\text {true }} \mathcal{G}(0,1),
\end{equation}
where $T_{obs}^{noise}$ and $T_{obs}^{true}$ represent observations with and without noises, respectively. The $\epsilon$ determines the noise level and $\mathcal{G}(0,1)$ is a random value sampled from the Gaussian distribution with mean and standard deviation of 0 and 1. Sets of positions randomly sampled from LHS and LDS are used as baselines.

\begin{figure*}[!htbp]
	\centering
	\includegraphics[width=0.7\linewidth]{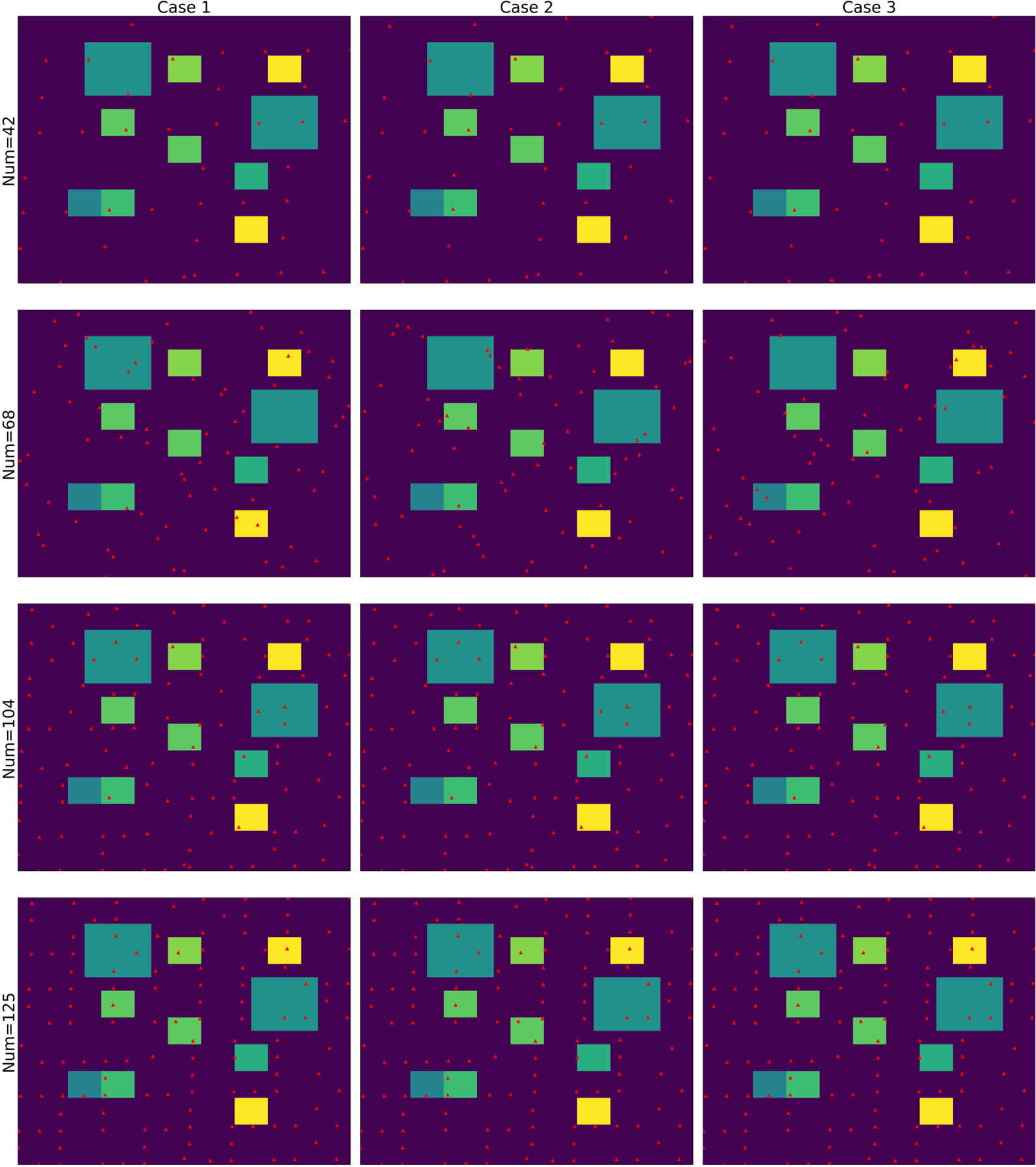}
	\caption{The optimal positions under different number of observations by the CMCN-PSO method.}
	\label{fig:CMCN_position}
\end{figure*}

Table \ref{tab:optimal set} shows the optimal set of positions with the CMCN-PSO method under the different number. The condition number of optimal positions is much smaller than that of positions from LHS and LDS. For three cases, optimal positions are different under 42 and 68 observations but are the same under 104 and 125 observations. It can be seen that the more the number of observations, the less boundary conditions affect positions of observations. It also means that when the number of observations is large enough, optima positions under different boundaries are the same. In addition, as boundary conditions of case 2 and case 3 are similar, the condition numbers obtained are also relatively close. Fig. \ref{fig:CMCN_position} shows the optimal positions under different number of observations. 
For 42 observations, the number of observations is relatively small and the distribution of observations is mainly concentrated on components and boundaries. For 125 observations, the number of observations is relatively large, and observations are distributed around the components, on the components, and on the boundary.

\begin{table*}[!htbp]
	\caption{The optimal set of positions with the CMCN-PSO method under different number of observations.}
	\label{tab:optimal set}
	\centering
	\scalebox{0.74}{
		\begin{tabular}{cccclcclcc}
			\hline
			\multirow{2}{*}{Num} & \multirow{2}{*}{Method} & \multicolumn{2}{c}{Case 1}   &  & \multicolumn{2}{c}{Case 2}   &  & \multicolumn{2}{c}{Case 3}   \\ \cline{3-4} \cline{6-7} \cline{9-10} 
			&                         & position & condition number  &  & position & condition number  &  & position & condition number  \\ \hline
			\multirow{3}{*}{42}  & CMCN-PSO                & 45(LDS)  & \textbf{154.7281} &  & 23(LDS)  & \textbf{154.3268} &  & 18(LDS)  & \textbf{158.3182} \\
			& LHS                     & -        & 250.8578          &  & -        & 332.8663          &  & -        & 265.5762          \\
			& LDS                     & -        & 201.7295          &  & -        & 245.5780          &  & -        & 259.9114          \\ \hline
			\multirow{3}{*}{68}  & CMCN-PSO                & 0(LHS)   & \textbf{128.7127} &  & 37(LHS)  & \textbf{135.1052} &  & 13(LHS)  & \textbf{138.1758} \\
			& LHS                     & -        & 224.8941          &  & -        & 226.2464          &  & -        & 267.6193          \\
			& LDS                     & -        & 166.6221          &  & -        & 169.4719          &  & -        & 241.2854          \\ \hline
			\multirow{3}{*}{104} & CMCN-PSO                & 5(LDS)   & \textbf{85.5572}  &  & 5(LDS)   & \textbf{86.0331}  &  & 5(LDS)   & \textbf{86.0380}  \\
			& LHS                     & -        & 165.8742          &  & -        & 167.8742          &  & -        & 150.0379          \\
			& LDS                     & -        & 158.8168          &  & -        & 142.2804          &  & -        & 127.8972          \\ \hline
			\multirow{3}{*}{125} & CMCN-PSO                & 7(LDS)   & \textbf{84.1118}  &  & 7(LDS)   & \textbf{84.6107}  &  & 7(LDS)   & \textbf{84.6015}  \\
			& LHS                     & -        & 140.2223          &  & -        & 130.2663          &  & -        & 133.6214          \\
			& LDS                     & -        & 129.6307          &  & -        & 131.3137          &  & -        & 122.4524866       \\ \hline
		\end{tabular}
	}
\end{table*}

\subsubsection{Performance of the CMCN-PSO method with the different number of noise observations}

To validate the effectiveness of optimal positions under the different number of observations, this work uses optimal positions to reconstruct the temperature field by the PINN-TFI method. The noise level $\epsilon$ is set to be $1 \%$. Two sets of positions randomly sampled from LHS and LDS are used as baselines.

Table \ref{tab:noise_num} shows the performance of optimal positions of the CMCN-PSO method and positions from LHS and LDS under the different number of observations when $\epsilon=1 \%$. For three cases, MAE and BMAE of the CMCN-PSO method are the smallest under the same number. For CMAE and M-CAE, the CMCN-PSO method can obtain minimum values in most instances. In addition, for 104 and 125 observations, four evaluation metrics of the CMCN-PSO method are best. When observations are perturbed by noises, as expected, the more the number of observations, the better reconstruction performance of the temperature field. Under 42 observations, the MAE of CMCN-PSO for case 1 is around 0.1K smaller than that of LHS, and MAEs of the CMCN-PSO method for case 2 and case 3 are around 0.5K and 0.4K smaller than that of LHS, respectively. This also indicates that the CMCN-PSO method might alleviate the effect of noises more significantly for more complex boundaries.

\begin{table*}[!htbp]
	\caption{Performance of the optimal positions of the CMCN-PSO method and positions from LHS and LDS under different number of observations when $\epsilon=0.01$. The best results under different numbers are highlight.}
	\label{tab:noise_num}
	\centering
	\scalebox{0.74}{
		\begin{tabular}{ccllllllllllllll}
			\hline
			\multirow{2}{*}{Num} & \multirow{2}{*}{Method} & \multicolumn{4}{c}{Case 1}                                                                                 &  & \multicolumn{4}{c}{Case 2}                                                                                 &  & \multicolumn{4}{c}{Case 3}                                                                                 \\ \cline{3-6} \cline{8-11} \cline{13-16} 
			&                         & \multicolumn{1}{c}{MAE} & \multicolumn{1}{c}{CMAE} & \multicolumn{1}{c}{BMAE} & \multicolumn{1}{c}{M-CAE} &  & \multicolumn{1}{c}{MAE} & \multicolumn{1}{c}{CMAE} & \multicolumn{1}{c}{BMAE} & \multicolumn{1}{c}{M-CAE} &  & \multicolumn{1}{c}{MAE} & \multicolumn{1}{c}{CMAE} & \multicolumn{1}{c}{BMAE} & \multicolumn{1}{c}{M-CAE} \\ \hline
			& CMCN-PSO                & \textbf{0.8849}         & 0.9684                   & \textbf{0.8132}          & \textbf{3.0784}           &  & \textbf{1.2722}         & \textbf{1.3888}          & \textbf{1.1286}          & 4.5438                    &  & \textbf{2.2899}         & \textbf{1.8989}          & \textbf{3.336}           & \textbf{6.3405}           \\
			42                   & LHS                     & 0.9861                  & 1.1839                   & 1.2454                   & 3.20171                   &  & 1.7742                  & 2.0686                   & 1.6654                   & 5.4797                    &  & 2.6843                  & 2.6879                   & 4.0735                   & 9.4054                    \\
			& LDS                     & 0.9064                  & \textbf{0.9611}          & 1.0361                   & 3.0813                    &  & 1.3412                  & 1.5411                   & 1.9054                   & \textbf{3.9012}           &  & 2.4882                  & 2.6626                   & 4.1432                   & 8.1555                    \\ \hline
			& CMCN-PSO                & \textbf{0.7057}         & \textbf{0.8385}          & \textbf{0.6280}          & 3.2573                    &  & \textbf{1.0757}         & 1.4777                   & \textbf{1.2119}          & \textbf{3.0310}           &  & \textbf{2.0185}         & \textbf{1.5687}          & 3.0101                   & \textbf{4.7467}           \\
			68                   & LHS                     & 0.9227                  & 1.0771                   & 0.7690                   & \textbf{3.2407}           &  & 1.3559                  & \textbf{1.0771}          & 2.4239                   & 3.1023                    &  & 2.4958                  & 1.8365                   & 5.3349                   & 5.6113                    \\
			& LDS                     & 0.8946                  & 0.9381                   & 0.8918                   & 3.4484                    &  & 1.2111                  & 1.3066                   & 1.6656                   & 3.4262                    &  & 2.385                   & 1.6333                   & \textbf{2.7971}          & 6.5685                    \\ \hline
			& CMCN-PSO                & \textbf{0.6502}         & \textbf{0.5978}          & \textbf{0.6270}          & \textbf{1.6969}           &  & \textbf{0.8929}         & \textbf{0.8374}          & \textbf{1.1587}          & \textbf{2.5145}           &  & \textbf{1.3054}         & \textbf{1.3568}          & \textbf{1.6258}          & \textbf{3.7029}           \\
			104                  & LHS                     & 0.8604                  & 0.8682                   & 1.0192                   & 3.4401                    &  & 1.3010                  & 0.9180                   & 1.7530                   & 3.1858                    &  & 1.8954                  & 1.5200                   & 3.6248                   & 4.0593                    \\
			& LDS                     & 0.8577                  & 0.7621                   & 0.9487                   & 3.1495                    &  & 1.1517                  & 1.0744                   & 1.3964                   & 3.6843                    &  & 1.7339                  & 1.5567                   & 2.402                    & 4.9645                    \\ \hline
			& CMCN-PSO                & \textbf{0.5871}         & \textbf{0.4700}          & \textbf{0.7871}          & \textbf{1.8628}           &  & \textbf{0.7891}         & \textbf{0.8074}          & \textbf{1.0721}          & \textbf{2.1150}           &  & \textbf{1.1694}         & \textbf{1.0035}          & \textbf{2.0039}          & \textbf{3.1011}           \\
			125                  & LHS                     & 0.7888                  & 0.4846                   & 0.9219                   & 2.5283                    &  & 1.2531                  & 1.1884                   & 1.7325                   & 4.1278                    &  & 1.6627                  & 1.2016                   & 3.3963                   & 3.8837                    \\
			& LDS                     & 0.7011                  & 0.6131                   & 0.9556                   & 2.4351                    &  & 1.1676                  & 1.0332                   & 1.2427                   & 4.9933                    &  & 1.6328                  & 1.6830                   & 2.2841                   & 5.2096                    \\ \hline
		\end{tabular}
	}
\end{table*}


\subsubsection{Performance with different noise levels}
This experiment is designed to validate the effectiveness of the CMCN-PSO method under various noise levels. The number of observations is set to be 68. The noise level $\epsilon$ is chosen from
$\{0.25 \%$, $0.5 \%$, $1 \%$, $2 \%\}$. Two sets of positions randomly sampled from LHS and LDS are used as baselines.

Table \ref{tab:noise_level} shows the performance of optimal positions by the CMCN-PSO method and positions from LHS and LDS under different noise levels. Except for case 1 under the $1 \%$ noise level, MAEs of the CMCN-PSO method are best. The best MAE for case 1 under the $1 \%$ noise level tends to be $0.3908$K, which is very close to $0.7057$K by the CMCN-PSO method. It means that the CMCN-PSO method at different noise levels can achieve superior performance. 
As expected, the table shows that the predicted accuracy decreases as noise levels go up. Especially for case 3 under the $2 \%$ noise level, M-CAEs of LHS and LDS tend to be 16.1207K and 13.206K, which means a large noise level can cause a very terrible maximum error. But, the CMCN-PSO method can reduce the M-CAE to 7.8713K, which indicates that reasonable position selection can reduce the maximum error to a certain extent.
\begin{table*}[!htbp]
	\caption{Performance of optimal positions by the CMCN-PSO method and positions from LHS and LDS under different noise level when the number of observations is equal to 68. The best results under different numbers are highlight.}
	\label{tab:noise_level}
	\centering
	\scalebox{0.7}{
		\begin{tabular}{cccccclcccclcccc}
			\hline
			\multirow{2}{*}{$\epsilon$} & \multirow{2}{*}{Method} & \multicolumn{4}{c}{Case 1}                                             &  & \multicolumn{4}{c}{Case 2}                                             &  & \multicolumn{4}{c}{Case 3}                                             \\ \cline{3-6} \cline{8-11} \cline{13-16} 
			&                         & MAE             & CMAE            & BMAE            & M-CAE           &  & MAE             & CMAE            & BMAE            & M-CAE           &  & MAE             & CMAE            & BMAE            & M-CAE           \\ \hline
			& CMCN-PSO                & \textbf{0.2814} & \textbf{0.2437} & 0.2669          & 1.2041          &  & \textbf{0.3792} & 0.3258          & \textbf{0.6133} & 1.4276          &  & \textbf{0.4632} & \textbf{0.3642} & \textbf{1.0145} & \textbf{1.4645} \\
			$0.25 \%$                   & LHS                     & 0.3307          & 0.4341          & 0.3298          & 1.3273          &  & 0.6474          & 0.5319          & 0.7124          & 1.9306          &  & 0.8011          & 0.7824          & 1.5316          & 2.3995          \\
			& LDS                     & 0.2975          & 0.3523          & \textbf{0.2433} & \textbf{1.0332} &  & 0.4098          & \textbf{0.2934} & 0.7878          & \textbf{1.1030} &  & 0.6871          & 0.5647          & 1.4279          & 1.6981          \\ \hline
			& CMCN-PSO                & 0.4027          & 0.3991          & \textbf{0.3262} & \textbf{1.5099} &  & \textbf{0.5471} & \textbf{0.5348} & \textbf{0.6267} & 1.8197          &  & \textbf{0.7231} & \textbf{0.5639} & 1.4998          & \textbf{1.7774} \\
			$0.5 \%$                   & LHS                     & 0.4911          & 0.6706          & 0.4566          & 2.1253          &  & 1.1086          & 1.0223          & 1.6337          & 3.4122          &  & 1.4456          & 1.5260          & 2.1647          & 4.8312          \\
			& LDS                     & \textbf{0.3908} & \textbf{0.3447} & 0.4023          & 1.9413          &  & 0.5611          & 0.4383          & 0.9114          & \textbf{1.7046} &  & 0.8014          & 0.7939          & \textbf{1.3544} & 3.2351          \\ \hline
			& CMCN-PSO                & \textbf{0.7057} & \textbf{0.8385} & \textbf{0.6282} & 3.2573          &  & \textbf{1.0757} & 1.4777          & \textbf{1.2119} & \textbf{3.0310} &  & \textbf{2.0185} & \textbf{1.5687} & 3.0101          & \textbf{4.7467} \\
			$1 \%$                  & LHS                     & 0.9227          & 1.0771          & 0.7691          & \textbf{3.2407} &  & 1.3559          & \textbf{1.0771} & 2.4239          & 3.1023          &  & 2.4958          & 1.8365          & 5.3349          & 5.6113          \\
			& LDS                     & 0.8946          & 0.9381          & 0.8918          & 3.4484          &  & 1.2111          & 1.3066          & 1.6656          & 3.4262          &  & 2.3850          & 1.6333          & \textbf{2.7971} & 6.5685          \\ \hline
			& CMCN-PSO                & \textbf{1.1589} & 1.2517          & \textbf{1.0908} & \textbf{4.8594} &  & \textbf{2.0832} & 1.9856          & \textbf{2.0051} & \textbf{7.9203} &  & \textbf{3.2935} & \textbf{2.0461} & 5.3258          & \textbf{7.8713} \\
			$2 \%$                  & LHS                     & 1.7186          & 1.9954          & 1.3693          & 5.5975          &  & 3.2203          & 3.4256          & 2.9096          & 10.6826         &  & 4.4206          & 4.7631          & 5.8981          & 16.1207         \\
			& LDS                     & 1.4886          & \textbf{1.1975} & 1.2756          & 6.1999          &  & 2.1891          & \textbf{1.7512} & 2.4179          & 9.4052          &  & 3.4007          & 3.7081          & \textbf{4.1591} & 13.2061         \\ \hline
		\end{tabular}
	}
\end{table*}

\section{Conclusions}
\label{sec:con}
In this paper, the TFI-HSS task is defined by giving the formula expression. Then, we further develop a physics-informed neural network-based temperature field inversion (PINN-TFI) method and a coefficient matrix condition number based position selection of observations (CMCN-PSO) method. 
To solve the TFI-HSS task with limited observations, the PINN-TFI method firstly encodes physics and data constrains terms into the loss function. Then, the problem of TFI-HSS is transformed into an optimization problem of minimizing the loss function. Finally, the PINN model is trained to acquire the surrogate model of the reconstructed temperature field, where transfer learning strategy is used to accelerate the training process. However, noise observations significantly affect the reconstruction performance of the PINN-TFI method. To this end, the CMCN-PSO method is proposed to alleviate the effect of noise observations. We have proved the upper bound of the reconstruction error is related to the coefficient matrix condition number, which is determined by positions of noise observations. Subsequently, the condition number is used as the principle to find optima positions. After that, the PINN-TFI method with optima positions can achieve better reconstructed performance. Experiments have shown that the PINN-TFI method can reconstruct the temperature field well in the small data setting, where transfer learning strategy can accelerate training process significantly. Besides, the CMCN-PSO method can find optimal positions to alleviate the effect of noises under different noise levels and numbers of observations, making the PINN-TFI method to reconstruct a more robust temperature field. In future work, reducing the number of observations while maintaining reconstructed performances is an interesting research point.

\section*{Declaration of competing interest}
The authors declare that they have no known competing financial interests or personal relationships that could have appeared to influence the work reported in this paper.

\section*{Replication of results}
The code can be downloaded at: \url{https://github.com/liuxu97531/PINN_TFI-HSS}.

\section*{Acknowledgements} 
This work was supported by National Natural Science Foundation of China under Grant No.11725211, 52005505, and 62001502.

\bibliographystyle{unsrtnat}
\bibliography{mybibfile}  

\end{document}